\documentclass{article}
\usepackage{arxiv}
\usepackage[T1]{fontenc}
\usepackage[utf8]{inputenc}
\usepackage[numbers,sort&compress]{natbib}
\usepackage{amssymb}
\usepackage{colortbl}
\usepackage{float}
\usepackage[colorlinks=true, urlcolor=blue]{hyperref}
\usepackage{xcolor}
\definecolor{oursblue}{RGB}{214,232,248}  % light blue for our entries in the Learn2Reg table

\usepackage{lineno}
\usepackage{amsmath}
\usepackage{graphicx}

\usepackage{multirow}
\usepackage{booktabs}
\usepackage{longtable}
\usepackage{tabularx}
\usepackage{pdflscape}
\usepackage{placeins}  % \FloatBarrier: keep the main-text floats before the references
\usepackage{needspace}
\newsavebox{\fittablebox}
\newcommand{\fittable}[1]{\begingroup\sbox{\fittablebox}{#1}%
\ifdim\wd\fittablebox>\linewidth
\resizebox{\linewidth}{!}{\usebox{\fittablebox}}%
\else\usebox{\fittablebox}\fi\endgroup}
\newif\ifextendedsupplement
\newcommand{\amosmrmanual}{16}
\newcommand{\amosmrtotal}{58}
\newcommand{\totalsmrmanual}{218}
\newcommand{\totalsmrtotal}{374}

\providecommand{\res}[1]{\ifcsname res@#1\endcsname\csname res@#1\endcsname\else\textbf{??\detokenize{#1}}\fi}
\expandafter\def\csname res@amos.cases\endcsname{20}
\expandafter\def\csname res@amos.ims25.assd\endcsname{1.77}
\expandafter\def\csname res@amos.ims25.ci\endcsname{[0.789, 0.827]}
\expandafter\def\csname res@amos.ims25.dice\endcsname{0.808}
\expandafter\def\csname res@amos.ims25.dicesd\endcsname{0.045}
\expandafter\def\csname res@amos.ims25.hd95\endcsname{6.3}
\expandafter\def\csname res@amos.ims25.nsd\endcsname{0.485}
\expandafter\def\csname res@amos.ims25.pthreed\endcsname{$p=0.051$}
\expandafter\def\csname res@amos.ims3d.assd\endcsname{1.60}
\expandafter\def\csname res@amos.ims3d.ci\endcsname{[0.797, 0.834]}
\expandafter\def\csname res@amos.ims3d.dice\endcsname{0.816}
\expandafter\def\csname res@amos.ims3d.dicesd\endcsname{0.043}
\expandafter\def\csname res@amos.ims3d.diff\endcsname{-0.008}
\expandafter\def\csname res@amos.ims3d.hd95\endcsname{5.5}
\expandafter\def\csname res@amos.ims3d.nsd\endcsname{0.485}
\expandafter\def\csname res@amos.ims3d.p\endcsname{$p=0.064$}
\expandafter\def\csname res@amos.labels\endcsname{36}
\expandafter\def\csname res@amos.mrseg.assd\endcsname{2.16}
\expandafter\def\csname res@amos.mrseg.ci\endcsname{[0.765, 0.808]}
\expandafter\def\csname res@amos.mrseg.dice\endcsname{0.787}
\expandafter\def\csname res@amos.mrseg.dicesd\endcsname{0.050}
\expandafter\def\csname res@amos.mrseg.diff\endcsname{+0.020}
\expandafter\def\csname res@amos.mrseg.hd95\endcsname{7.7}
\expandafter\def\csname res@amos.mrseg.nsd\endcsname{0.435}
\expandafter\def\csname res@amos.mrseg.p\endcsname{$p=0.011$}
\expandafter\def\csname res@amos.mrseg.pthreed\endcsname{$p<0.001$}
\expandafter\def\csname res@amos.nv.assd\endcsname{1.97}
\expandafter\def\csname res@amos.nv.ci\endcsname{[0.797, 0.838]}
\expandafter\def\csname res@amos.nv.dice\endcsname{0.818}
\expandafter\def\csname res@amos.nv.dicesd\endcsname{0.048}
\expandafter\def\csname res@amos.nv.diff\endcsname{-0.008}
\expandafter\def\csname res@amos.nv.hd95\endcsname{7.4}
\expandafter\def\csname res@amos.nv.nsd\endcsname{0.487}
\expandafter\def\csname res@amos.nv.p\endcsname{$p=0.46$}
\expandafter\def\csname res@amos.nv.pthreed\endcsname{$p=1.00$}
\expandafter\def\csname res@amos.ts.assd\endcsname{1.94}
\expandafter\def\csname res@amos.ts.ci\endcsname{[0.773, 0.808]}
\expandafter\def\csname res@amos.ts.dice\endcsname{0.791}
\expandafter\def\csname res@amos.ts.dicesd\endcsname{0.040}
\expandafter\def\csname res@amos.ts.diff\endcsname{+0.018}
\expandafter\def\csname res@amos.ts.hd95\endcsname{6.5}
\expandafter\def\csname res@amos.ts.nsd\endcsname{0.425}
\expandafter\def\csname res@amos.ts.p\endcsname{$p=0.005$}
\expandafter\def\csname res@amos.ts.pthreed\endcsname{$p<0.001$}
\expandafter\def\csname res@conscbct.cads.agree\endcsname{0.489}
\expandafter\def\csname res@conscbct.cads.agreementp\endcsname{$p<0.001$}
\expandafter\def\csname res@conscbct.cads.ct\endcsname{0.999}
\expandafter\def\csname res@conscbct.cads.ctp\endcsname{$p<0.001$}
\expandafter\def\csname res@conscbct.cads.gap\endcsname{-0.510}
\expandafter\def\csname res@conscbct.cads.gapp\endcsname{$p<0.001$}
\expandafter\def\csname res@conscbct.cads.mod\endcsname{0.489}
\expandafter\def\csname res@conscbct.cads.modp\endcsname{$p<0.001$}
\expandafter\def\csname res@conscbct.cases\endcsname{139}
\expandafter\def\csname res@conscbct.ims25.agree\endcsname{0.864}
\expandafter\def\csname res@conscbct.ims25.ct\endcsname{0.877}
\expandafter\def\csname res@conscbct.ims25.gap\endcsname{-0.058}
\expandafter\def\csname res@conscbct.ims25.mod\endcsname{0.819}
\expandafter\def\csname res@conscbct.ims3d.agree\endcsname{0.881}
\expandafter\def\csname res@conscbct.ims3d.agreementp\endcsname{$p<0.001$}
\expandafter\def\csname res@conscbct.ims3d.ct\endcsname{0.896}
\expandafter\def\csname res@conscbct.ims3d.ctp\endcsname{$p<0.001$}
\expandafter\def\csname res@conscbct.ims3d.gap\endcsname{-0.054}
\expandafter\def\csname res@conscbct.ims3d.gapp\endcsname{$p=0.075$}
\expandafter\def\csname res@conscbct.ims3d.mod\endcsname{0.842}
\expandafter\def\csname res@conscbct.ims3d.modp\endcsname{$p<0.001$}
\expandafter\def\csname res@conscbct.nv.agree\endcsname{0.692}
\expandafter\def\csname res@conscbct.nv.agreementp\endcsname{$p<0.001$}
\expandafter\def\csname res@conscbct.nv.ct\endcsname{0.823}
\expandafter\def\csname res@conscbct.nv.ctp\endcsname{$p<0.001$}
\expandafter\def\csname res@conscbct.nv.gap\endcsname{-0.152}
\expandafter\def\csname res@conscbct.nv.gapp\endcsname{$p<0.001$}
\expandafter\def\csname res@conscbct.nv.mod\endcsname{0.671}
\expandafter\def\csname res@conscbct.nv.modp\endcsname{$p<0.001$}
\expandafter\def\csname res@conscbct.ts.agree\endcsname{0.492}
\expandafter\def\csname res@conscbct.ts.agreementp\endcsname{$p<0.001$}
\expandafter\def\csname res@conscbct.ts.ct\endcsname{0.883}
\expandafter\def\csname res@conscbct.ts.ctp\endcsname{$p=0.19$}
\expandafter\def\csname res@conscbct.ts.gap\endcsname{-0.430}
\expandafter\def\csname res@conscbct.ts.gapp\endcsname{$p<0.001$}
\expandafter\def\csname res@conscbct.ts.mod\endcsname{0.454}
\expandafter\def\csname res@conscbct.ts.modp\endcsname{$p<0.001$}
\expandafter\def\csname res@consmri.cases\endcsname{52}
\expandafter\def\csname res@consmri.ims25.agree\endcsname{0.785}
\expandafter\def\csname res@consmri.ims25.ct\endcsname{0.810}
\expandafter\def\csname res@consmri.ims25.gap\endcsname{-0.001}
\expandafter\def\csname res@consmri.ims25.mod\endcsname{0.808}
\expandafter\def\csname res@consmri.ims3d.agree\endcsname{0.790}
\expandafter\def\csname res@consmri.ims3d.agreementp\endcsname{$p=0.41$}
\expandafter\def\csname res@consmri.ims3d.ct\endcsname{0.815}
\expandafter\def\csname res@consmri.ims3d.ctp\endcsname{$p=0.12$}
\expandafter\def\csname res@consmri.ims3d.gap\endcsname{-0.000}
\expandafter\def\csname res@consmri.ims3d.gapp\endcsname{$p=0.86$}
\expandafter\def\csname res@consmri.ims3d.mod\endcsname{0.814}
\expandafter\def\csname res@consmri.ims3d.modp\endcsname{$p=0.12$}
\expandafter\def\csname res@consmri.mrseg.agree\endcsname{0.762}
\expandafter\def\csname res@consmri.mrseg.agreementp\endcsname{$p=0.027$}
\expandafter\def\csname res@consmri.mrseg.ct\endcsname{0.834}
\expandafter\def\csname res@consmri.mrseg.ctp\endcsname{$p<0.001$}
\expandafter\def\csname res@consmri.mrseg.gap\endcsname{-0.040}
\expandafter\def\csname res@consmri.mrseg.gapp\endcsname{$p<0.001$}
\expandafter\def\csname res@consmri.mrseg.mod\endcsname{0.793}
\expandafter\def\csname res@consmri.mrseg.modp\endcsname{$p=0.41$}
\expandafter\def\csname res@consmri.nv.agree\endcsname{0.588}
\expandafter\def\csname res@consmri.nv.agreementp\endcsname{$p<0.001$}
\expandafter\def\csname res@consmri.nv.ct\endcsname{0.772}
\expandafter\def\csname res@consmri.nv.ctp\endcsname{$p<0.001$}
\expandafter\def\csname res@consmri.nv.gap\endcsname{-0.118}
\expandafter\def\csname res@consmri.nv.gapp\endcsname{$p<0.001$}
\expandafter\def\csname res@consmri.nv.mod\endcsname{0.654}
\expandafter\def\csname res@consmri.nv.modp\endcsname{$p<0.001$}
\expandafter\def\csname res@consmri.ts.agree\endcsname{0.673}
\expandafter\def\csname res@consmri.ts.agreementp\endcsname{$p<0.001$}
\expandafter\def\csname res@consmri.ts.ct\endcsname{0.795}
\expandafter\def\csname res@consmri.ts.ctp\endcsname{$p=0.18$}
\expandafter\def\csname res@consmri.ts.gap\endcsname{+0.013}
\expandafter\def\csname res@consmri.ts.gapp\endcsname{$p=0.42$}
\expandafter\def\csname res@consmri.ts.mod\endcsname{0.808}
\expandafter\def\csname res@consmri.ts.modp\endcsname{$p=0.86$}
\expandafter\def\csname res@gap.cads.agree\endcsname{0.774}
\expandafter\def\csname res@gap.cads.agreementp\endcsname{$p<0.001$}
\expandafter\def\csname res@gap.cads.cbct\endcsname{0.740}
\expandafter\def\csname res@gap.cads.ct\endcsname{0.917}
\expandafter\def\csname res@gap.cads.dice_cbctp\endcsname{$p<0.001$}
\expandafter\def\csname res@gap.cads.gap\endcsname{-0.176}
\expandafter\def\csname res@gap.cads.gapci\endcsname{[-0.185, -0.168]}
\expandafter\def\csname res@gap.cads.gapp\endcsname{$p<0.001$}
\expandafter\def\csname res@gap.cases\endcsname{497}
\expandafter\def\csname res@gap.ims25.agree\endcsname{0.908}
\expandafter\def\csname res@gap.ims25.cbct\endcsname{0.833}
\expandafter\def\csname res@gap.ims25.ct\endcsname{0.858}
\expandafter\def\csname res@gap.ims25.gap\endcsname{-0.025}
\expandafter\def\csname res@gap.ims25.gapci\endcsname{[-0.028, -0.023]}
\expandafter\def\csname res@gap.ims3d.agree\endcsname{0.926}
\expandafter\def\csname res@gap.ims3d.agreementp\endcsname{$p<0.001$}
\expandafter\def\csname res@gap.ims3d.cbct\endcsname{0.855}
\expandafter\def\csname res@gap.ims3d.ct\endcsname{0.877}
\expandafter\def\csname res@gap.ims3d.dice_cbctp\endcsname{$p<0.001$}
\expandafter\def\csname res@gap.ims3d.gap\endcsname{-0.022}
\expandafter\def\csname res@gap.ims3d.gapci\endcsname{[-0.024, -0.020]}
\expandafter\def\csname res@gap.ims3d.gapp\endcsname{$p<0.001$}
\expandafter\def\csname res@gap.nv.agree\endcsname{0.748}
\expandafter\def\csname res@gap.nv.agreementp\endcsname{$p<0.001$}
\expandafter\def\csname res@gap.nv.cbct\endcsname{0.710}
\expandafter\def\csname res@gap.nv.ct\endcsname{0.862}
\expandafter\def\csname res@gap.nv.dice_cbctp\endcsname{$p<0.001$}
\expandafter\def\csname res@gap.nv.gap\endcsname{-0.153}
\expandafter\def\csname res@gap.nv.gapci\endcsname{[-0.162, -0.144]}
\expandafter\def\csname res@gap.nv.gapp\endcsname{$p<0.001$}
\expandafter\def\csname res@gap.ts.agree\endcsname{0.776}
\expandafter\def\csname res@gap.ts.agreementp\endcsname{$p<0.001$}
\expandafter\def\csname res@gap.ts.cbct\endcsname{0.741}
\expandafter\def\csname res@gap.ts.ct\endcsname{0.918}
\expandafter\def\csname res@gap.ts.dice_cbctp\endcsname{$p<0.001$}
\expandafter\def\csname res@gap.ts.gap\endcsname{-0.178}
\expandafter\def\csname res@gap.ts.gapci\endcsname{[-0.186, -0.170]}
\expandafter\def\csname res@gap.ts.gapp\endcsname{$p<0.001$}
\expandafter\def\csname res@ident.cads.ribs.correct\endcsname{98\%}
\expandafter\def\csname res@ident.cads.ribs.level\endcsname{0.900}
\expandafter\def\csname res@ident.cads.ribs.merged\endcsname{0.935}
\expandafter\def\csname res@ident.cads.ribs.offone\endcsname{2\%}
\expandafter\def\csname res@ident.cads.vertebrae.correct\endcsname{99\%}
\expandafter\def\csname res@ident.cads.vertebrae.level\endcsname{0.926}
\expandafter\def\csname res@ident.cads.vertebrae.merged\endcsname{0.962}
\expandafter\def\csname res@ident.cads.vertebrae.offone\endcsname{1\%}
\expandafter\def\csname res@ident.ims25.ribs.correct\endcsname{95\%}
\expandafter\def\csname res@ident.ims25.ribs.level\endcsname{0.767}
\expandafter\def\csname res@ident.ims25.ribs.merged\endcsname{0.849}
\expandafter\def\csname res@ident.ims25.ribs.offone\endcsname{5\%}
\expandafter\def\csname res@ident.ims25.vertebrae.correct\endcsname{93\%}
\expandafter\def\csname res@ident.ims25.vertebrae.level\endcsname{0.809}
\expandafter\def\csname res@ident.ims25.vertebrae.merged\endcsname{0.913}
\expandafter\def\csname res@ident.ims25.vertebrae.offone\endcsname{7\%}
\expandafter\def\csname res@ident.ims3d.ribs.correct\endcsname{98\%}
\expandafter\def\csname res@ident.ims3d.ribs.level\endcsname{0.798}
\expandafter\def\csname res@ident.ims3d.ribs.merged\endcsname{0.850}
\expandafter\def\csname res@ident.ims3d.ribs.offone\endcsname{2\%}
\expandafter\def\csname res@ident.ims3d.vertebrae.correct\endcsname{97\%}
\expandafter\def\csname res@ident.ims3d.vertebrae.level\endcsname{0.843}
\expandafter\def\csname res@ident.ims3d.vertebrae.merged\endcsname{0.917}
\expandafter\def\csname res@ident.ims3d.vertebrae.offone\endcsname{3\%}
\expandafter\def\csname res@ident.nv.ribs.correct\endcsname{98\%}
\expandafter\def\csname res@ident.nv.ribs.level\endcsname{0.787}
\expandafter\def\csname res@ident.nv.ribs.merged\endcsname{0.827}
\expandafter\def\csname res@ident.nv.ribs.offone\endcsname{2\%}
\expandafter\def\csname res@ident.nv.vertebrae.correct\endcsname{97\%}
\expandafter\def\csname res@ident.nv.vertebrae.level\endcsname{0.826}
\expandafter\def\csname res@ident.nv.vertebrae.merged\endcsname{0.896}
\expandafter\def\csname res@ident.nv.vertebrae.offone\endcsname{3\%}
\expandafter\def\csname res@ident.ts.ribs.correct\endcsname{97\%}
\expandafter\def\csname res@ident.ts.ribs.level\endcsname{0.802}
\expandafter\def\csname res@ident.ts.ribs.merged\endcsname{0.849}
\expandafter\def\csname res@ident.ts.ribs.offone\endcsname{3\%}
\expandafter\def\csname res@ident.ts.vertebrae.correct\endcsname{98\%}
\expandafter\def\csname res@ident.ts.vertebrae.level\endcsname{0.879}
\expandafter\def\csname res@ident.ts.vertebrae.merged\endcsname{0.929}
\expandafter\def\csname res@ident.ts.vertebrae.offone\endcsname{2\%}
\expandafter\def\csname res@identrtk.cads.ribs.correct\endcsname{85\%}
\expandafter\def\csname res@identrtk.cads.ribs.level\endcsname{0.590}
\expandafter\def\csname res@identrtk.cads.ribs.merged\endcsname{0.678}
\expandafter\def\csname res@identrtk.cads.ribs.offone\endcsname{3\%}
\expandafter\def\csname res@identrtk.cads.vertebrae.correct\endcsname{92\%}
\expandafter\def\csname res@identrtk.cads.vertebrae.level\endcsname{0.788}
\expandafter\def\csname res@identrtk.cads.vertebrae.merged\endcsname{0.847}
\expandafter\def\csname res@identrtk.cads.vertebrae.offone\endcsname{5\%}
\expandafter\def\csname res@identrtk.ims25.ribs.correct\endcsname{90\%}
\expandafter\def\csname res@identrtk.ims25.ribs.level\endcsname{0.744}
\expandafter\def\csname res@identrtk.ims25.ribs.merged\endcsname{0.847}
\expandafter\def\csname res@identrtk.ims25.ribs.offone\endcsname{8\%}
\expandafter\def\csname res@identrtk.ims25.vertebrae.correct\endcsname{91\%}
\expandafter\def\csname res@identrtk.ims25.vertebrae.level\endcsname{0.796}
\expandafter\def\csname res@identrtk.ims25.vertebrae.merged\endcsname{0.880}
\expandafter\def\csname res@identrtk.ims25.vertebrae.offone\endcsname{9\%}
\expandafter\def\csname res@identrtk.ims3d.ribs.correct\endcsname{92\%}
\expandafter\def\csname res@identrtk.ims3d.ribs.level\endcsname{0.770}
\expandafter\def\csname res@identrtk.ims3d.ribs.merged\endcsname{0.846}
\expandafter\def\csname res@identrtk.ims3d.ribs.offone\endcsname{5\%}
\expandafter\def\csname res@identrtk.ims3d.vertebrae.correct\endcsname{96\%}
\expandafter\def\csname res@identrtk.ims3d.vertebrae.level\endcsname{0.852}
\expandafter\def\csname res@identrtk.ims3d.vertebrae.merged\endcsname{0.884}
\expandafter\def\csname res@identrtk.ims3d.vertebrae.offone\endcsname{4\%}
\expandafter\def\csname res@identrtk.nv.ribs.correct\endcsname{86\%}
\expandafter\def\csname res@identrtk.nv.ribs.level\endcsname{0.566}
\expandafter\def\csname res@identrtk.nv.ribs.merged\endcsname{0.664}
\expandafter\def\csname res@identrtk.nv.ribs.offone\endcsname{7\%}
\expandafter\def\csname res@identrtk.nv.vertebrae.correct\endcsname{91\%}
\expandafter\def\csname res@identrtk.nv.vertebrae.level\endcsname{0.745}
\expandafter\def\csname res@identrtk.nv.vertebrae.merged\endcsname{0.818}
\expandafter\def\csname res@identrtk.nv.vertebrae.offone\endcsname{6\%}
\expandafter\def\csname res@identrtk.ts.ribs.correct\endcsname{87\%}
\expandafter\def\csname res@identrtk.ts.ribs.level\endcsname{0.585}
\expandafter\def\csname res@identrtk.ts.ribs.merged\endcsname{0.675}
\expandafter\def\csname res@identrtk.ts.ribs.offone\endcsname{3\%}
\expandafter\def\csname res@identrtk.ts.vertebrae.correct\endcsname{93\%}
\expandafter\def\csname res@identrtk.ts.vertebrae.level\endcsname{0.817}
\expandafter\def\csname res@identrtk.ts.vertebrae.merged\endcsname{0.872}
\expandafter\def\csname res@identrtk.ts.vertebrae.offone\endcsname{5\%}
\expandafter\def\csname res@identvthreect.cads.ribs.correct\endcsname{95\%}
\expandafter\def\csname res@identvthreect.cads.ribs.level\endcsname{0.862}
\expandafter\def\csname res@identvthreect.cads.ribs.merged\endcsname{0.912}
\expandafter\def\csname res@identvthreect.cads.ribs.offone\endcsname{3\%}
\expandafter\def\csname res@identvthreect.cads.vertebrae.correct\endcsname{95\%}
\expandafter\def\csname res@identvthreect.cads.vertebrae.level\endcsname{0.912}
\expandafter\def\csname res@identvthreect.cads.vertebrae.merged\endcsname{0.963}
\expandafter\def\csname res@identvthreect.cads.vertebrae.offone\endcsname{4\%}
\expandafter\def\csname res@identvthreect.ims25.ribs.correct\endcsname{88\%}
\expandafter\def\csname res@identvthreect.ims25.ribs.level\endcsname{0.724}
\expandafter\def\csname res@identvthreect.ims25.ribs.merged\endcsname{0.871}
\expandafter\def\csname res@identvthreect.ims25.ribs.offone\endcsname{10\%}
\expandafter\def\csname res@identvthreect.ims25.vertebrae.correct\endcsname{87\%}
\expandafter\def\csname res@identvthreect.ims25.vertebrae.level\endcsname{0.774}
\expandafter\def\csname res@identvthreect.ims25.vertebrae.merged\endcsname{0.914}
\expandafter\def\csname res@identvthreect.ims25.vertebrae.offone\endcsname{11\%}
\expandafter\def\csname res@identvthreect.ims3d.ribs.correct\endcsname{92\%}
\expandafter\def\csname res@identvthreect.ims3d.ribs.level\endcsname{0.763}
\expandafter\def\csname res@identvthreect.ims3d.ribs.merged\endcsname{0.869}
\expandafter\def\csname res@identvthreect.ims3d.ribs.offone\endcsname{7\%}
\expandafter\def\csname res@identvthreect.ims3d.vertebrae.correct\endcsname{92\%}
\expandafter\def\csname res@identvthreect.ims3d.vertebrae.level\endcsname{0.838}
\expandafter\def\csname res@identvthreect.ims3d.vertebrae.merged\endcsname{0.927}
\expandafter\def\csname res@identvthreect.ims3d.vertebrae.offone\endcsname{5\%}
\expandafter\def\csname res@identvthreect.nv.ribs.correct\endcsname{90\%}
\expandafter\def\csname res@identvthreect.nv.ribs.level\endcsname{0.778}
\expandafter\def\csname res@identvthreect.nv.ribs.merged\endcsname{0.864}
\expandafter\def\csname res@identvthreect.nv.ribs.offone\endcsname{7\%}
\expandafter\def\csname res@identvthreect.nv.vertebrae.correct\endcsname{90\%}
\expandafter\def\csname res@identvthreect.nv.vertebrae.level\endcsname{0.797}
\expandafter\def\csname res@identvthreect.nv.vertebrae.merged\endcsname{0.873}
\expandafter\def\csname res@identvthreect.nv.vertebrae.offone\endcsname{6\%}
\expandafter\def\csname res@identvthreect.ts.ribs.correct\endcsname{96\%}
\expandafter\def\csname res@identvthreect.ts.ribs.level\endcsname{0.922}
\expandafter\def\csname res@identvthreect.ts.ribs.merged\endcsname{0.970}
\expandafter\def\csname res@identvthreect.ts.ribs.offone\endcsname{3\%}
\expandafter\def\csname res@identvthreect.ts.vertebrae.correct\endcsname{97\%}
\expandafter\def\csname res@identvthreect.ts.vertebrae.level\endcsname{0.944}
\expandafter\def\csname res@identvthreect.ts.vertebrae.merged\endcsname{0.976}
\expandafter\def\csname res@identvthreect.ts.vertebrae.offone\endcsname{2\%}
\expandafter\def\csname res@internal.ALL.cases\endcsname{839}
\expandafter\def\csname res@internal.ALL.dice\endcsname{0.747}
\expandafter\def\csname res@internal.ALL.labelmedian\endcsname{0.864}
\expandafter\def\csname res@internal.CBCT.cases\endcsname{277}
\expandafter\def\csname res@internal.CBCT.dice\endcsname{0.777}
\expandafter\def\csname res@internal.CBCT.labelmedian\endcsname{0.876}
\expandafter\def\csname res@internal.CT.cases\endcsname{281}
\expandafter\def\csname res@internal.CT.dice\endcsname{0.834}
\expandafter\def\csname res@internal.CT.labelmedian\endcsname{0.895}
\expandafter\def\csname res@internal.MRI.cases\endcsname{281}
\expandafter\def\csname res@internal.MRI.dice\endcsname{0.630}
\expandafter\def\csname res@internal.MRI.labelmedian\endcsname{0.760}
\expandafter\def\csname res@internal3d.ALL.cases\endcsname{839}
\expandafter\def\csname res@internal3d.ALL.dice\endcsname{0.765}
\expandafter\def\csname res@internal3d.ALL.labelmedian\endcsname{0.874}
\expandafter\def\csname res@internal3d.CBCT.cases\endcsname{277}
\expandafter\def\csname res@internal3d.CBCT.dice\endcsname{0.798}
\expandafter\def\csname res@internal3d.CBCT.labelmedian\endcsname{0.888}
\expandafter\def\csname res@internal3d.CT.cases\endcsname{281}
\expandafter\def\csname res@internal3d.CT.dice\endcsname{0.852}
\expandafter\def\csname res@internal3d.CT.labelmedian\endcsname{0.902}
\expandafter\def\csname res@internal3d.MRI.cases\endcsname{281}
\expandafter\def\csname res@internal3d.MRI.dice\endcsname{0.647}
\expandafter\def\csname res@internal3d.MRI.labelmedian\endcsname{0.770}
\expandafter\def\csname res@nvrsynthradcbct.cads.dice\endcsname{0.491}
\expandafter\def\csname res@nvrsynthradcbct.ims25.dice\endcsname{0.834}
\expandafter\def\csname res@nvrsynthradcbct.ims3d.dice\endcsname{0.849}
\expandafter\def\csname res@nvrsynthradcbct.nv.dice\endcsname{0.679}
\expandafter\def\csname res@nvrsynthradcbct.ts.dice\endcsname{0.454}
\expandafter\def\csname res@nvrsynthradmri.ims25.dice\endcsname{0.817}
\expandafter\def\csname res@nvrsynthradmri.ims3d.dice\endcsname{0.822}
\expandafter\def\csname res@nvrsynthradmri.mrseg.dice\endcsname{0.803}
\expandafter\def\csname res@nvrsynthradmri.nv.dice\endcsname{0.667}
\expandafter\def\csname res@nvrsynthradmri.ts.dice\endcsname{0.820}
\expandafter\def\csname res@nvrtsrtk.cads.dice\endcsname{0.746}
\expandafter\def\csname res@nvrtsrtk.ims25.dice\endcsname{0.868}
\expandafter\def\csname res@nvrtsrtk.ims3d.dice\endcsname{0.874}
\expandafter\def\csname res@nvrtsrtk.nv.dice\endcsname{0.745}
\expandafter\def\csname res@nvrtsrtk.ts.dice\endcsname{0.747}
\expandafter\def\csname res@nvrtsvthreect.cads.dice\endcsname{0.881}
\expandafter\def\csname res@nvrtsvthreect.ims25.dice\endcsname{0.840}
\expandafter\def\csname res@nvrtsvthreect.ims25ens.dice\endcsname{0.852}
\expandafter\def\csname res@nvrtsvthreect.ims3d.dice\endcsname{0.856}
\expandafter\def\csname res@nvrtsvthreect.ims3dens.dice\endcsname{0.867}
\expandafter\def\csname res@nvrtsvthreect.nv.dice\endcsname{0.808}
\expandafter\def\csname res@nvrtsvthreect.ts.dice\endcsname{0.922}
\expandafter\def\csname res@nvrtsvthreemri.ims25.dice\endcsname{0.758}
\expandafter\def\csname res@nvrtsvthreemri.ims25ens.dice\endcsname{0.780}
\expandafter\def\csname res@nvrtsvthreemri.ims3d.dice\endcsname{0.778}
\expandafter\def\csname res@nvrtsvthreemri.ims3dens.dice\endcsname{0.795}
\expandafter\def\csname res@nvrtsvthreemri.mrseg.dice\endcsname{0.812}
\expandafter\def\csname res@nvrtsvthreemri.nv.dice\endcsname{0.718}
\expandafter\def\csname res@nvrtsvthreemri.ts.dice\endcsname{0.843}
\expandafter\def\csname res@nvrvisceral.ims25.dice\endcsname{0.866}
\expandafter\def\csname res@nvrvisceral.ims3d.dice\endcsname{0.868}
\expandafter\def\csname res@nvrvisceral.nv.dice\endcsname{0.784}
\expandafter\def\csname res@nvrvisceral.ts.dice\endcsname{0.857}
\expandafter\def\csname res@synthradcbct.cads.assd\endcsname{6.00}
\expandafter\def\csname res@synthradcbct.cads.ci\endcsname{[0.454, 0.524]}
\expandafter\def\csname res@synthradcbct.cads.dice\endcsname{0.489}
\expandafter\def\csname res@synthradcbct.cads.dicesd\endcsname{0.208}
\expandafter\def\csname res@synthradcbct.cads.diff\endcsname{+0.329}
\expandafter\def\csname res@synthradcbct.cads.hd95\endcsname{19.2}
\expandafter\def\csname res@synthradcbct.cads.nsd\endcsname{0.362}
\expandafter\def\csname res@synthradcbct.cads.p\endcsname{$p<0.001$}
\expandafter\def\csname res@synthradcbct.cads.pthreed\endcsname{$p<0.001$}
\expandafter\def\csname res@synthradcbct.cases\endcsname{139}
\expandafter\def\csname res@synthradcbct.ims25.assd\endcsname{1.93}
\expandafter\def\csname res@synthradcbct.ims25.ci\endcsname{[0.804, 0.832]}
\expandafter\def\csname res@synthradcbct.ims25.dice\endcsname{0.819}
\expandafter\def\csname res@synthradcbct.ims25.dicesd\endcsname{0.086}
\expandafter\def\csname res@synthradcbct.ims25.hd95\endcsname{6.8}
\expandafter\def\csname res@synthradcbct.ims25.nsd\endcsname{0.667}
\expandafter\def\csname res@synthradcbct.ims25.pthreed\endcsname{$p<0.001$}
\expandafter\def\csname res@synthradcbct.ims3d.assd\endcsname{1.57}
\expandafter\def\csname res@synthradcbct.ims3d.ci\endcsname{[0.831, 0.852]}
\expandafter\def\csname res@synthradcbct.ims3d.dice\endcsname{0.842}
\expandafter\def\csname res@synthradcbct.ims3d.dicesd\endcsname{0.063}
\expandafter\def\csname res@synthradcbct.ims3d.diff\endcsname{-0.014}
\expandafter\def\csname res@synthradcbct.ims3d.hd95\endcsname{5.6}
\expandafter\def\csname res@synthradcbct.ims3d.nsd\endcsname{0.703}
\expandafter\def\csname res@synthradcbct.ims3d.p\endcsname{$p<0.001$}
\expandafter\def\csname res@synthradcbct.labels\endcsname{89}
\expandafter\def\csname res@synthradcbct.nv.assd\endcsname{10.00}
\expandafter\def\csname res@synthradcbct.nv.ci\endcsname{[0.644, 0.697]}
\expandafter\def\csname res@synthradcbct.nv.dice\endcsname{0.671}
\expandafter\def\csname res@synthradcbct.nv.dicesd\endcsname{0.157}
\expandafter\def\csname res@synthradcbct.nv.diff\endcsname{+0.087}
\expandafter\def\csname res@synthradcbct.nv.hd95\endcsname{30.8}
\expandafter\def\csname res@synthradcbct.nv.nsd\endcsname{0.475}
\expandafter\def\csname res@synthradcbct.nv.p\endcsname{$p<0.001$}
\expandafter\def\csname res@synthradcbct.nv.pthreed\endcsname{$p<0.001$}
\expandafter\def\csname res@synthradcbct.ts.assd\endcsname{12.58}
\expandafter\def\csname res@synthradcbct.ts.ci\endcsname{[0.410, 0.496]}
\expandafter\def\csname res@synthradcbct.ts.dice\endcsname{0.454}
\expandafter\def\csname res@synthradcbct.ts.dicesd\endcsname{0.259}
\expandafter\def\csname res@synthradcbct.ts.diff\endcsname{+0.334}
\expandafter\def\csname res@synthradcbct.ts.hd95\endcsname{31.4}
\expandafter\def\csname res@synthradcbct.ts.nsd\endcsname{0.347}
\expandafter\def\csname res@synthradcbct.ts.p\endcsname{$p<0.001$}
\expandafter\def\csname res@synthradcbct.ts.pthreed\endcsname{$p<0.001$}
\expandafter\def\csname res@synthradmri.cases\endcsname{52}
\expandafter\def\csname res@synthradmri.ims25.assd\endcsname{1.98}
\expandafter\def\csname res@synthradmri.ims25.ci\endcsname{[0.803, 0.831]}
\expandafter\def\csname res@synthradmri.ims25.dice\endcsname{0.817}
\expandafter\def\csname res@synthradmri.ims25.dicesd\endcsname{0.052}
\expandafter\def\csname res@synthradmri.ims25.hd95\endcsname{7.3}
\expandafter\def\csname res@synthradmri.ims25.nsd\endcsname{0.619}
\expandafter\def\csname res@synthradmri.ims25.pthreed\endcsname{$p=0.14$}
\expandafter\def\csname res@synthradmri.ims3d.assd\endcsname{1.93}
\expandafter\def\csname res@synthradmri.ims3d.ci\endcsname{[0.807, 0.837]}
\expandafter\def\csname res@synthradmri.ims3d.dice\endcsname{0.822}
\expandafter\def\csname res@synthradmri.ims3d.dicesd\endcsname{0.056}
\expandafter\def\csname res@synthradmri.ims3d.diff\endcsname{-0.007}
\expandafter\def\csname res@synthradmri.ims3d.hd95\endcsname{6.8}
\expandafter\def\csname res@synthradmri.ims3d.nsd\endcsname{0.620}
\expandafter\def\csname res@synthradmri.ims3d.p\endcsname{$p=0.16$}
\expandafter\def\csname res@synthradmri.labels\endcsname{39}
\expandafter\def\csname res@synthradmri.mrseg.assd\endcsname{2.75}
\expandafter\def\csname res@synthradmri.mrseg.ci\endcsname{[0.791, 0.815]}
\expandafter\def\csname res@synthradmri.mrseg.dice\endcsname{0.803}
\expandafter\def\csname res@synthradmri.mrseg.dicesd\endcsname{0.045}
\expandafter\def\csname res@synthradmri.mrseg.diff\endcsname{+0.015}
\expandafter\def\csname res@synthradmri.mrseg.hd95\endcsname{11.0}
\expandafter\def\csname res@synthradmri.mrseg.nsd\endcsname{0.550}
\expandafter\def\csname res@synthradmri.mrseg.p\endcsname{$p=0.24$}
\expandafter\def\csname res@synthradmri.mrseg.pthreed\endcsname{$p=0.009$}
\expandafter\def\csname res@synthradmri.nv.assd\endcsname{10.97}
\expandafter\def\csname res@synthradmri.nv.ci\endcsname{[0.624, 0.704]}
\expandafter\def\csname res@synthradmri.nv.dice\endcsname{0.667}
\expandafter\def\csname res@synthradmri.nv.dicesd\endcsname{0.147}
\expandafter\def\csname res@synthradmri.nv.diff\endcsname{+0.097}
\expandafter\def\csname res@synthradmri.nv.hd95\endcsname{37.6}
\expandafter\def\csname res@synthradmri.nv.nsd\endcsname{0.432}
\expandafter\def\csname res@synthradmri.nv.p\endcsname{$p<0.001$}
\expandafter\def\csname res@synthradmri.nv.pthreed\endcsname{$p<0.001$}
\expandafter\def\csname res@synthradmri.ts.assd\endcsname{2.51}
\expandafter\def\csname res@synthradmri.ts.ci\endcsname{[0.792, 0.848]}
\expandafter\def\csname res@synthradmri.ts.dice\endcsname{0.820}
\expandafter\def\csname res@synthradmri.ts.dicesd\endcsname{0.101}
\expandafter\def\csname res@synthradmri.ts.diff\endcsname{-0.037}
\expandafter\def\csname res@synthradmri.ts.hd95\endcsname{10.0}
\expandafter\def\csname res@synthradmri.ts.nsd\endcsname{0.635}
\expandafter\def\csname res@synthradmri.ts.p\endcsname{$p=1.00$}
\expandafter\def\csname res@synthradmri.ts.pthreed\endcsname{$p=1.00$}
\expandafter\def\csname res@tsmr.cases\endcsname{120}
\expandafter\def\csname res@tsmr.ims25.assd\endcsname{3.67}
\expandafter\def\csname res@tsmr.ims25.ci\endcsname{[0.711, 0.751]}
\expandafter\def\csname res@tsmr.ims25.dice\endcsname{0.732}
\expandafter\def\csname res@tsmr.ims25.dicesd\endcsname{0.115}
\expandafter\def\csname res@tsmr.ims25.hd95\endcsname{13.8}
\expandafter\def\csname res@tsmr.ims25.nsd\endcsname{0.453}
\expandafter\def\csname res@tsmr.ims25.pthreed\endcsname{$p<0.001$}
\expandafter\def\csname res@tsmr.ims3d.assd\endcsname{2.85}
\expandafter\def\csname res@tsmr.ims3d.ci\endcsname{[0.753, 0.783]}
\expandafter\def\csname res@tsmr.ims3d.dice\endcsname{0.768}
\expandafter\def\csname res@tsmr.ims3d.dicesd\endcsname{0.085}
\expandafter\def\csname res@tsmr.ims3d.diff\endcsname{-0.016}
\expandafter\def\csname res@tsmr.ims3d.hd95\endcsname{10.6}
\expandafter\def\csname res@tsmr.ims3d.nsd\endcsname{0.487}
\expandafter\def\csname res@tsmr.ims3d.p\endcsname{$p<0.001$}
\expandafter\def\csname res@tsmr.labels\endcsname{39}
\expandafter\def\csname res@tsmr.mrseg.assd\endcsname{4.32}
\expandafter\def\csname res@tsmr.mrseg.ci\endcsname{[0.696, 0.760]}
\expandafter\def\csname res@tsmr.mrseg.dice\endcsname{0.730}
\expandafter\def\csname res@tsmr.mrseg.dicesd\endcsname{0.181}
\expandafter\def\csname res@tsmr.mrseg.diff\endcsname{-0.018}
\expandafter\def\csname res@tsmr.mrseg.hd95\endcsname{15.0}
\expandafter\def\csname res@tsmr.mrseg.nsd\endcsname{0.455}
\expandafter\def\csname res@tsmr.mrseg.p\endcsname{$p=0.20$}
\expandafter\def\csname res@tsmr.mrseg.pthreed\endcsname{$p=1.00$}
\expandafter\def\csname res@tsmr.nv.assd\endcsname{8.48}
\expandafter\def\csname res@tsmr.nv.ci\endcsname{[0.657, 0.734]}
\expandafter\def\csname res@tsmr.nv.dice\endcsname{0.696}
\expandafter\def\csname res@tsmr.nv.dicesd\endcsname{0.216}
\expandafter\def\csname res@tsmr.nv.diff\endcsname{-0.022}
\expandafter\def\csname res@tsmr.nv.hd95\endcsname{22.8}
\expandafter\def\csname res@tsmr.nv.nsd\endcsname{0.445}
\expandafter\def\csname res@tsmr.nv.p\endcsname{$p=1.00$}
\expandafter\def\csname res@tsmr.nv.pthreed\endcsname{$p=1.00$}
\expandafter\def\csname res@tsrtk.cads.assd\endcsname{2.98}
\expandafter\def\csname res@tsrtk.cads.ci\endcsname{[0.731, 0.750]}
\expandafter\def\csname res@tsrtk.cads.dice\endcsname{0.740}
\expandafter\def\csname res@tsrtk.cads.dicesd\endcsname{0.108}
\expandafter\def\csname res@tsrtk.cads.diff\endcsname{+0.078}
\expandafter\def\csname res@tsrtk.cads.hd95\endcsname{10.8}
\expandafter\def\csname res@tsrtk.cads.nsd\endcsname{0.432}
\expandafter\def\csname res@tsrtk.cads.p\endcsname{$p<0.001$}
\expandafter\def\csname res@tsrtk.cads.pthreed\endcsname{$p<0.001$}
\expandafter\def\csname res@tsrtk.cases\endcsname{497}
\expandafter\def\csname res@tsrtk.ims25.assd\endcsname{1.77}
\expandafter\def\csname res@tsrtk.ims25.ci\endcsname{[0.826, 0.840]}
\expandafter\def\csname res@tsrtk.ims25.dice\endcsname{0.833}
\expandafter\def\csname res@tsrtk.ims25.dicesd\endcsname{0.079}
\expandafter\def\csname res@tsrtk.ims25.hd95\endcsname{6.1}
\expandafter\def\csname res@tsrtk.ims25.nsd\endcsname{0.524}
\expandafter\def\csname res@tsrtk.ims25.pthreed\endcsname{$p<0.001$}
\expandafter\def\csname res@tsrtk.ims3d.assd\endcsname{1.42}
\expandafter\def\csname res@tsrtk.ims3d.ci\endcsname{[0.849, 0.860]}
\expandafter\def\csname res@tsrtk.ims3d.dice\endcsname{0.855}
\expandafter\def\csname res@tsrtk.ims3d.dicesd\endcsname{0.066}
\expandafter\def\csname res@tsrtk.ims3d.diff\endcsname{-0.009}
\expandafter\def\csname res@tsrtk.ims3d.hd95\endcsname{4.6}
\expandafter\def\csname res@tsrtk.ims3d.nsd\endcsname{0.531}
\expandafter\def\csname res@tsrtk.ims3d.p\endcsname{$p<0.001$}
\expandafter\def\csname res@tsrtk.labels\endcsname{93}
\expandafter\def\csname res@tsrtk.nv.assd\endcsname{4.10}
\expandafter\def\csname res@tsrtk.nv.ci\endcsname{[0.697, 0.721]}
\expandafter\def\csname res@tsrtk.nv.dice\endcsname{0.710}
\expandafter\def\csname res@tsrtk.nv.dicesd\endcsname{0.135}
\expandafter\def\csname res@tsrtk.nv.diff\endcsname{+0.106}
\expandafter\def\csname res@tsrtk.nv.hd95\endcsname{14.3}
\expandafter\def\csname res@tsrtk.nv.nsd\endcsname{0.361}
\expandafter\def\csname res@tsrtk.nv.p\endcsname{$p<0.001$}
\expandafter\def\csname res@tsrtk.nv.pthreed\endcsname{$p<0.001$}
\expandafter\def\csname res@tsrtk.ts.assd\endcsname{3.32}
\expandafter\def\csname res@tsrtk.ts.ci\endcsname{[0.732, 0.748]}
\expandafter\def\csname res@tsrtk.ts.dice\endcsname{0.741}
\expandafter\def\csname res@tsrtk.ts.dicesd\endcsname{0.094}
\expandafter\def\csname res@tsrtk.ts.diff\endcsname{+0.080}
\expandafter\def\csname res@tsrtk.ts.hd95\endcsname{11.7}
\expandafter\def\csname res@tsrtk.ts.nsd\endcsname{0.451}
\expandafter\def\csname res@tsrtk.ts.p\endcsname{$p<0.001$}
\expandafter\def\csname res@tsrtk.ts.pthreed\endcsname{$p<0.001$}
\expandafter\def\csname res@tsvthreect.cads.assd\endcsname{1.15}
\expandafter\def\csname res@tsvthreect.cads.ci\endcsname{[0.878, 0.900]}
\expandafter\def\csname res@tsvthreect.cads.dice\endcsname{0.890}
\expandafter\def\csname res@tsvthreect.cads.dicesd\endcsname{0.091}
\expandafter\def\csname res@tsvthreect.cads.diff\endcsname{-0.067}
\expandafter\def\csname res@tsvthreect.cads.hd95\endcsname{4.7}
\expandafter\def\csname res@tsvthreect.cads.nsd\endcsname{0.696}
\expandafter\def\csname res@tsvthreect.cads.p\endcsname{$p<0.001$}
\expandafter\def\csname res@tsvthreect.cads.pthreed\endcsname{$p<0.001$}
\expandafter\def\csname res@tsvthreect.cases\endcsname{247}
\expandafter\def\csname res@tsvthreect.ims25.assd\endcsname{3.10}
\expandafter\def\csname res@tsvthreect.ims25.ci\endcsname{[0.783, 0.814]}
\expandafter\def\csname res@tsvthreect.ims25.dice\endcsname{0.799}
\expandafter\def\csname res@tsvthreect.ims25.dicesd\endcsname{0.128}
\expandafter\def\csname res@tsvthreect.ims25.hd95\endcsname{11.4}
\expandafter\def\csname res@tsvthreect.ims25.nsd\endcsname{0.527}
\expandafter\def\csname res@tsvthreect.ims25.pthreed\endcsname{$p<0.001$}
\expandafter\def\csname res@tsvthreect.ims25ens.assd\endcsname{2.40}
\expandafter\def\csname res@tsvthreect.ims25ens.ci\endcsname{[0.799, 0.831]}
\expandafter\def\csname res@tsvthreect.ims25ens.dice\endcsname{0.816}
\expandafter\def\csname res@tsvthreect.ims25ens.dicesd\endcsname{0.130}
\expandafter\def\csname res@tsvthreect.ims25ens.diff\endcsname{-0.014}
\expandafter\def\csname res@tsvthreect.ims25ens.hd95\endcsname{8.6}
\expandafter\def\csname res@tsvthreect.ims25ens.nsd\endcsname{0.558}
\expandafter\def\csname res@tsvthreect.ims25ens.p\endcsname{$p<0.001$}
\expandafter\def\csname res@tsvthreect.ims25ens.pthreed\endcsname{$p<0.001$}
\expandafter\def\csname res@tsvthreect.ims3d.assd\endcsname{2.30}
\expandafter\def\csname res@tsvthreect.ims3d.ci\endcsname{[0.815, 0.845]}
\expandafter\def\csname res@tsvthreect.ims3d.dice\endcsname{0.831}
\expandafter\def\csname res@tsvthreect.ims3d.dicesd\endcsname{0.122}
\expandafter\def\csname res@tsvthreect.ims3d.diff\endcsname{-0.019}
\expandafter\def\csname res@tsvthreect.ims3d.hd95\endcsname{7.8}
\expandafter\def\csname res@tsvthreect.ims3d.nsd\endcsname{0.549}
\expandafter\def\csname res@tsvthreect.ims3d.p\endcsname{$p<0.001$}
\expandafter\def\csname res@tsvthreect.ims3dens.assd\endcsname{1.77}
\expandafter\def\csname res@tsvthreect.ims3dens.ci\endcsname{[0.831, 0.860]}
\expandafter\def\csname res@tsvthreect.ims3dens.dice\endcsname{0.846}
\expandafter\def\csname res@tsvthreect.ims3dens.dicesd\endcsname{0.119}
\expandafter\def\csname res@tsvthreect.ims3dens.diff\endcsname{-0.031}
\expandafter\def\csname res@tsvthreect.ims3dens.hd95\endcsname{5.9}
\expandafter\def\csname res@tsvthreect.ims3dens.nsd\endcsname{0.578}
\expandafter\def\csname res@tsvthreect.ims3dens.p\endcsname{$p<0.001$}
\expandafter\def\csname res@tsvthreect.ims3dens.pthreed\endcsname{$p<0.001$}
\expandafter\def\csname res@tsvthreect.labels\endcsname{93}
\expandafter\def\csname res@tsvthreect.nv.assd\endcsname{8.68}
\expandafter\def\csname res@tsvthreect.nv.ci\endcsname{[0.776, 0.818]}
\expandafter\def\csname res@tsvthreect.nv.dice\endcsname{0.798}
\expandafter\def\csname res@tsvthreect.nv.dicesd\endcsname{0.171}
\expandafter\def\csname res@tsvthreect.nv.diff\endcsname{-0.025}
\expandafter\def\csname res@tsvthreect.nv.hd95\endcsname{32.9}
\expandafter\def\csname res@tsvthreect.nv.nsd\endcsname{0.533}
\expandafter\def\csname res@tsvthreect.nv.p\endcsname{$p<0.001$}
\expandafter\def\csname res@tsvthreect.nv.pthreed\endcsname{$p=0.006$}
\expandafter\def\csname res@tsvthreect.ts.assd\endcsname{0.71}
\expandafter\def\csname res@tsvthreect.ts.ci\endcsname{[0.925, 0.939]}
\expandafter\def\csname res@tsvthreect.ts.dice\endcsname{0.932}
\expandafter\def\csname res@tsvthreect.ts.dicesd\endcsname{0.056}
\expandafter\def\csname res@tsvthreect.ts.diff\endcsname{-0.103}
\expandafter\def\csname res@tsvthreect.ts.hd95\endcsname{2.8}
\expandafter\def\csname res@tsvthreect.ts.nsd\endcsname{0.775}
\expandafter\def\csname res@tsvthreect.ts.p\endcsname{$p<0.001$}
\expandafter\def\csname res@tsvthreect.ts.pthreed\endcsname{$p<0.001$}
\expandafter\def\csname res@tsvthreemri.cases\endcsname{253}
\expandafter\def\csname res@tsvthreemri.ims25.assd\endcsname{4.13}
\expandafter\def\csname res@tsvthreemri.ims25.ci\endcsname{[0.744, 0.771]}
\expandafter\def\csname res@tsvthreemri.ims25.dice\endcsname{0.758}
\expandafter\def\csname res@tsvthreemri.ims25.dicesd\endcsname{0.109}
\expandafter\def\csname res@tsvthreemri.ims25.hd95\endcsname{15.0}
\expandafter\def\csname res@tsvthreemri.ims25.nsd\endcsname{0.364}
\expandafter\def\csname res@tsvthreemri.ims25.pthreed\endcsname{$p<0.001$}
\expandafter\def\csname res@tsvthreemri.ims25ens.assd\endcsname{3.20}
\expandafter\def\csname res@tsvthreemri.ims25ens.ci\endcsname{[0.766, 0.793]}
\expandafter\def\csname res@tsvthreemri.ims25ens.dice\endcsname{0.780}
\expandafter\def\csname res@tsvthreemri.ims25ens.dicesd\endcsname{0.109}
\expandafter\def\csname res@tsvthreemri.ims25ens.diff\endcsname{-0.021}
\expandafter\def\csname res@tsvthreemri.ims25ens.hd95\endcsname{11.7}
\expandafter\def\csname res@tsvthreemri.ims25ens.nsd\endcsname{0.395}
\expandafter\def\csname res@tsvthreemri.ims25ens.p\endcsname{$p<0.001$}
\expandafter\def\csname res@tsvthreemri.ims25ens.pthreed\endcsname{$p=0.16$}
\expandafter\def\csname res@tsvthreemri.ims3d.assd\endcsname{3.30}
\expandafter\def\csname res@tsvthreemri.ims3d.ci\endcsname{[0.765, 0.791]}
\expandafter\def\csname res@tsvthreemri.ims3d.dice\endcsname{0.778}
\expandafter\def\csname res@tsvthreemri.ims3d.dicesd\endcsname{0.102}
\expandafter\def\csname res@tsvthreemri.ims3d.diff\endcsname{-0.018}
\expandafter\def\csname res@tsvthreemri.ims3d.hd95\endcsname{11.7}
\expandafter\def\csname res@tsvthreemri.ims3d.nsd\endcsname{0.381}
\expandafter\def\csname res@tsvthreemri.ims3d.p\endcsname{$p<0.001$}
\expandafter\def\csname res@tsvthreemri.ims3dens.assd\endcsname{2.85}
\expandafter\def\csname res@tsvthreemri.ims3dens.ci\endcsname{[0.781, 0.807]}
\expandafter\def\csname res@tsvthreemri.ims3dens.dice\endcsname{0.795}
\expandafter\def\csname res@tsvthreemri.ims3dens.dicesd\endcsname{0.108}
\expandafter\def\csname res@tsvthreemri.ims3dens.diff\endcsname{-0.036}
\expandafter\def\csname res@tsvthreemri.ims3dens.hd95\endcsname{10.0}
\expandafter\def\csname res@tsvthreemri.ims3dens.nsd\endcsname{0.408}
\expandafter\def\csname res@tsvthreemri.ims3dens.p\endcsname{$p<0.001$}
\expandafter\def\csname res@tsvthreemri.ims3dens.pthreed\endcsname{$p<0.001$}
\expandafter\def\csname res@tsvthreemri.labels\endcsname{39}
\expandafter\def\csname res@tsvthreemri.mrseg.assd\endcsname{2.51}
\expandafter\def\csname res@tsvthreemri.mrseg.ci\endcsname{[0.799, 0.823]}
\expandafter\def\csname res@tsvthreemri.mrseg.dice\endcsname{0.812}
\expandafter\def\csname res@tsvthreemri.mrseg.dicesd\endcsname{0.098}
\expandafter\def\csname res@tsvthreemri.mrseg.diff\endcsname{-0.042}
\expandafter\def\csname res@tsvthreemri.mrseg.hd95\endcsname{9.9}
\expandafter\def\csname res@tsvthreemri.mrseg.nsd\endcsname{0.438}
\expandafter\def\csname res@tsvthreemri.mrseg.p\endcsname{$p<0.001$}
\expandafter\def\csname res@tsvthreemri.mrseg.pthreed\endcsname{$p<0.001$}
\expandafter\def\csname res@tsvthreemri.nv.assd\endcsname{12.04}
\expandafter\def\csname res@tsvthreemri.nv.ci\endcsname{[0.699, 0.737]}
\expandafter\def\csname res@tsvthreemri.nv.dice\endcsname{0.718}
\expandafter\def\csname res@tsvthreemri.nv.dicesd\endcsname{0.155}
\expandafter\def\csname res@tsvthreemri.nv.diff\endcsname{+0.007}
\expandafter\def\csname res@tsvthreemri.nv.hd95\endcsname{45.4}
\expandafter\def\csname res@tsvthreemri.nv.nsd\endcsname{0.363}
\expandafter\def\csname res@tsvthreemri.nv.p\endcsname{$p<0.001$}
\expandafter\def\csname res@tsvthreemri.nv.pthreed\endcsname{$p<0.001$}
\expandafter\def\csname res@tsvthreemri.ts.assd\endcsname{2.24}
\expandafter\def\csname res@tsvthreemri.ts.ci\endcsname{[0.832, 0.853]}
\expandafter\def\csname res@tsvthreemri.ts.dice\endcsname{0.843}
\expandafter\def\csname res@tsvthreemri.ts.dicesd\endcsname{0.087}
\expandafter\def\csname res@tsvthreemri.ts.diff\endcsname{-0.076}
\expandafter\def\csname res@tsvthreemri.ts.hd95\endcsname{8.8}
\expandafter\def\csname res@tsvthreemri.ts.nsd\endcsname{0.534}
\expandafter\def\csname res@tsvthreemri.ts.p\endcsname{$p<0.001$}
\expandafter\def\csname res@tsvthreemri.ts.pthreed\endcsname{$p<0.001$}
\expandafter\def\csname res@visceral.cases\endcsname{58}
\expandafter\def\csname res@visceral.ims25.assd\endcsname{1.88}
\expandafter\def\csname res@visceral.ims25.ci\endcsname{[0.822, 0.844]}
\expandafter\def\csname res@visceral.ims25.dice\endcsname{0.833}
\expandafter\def\csname res@visceral.ims25.dicesd\endcsname{0.043}
\expandafter\def\csname res@visceral.ims25.hd95\endcsname{7.4}
\expandafter\def\csname res@visceral.ims25.nsd\endcsname{0.711}
\expandafter\def\csname res@visceral.ims25.pthreed\endcsname{$p<0.001$}
\expandafter\def\csname res@visceral.ims3d.assd\endcsname{1.52}
\expandafter\def\csname res@visceral.ims3d.ci\endcsname{[0.841, 0.859]}
\expandafter\def\csname res@visceral.ims3d.dice\endcsname{0.850}
\expandafter\def\csname res@visceral.ims3d.dicesd\endcsname{0.035}
\expandafter\def\csname res@visceral.ims3d.diff\endcsname{-0.005}
\expandafter\def\csname res@visceral.ims3d.hd95\endcsname{6.1}
\expandafter\def\csname res@visceral.ims3d.nsd\endcsname{0.730}
\expandafter\def\csname res@visceral.ims3d.p\endcsname{$p<0.001$}
\expandafter\def\csname res@visceral.labels\endcsname{103}
\expandafter\def\csname res@visceral.nv.assd\endcsname{12.03}
\expandafter\def\csname res@visceral.nv.ci\endcsname{[0.785, 0.811]}
\expandafter\def\csname res@visceral.nv.dice\endcsname{0.798}
\expandafter\def\csname res@visceral.nv.dicesd\endcsname{0.053}
\expandafter\def\csname res@visceral.nv.diff\endcsname{+0.032}
\expandafter\def\csname res@visceral.nv.hd95\endcsname{41.7}
\expandafter\def\csname res@visceral.nv.nsd\endcsname{0.687}
\expandafter\def\csname res@visceral.nv.p\endcsname{$p<0.001$}
\expandafter\def\csname res@visceral.nv.pthreed\endcsname{$p<0.001$}
\expandafter\def\csname res@visceral.ts.assd\endcsname{1.62}
\expandafter\def\csname res@visceral.ts.ci\endcsname{[0.844, 0.862]}
\expandafter\def\csname res@visceral.ts.dice\endcsname{0.853}
\expandafter\def\csname res@visceral.ts.dicesd\endcsname{0.037}
\expandafter\def\csname res@visceral.ts.diff\endcsname{-0.014}
\expandafter\def\csname res@visceral.ts.hd95\endcsname{7.0}
\expandafter\def\csname res@visceral.ts.nsd\endcsname{0.776}
\expandafter\def\csname res@visceral.ts.p\endcsname{$p<0.001$}
\expandafter\def\csname res@visceral.ts.pthreed\endcsname{$p=0.24$}

\def\stripminus#1#2\relax{\ifx-#1#2\else#1#2\fi}
\newcommand{\resabs}[1]{\expandafter\expandafter\expandafter\stripminus\csname res@#1\endcsname\relax}

\extendedsupplementtrue
\renewcommand{\shorttitle}{IMPACT-Seg}
\renewcommand{\headeright}{Preprint}
\renewcommand{\undertitle}{Preprint}
\title{IMPACT-Seg: Consistent Anatomical Segmentation and Representation Learning Across CT, CBCT, and MRI}
\author{\begin{minipage}{0.95\textwidth}\centering
C{\'e}dric H{\'e}mon$^{1,3*}$\quad Valentin Boussot$^{4}$\quad Matthew McCormick$^{4}$\\[4pt]
Lukas Zimmermann$^{5}$\quad Christopher Kurz$^{1}$\quad Guillaume Landry$^{1,2,3}$\\[8pt]
\normalfont\small $^1$ Department of Radiation Oncology, LMU University Hospital, LMU Medizin, LMU Munich, Germany\\
$^2$ German Cancer Consortium (DKTK), partner site Munich, a partnership between DKFZ and LMU University Hospital Munich, Germany \\
$^3$ Bavarian Cancer Research Center (BZKF), Munich, Germany \\
$^4$ Fideus Labs, Durham, NC 27713, United States\\
$^5$ Department of Radiation Oncology, Medical University of Vienna, Vienna, Austria\\[4pt]
$^*$Corresponding author: \texttt{Cedric.Hemon@med.uni-muenchen.de}
\end{minipage}}
\date{}
\begin{document}
\maketitle
\begin{abstract}
Computed tomography (CT), cone-beam CT (CBCT), and magnetic resonance imaging (MRI) provide complementary representations of the same underlying anatomy, yet substantial differences in image appearance challenge consistent segmentation and deformable registration in adaptive radiotherapy. We investigate whether dense anatomical segmentation supervision can promote cross-modality correspondence without explicit feature alignment. We introduce IMPACT-Seg, which segments 126 structures across CT, CBCT, and MRI using shared network parameters. A collection of 8,290 volumes was assembled from public cohorts through label harmonization, model-assisted pseudo-label completion, iterative refinement, and 2,864 physics-based CBCT simulations with voxel-aligned CT labels. Unpaired images were used to jointly train 2.5D and 3D residual-encoder networks.

On 139 real CBCTs, IMPACT-Seg 3D achieved a mean Dice of 0.842 versus 0.671 for the strongest comparator. On 497 voxel-aligned CT–simulated-CBCT pairs, the CT-to-CBCT Dice decrease was 0.022 versus 0.153–0.178 for competing methods. On independent pancreatic radiotherapy CBCTs, gastrointestinal Dice was 0.735 versus 0.634 for TotalSegmentator. Prediction agreement reached 0.881 on registered CT–CBCT and 0.790 on CT–MRI pairs. Adding MRI training data reduced CT–MRI encoder feature retrieval error from 36.8 to 3.2 mm. Registration pipelines incorporating IMPACT-Seg features ranked first on two multimodal Learn2Reg benchmarks. The 3D model segmented a whole-body CT in 15 s.

These findings show that dense anatomical segmentation supervision can support consistent anatomical representations across modalities, with applications to both segmentation and registration. Code, datasets, and pretrained models are available at \url{https://github.com/HEMONCedric/IMPACTSeg}.
\end{abstract}
\section{Introduction} \label{sec:intro}

Adaptive radiotherapy relies on repeated imaging to account for anatomical changes during treatment. Planning computed tomography (CT), on-treatment cone-beam CT (CBCT), and magnetic resonance imaging (MRI) provide different representations of the same underlying anatomy, despite substantial differences in image contrast, quality, and artefacts. Although anatomical structures may change in shape and position between acquisitions, their identity and delineation conventions should remain consistent across modalities. Accurate and consistent delineation is essential for contour adaptation, dose evaluation, and longitudinal assessment. As manual delineation of numerous structures is impractical within the time constraints of online adaptation, automated segmentation and registration-based contour propagation offer valuable alternatives~\citep{liu2023cbctadaptive,rong2024nrgsegmentation}. Both approaches benefit from consistent anatomical identification despite differences in image appearance.

Existing segmentation tools provide broad anatomical coverage on CT, including TotalSegmentator \citep{wasserthal2023totalsegmentator}, CADS \citep{xu2025cads}, and VISTA3D \citep{he2024vista3d}. Joint CT-MRI segmentation is supported by MRSegmentator \citep{hantze2025mrsegmentator} and NV-Segment-CTMR \citep{nvidia2025nvsegmentctmr}, while TotalSegmentator MRI \citep{dantonoli2025totalsegmentatormri} provides a dedicated MRI model. Differences in training distributions and label conventions nevertheless complicate their combined use in radiotherapy. CBCT adds scatter, noise, truncation, and motion artefacts, while dense annotations remain scarce. High accuracy on diagnostic CT or MRI therefore does not establish cross-modality consistency in the same patient.

This raises a central question: can a single model trained with harmonized anatomical supervision produce accurate and consistent segmentations across modalities, without explicitly enforcing feature alignment? To investigate this, we combine harmonized anatomical labels, joint multimodal training with shared network parameters, and physics-based CBCT simulation. Beyond segmentation accuracy and cross-modality prediction agreement, we examine whether this supervision promotes spatially corresponding anatomical representations and assess their utility for deformable image registration.

\begin{figure*}[!tbp]
\centering
\includegraphics[width=\textwidth]{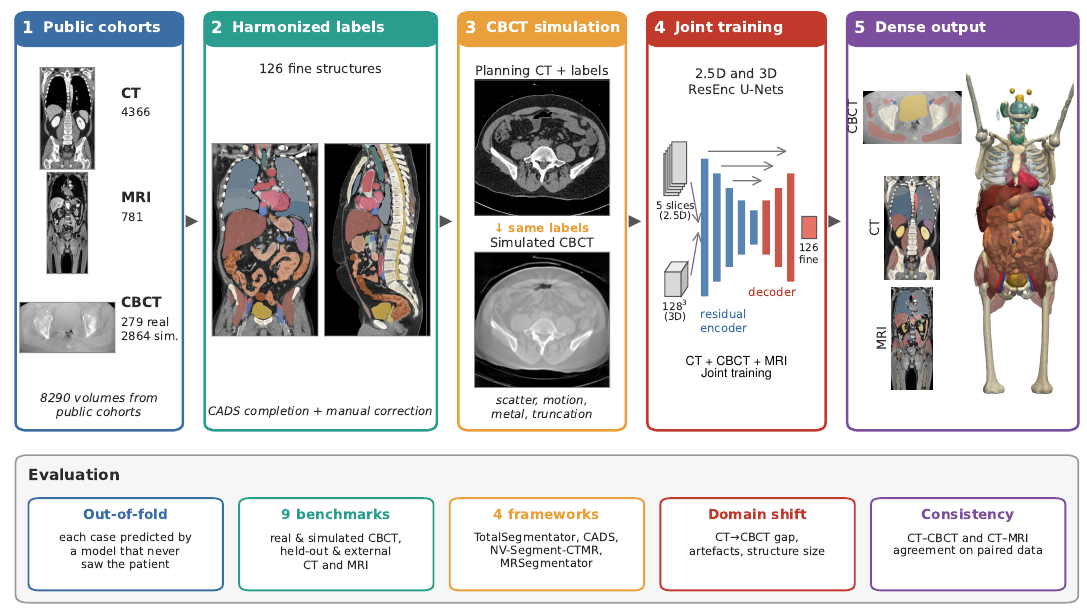}
\caption{Overview of IMPACT-Seg. Public CT, MRI and CBCT cohorts are harmonized into one label map of 126 structures (coronal and sagittal views of a held-out CT), CBCTs are simulated from source CT volumes with voxel-aligned labels, and two networks, a 2.5D and a 3D residual-encoder U-Net, are trained jointly on the three modalities with the same loss and optimization settings. Bottom: evaluation protocol.}
\label{fig:overview}
\end{figure*}

IMPACT-Seg segments 126 structures from the head and neck to the pelvis using a single network with shared parameters across CT, CBCT, and MRI, without requiring paired inputs. Implemented with KonfAI, its 2.5D and 3D architectures benefit from fast, memory-efficient inference, achieving substantially shorter end-to-end runtimes than the evaluated nnU-Net-based pipelines (Figure~\ref{fig:overview}).

The contributions are:
\begin{itemize}
\item a multimodal segmentation framework with a shared 126-structure label space and joint CT, CBCT, and MRI training;
\item a curated collection of 8290 public volumes, supported by annotation harmonization, model-assisted reference completion, and physics-based CBCT simulation with voxel-aligned labels;
\item a comprehensive evaluation of 2.5D and 3D segmentation models across CT, CBCT, and MRI, including comparisons with existing methods, training-data ablations, external validation, and reference-free failure detection;
\item an assessment of cross-modality correspondence in predictions and internal features, with downstream evaluation on two multimodal Learn2Reg registration tasks.
\end{itemize}

Code, datasets, and pretrained models are available at \url{https://github.com/HEMONCedric/IMPACTSeg}.

\section{Related Work} \label{sec:related}

\textbf{Dense anatomical segmentation.} nnU-Net \citep{isensee2021nnunet} and its residual-encoder variants \citep{isensee2024nnunetrevisited} provide strong volumetric baselines and underlie most of the frameworks below. Built on nnU-Net, TotalSegmentator \citep{wasserthal2023totalsegmentator} and CADS \citep{xu2025cads} established broad CT coverage, and MRSegmentator \citep{hantze2025mrsegmentator} and TotalSegmentator MRI \citep{dantonoli2025totalsegmentatormri} extended it to joint CT-MRI segmentation. VISTA3D \citep{he2024vista3d} instead uses its own architecture to support automatic and interactive segmentation, and its multimodal extension NV-Segment-CTMR \citep{nvidia2025nvsegmentctmr} also learns from CT and MRI jointly. AMOS \citep{ji2022amos} supports evaluation across CT and MRI, while SynthSeg \citep{billot2023synthseg} learns contrast-robust brain segmentation from synthesized images. Promptable models such as MedSAM \citep{ma2024medsam} and nnInteractive \citep{isensee2025nninteractive} instead segment one structure at a time from user clicks, boxes, or scribbles. They accelerate annotation and correction but require interaction for each structure. These advances motivate assessing dense anatomical coverage together with prediction correspondence across radiotherapy modalities.

\textbf{Learning across datasets and modalities.} MultiTalent \citep{ulrich2023multitalent} combines CT datasets with conflicting label definitions. Hermes \citep{gao2024hermes} conditions universal segmentation on task and modality priors, UniSeg \citep{ye2023uniseg} uses learned prompts for multi-task segmentation and representation transfer, and MulModSeg \citep{li2025mulmodseg} combines unpaired CT and MRI through modality-conditioned text embeddings and alternating training. COSST and INERRANT use iterative pseudo-labeling to learn from partial annotations~\citep{liu2024cosst,petit2021relabeling}.
IMPACT-Seg instead harmonizes references into one dense label map and shares a single network across modalities, without task- or modality-specific conditioning. Unlike the nnU-Net-based frameworks above, it is implemented and trained in KonfAI \citep{boussot2025konfai}, although its residual-encoder architecture follows similar design principles; memory-bounded patch inference segments a whole-body CT in seconds rather than minutes (Section~\ref{sec:results_cost}). The released models are distributed as KonfAI applications. The scientific contribution centres on extending this supervision to CBCT and on evaluating accuracy, prediction agreement, and spatial feature correspondence within the same framework.

\textbf{CBCT segmentation for radiotherapy.} Limited annotations and acquisition artefacts have motivated several strategies: propagation of planning contours by deformable registration \citep{liang2021dir}, CBCT-to-CT synthesis before or jointly with segmentation \citep{liang2019generating,kurz2019cbct,dahiya2021multitask}, direct training with limited CBCT annotations and complementary CT data \citep{fu2020pelvic,leger2020cross}, and training on CBCT simulated from CT \citep{alam2021esophagus,dahiya2021multitask}. Most studies address a small set of organs within one anatomical region. The SynthRAD challenges \citep{thummerer2023synthrad,rogowski2026generating} provide real CT-CBCT and CT-MRI data for broader multimodal evaluation, although correspondence-based analyses remain sensitive to the registration used to align the acquisitions \citep{boussot2026misalignment}.

\textbf{Anatomical representations and semantic similarity.} Learned features provide an alternative to raw intensities for comparing images across modalities.
SAM \citep{yan2022sam} learns pixel-wise anatomical embeddings through self-supervised contrastive learning, and UAE \citep{bai2025uae} learns embeddings for cross-modality landmark matching.
Anatomix \citep{dey2025learning} uses randomized volumetric synthesis and contrastive learning to obtain representations robust to appearance variation, with applications to registration and few-shot segmentation.
IMPACT loss \citep{boussot2025impact} compares features from pretrained segmentation models to define a semantic similarity objective for multimodal registration.
IMPACT-Seg uses dense segmentation in a shared anatomical label space as its sole training objective. After training, we assess the resulting cross-modality spatial correspondence and evaluate decoder feature similarity for registration.

\section{Methods}

\subsection{ Problem Setting}
\label{sec:formulation}

Let $x_m:\Omega_m\rightarrow\mathbb{R}$ denote an image of modality $m\in\{\mathrm{CT},\mathrm{CBCT},\mathrm{MRI}\}$, defined on the voxel grid $\Omega_m$. All modalities share the same label space $\mathcal{C}=\{0,\ldots,K\}$, with $K=126$ anatomical structures and background class 0. Each image has a reference label map $y_m:\Omega_m\rightarrow\mathcal{C}$.

A single network with parameters $\theta$ is trained jointly on unpaired images from all three modalities, without modality-specific conditioning. It predicts a segmentation $\widehat y_m$ from the
voxel-wise class probabilities $p_\theta(c\mid x_m,v)$, where $v\in\Omega_m$. Training uses a common segmentation loss (Section~\ref{sec:training}), without explicitly enforcing cross-modality consistency in predictions or features.

Accuracy compares the prediction $\widehat y_m$ with the reference $y_m$ of the same image. Prediction agreement instead compares the predictions obtained independently on two images of the same patient, $x_{m_1}$ and $x_{m_2}$, for instance a planning CT and a treatment CBCT. If segmentation were invariant to the imaging modality, both predictions would delineate the same anatomy and coincide once expressed on a common grid. Let $\phi:\Omega_{m_1}\rightarrow\Omega_{m_2}$ map each voxel $v$ of the first image to its anatomically corresponding position $\phi(v)$ in the second, so that $\widehat y_{m_2}\circ\phi$ is the second prediction resampled onto the grid of the first. Agreement is the mean per-structure Dice between the two predictions,
\begin{equation}
\mathrm{Agr}(x_{m_1},x_{m_2})=\frac{1}{|\mathcal{C}_{12}|}\sum_{c\in\mathcal{C}_{12}}
\frac{2\,\bigl|\{v:\widehat y_{m_1}(v)=c\}\cap\{v:\widehat y_{m_2}(\phi(v))=c\}\bigr|}
{\bigl|\{v:\widehat y_{m_1}(v)=c\}\bigr|+\bigl|\{v:\widehat y_{m_2}(\phi(v))=c\}\bigr|},
\label{eq:agreement}
\end{equation}
where $\mathcal{C}_{12}\subseteq\mathcal{C}\setminus\{0\}$ contains the structures present in the reference of the pair and included in the label space of the evaluated method. For simulated CBCTs, reconstructed on the grid of their source CT, $\phi$ is the identity; for real pairs, $\phi$ is obtained by deformable registration (Section~\ref{sec:evaluation}).

Agreement does not compare predictions with reference annotations and therefore measures consistency rather than correctness. A method that makes the same error on both images obtains a high agreement, whereas residual misregistration in $\phi$ lowers it independently of segmentation quality. Agreement is therefore always reported together with accuracy.

The framework combines a common non-overlapping label definition, CBCT simulation that preserves the source CT grid and annotations, and joint training with shared network parameters across modalities. Internal correspondence is then assessed through features $F_\theta^\ell(x_m)$ at network depth $\ell$, and downstream utility through feature-driven deformable registration.

\subsection{Dataset Curation and Harmonization} \label{sec:dataset}

Each public cohort was converted to a common per-patient format comprising the image, its reference label map, and its source cohort. Label maps were stored on the voxel grid of their image, with its original spacing and origin. The collection comprises 8290 volumes: 4366 CTs from 15 cohorts; 552 MRIs from AMOS22, TotalSegmentator MRI, and the MRI component of the multimodal total-body dynamic PET/CT/MRI dataset (MTBD) \citep{hinge2026multimodal}; 229 MRIs and 279 CBCTs from SynthRAD2023 and SynthRAD2025 \citep{thummerer2023synthrad,rogowski2026generating}; and 2865 CBCT simulations derived from 955 TotalSegmentator CTs (Section~\ref{sec:cbct_simulation}). Together, the cohorts cover the head and neck, thorax, abdomen, pelvis, and musculoskeletal anatomy, with MRI contributing abdominal, pelvic, and whole-body acquisitions. Table~\ref{tab:data_overview} lists the cohorts with their pairing, fold partition, and use in each experiment.

% Data overview: cohorts, pairing and use of each cohort in the experiments (generated layout; edit the rows directly).
% Counts: IMPACTSeg_GitHub/splits/fold*/ (train/validation/test), paper_v2/benchmarks/internal_test.csv
% (union of the five test partitions) and the evaluated case counts of Tables main_benchmarks and pancreas_cbct.
\begin{table*}[!tbp]
\centering
\footnotesize
\definecolor{grouprow}{RGB}{236,240,245}
\definecolor{zebra}{RGB}{248,249,251}
\definecolor{regiongray}{RGB}{95,95,95}
\definecolor{modCT}{RGB}{44,127,184}
\definecolor{modCBCT}{RGB}{230,126,34}
\definecolor{modMRI}{RGB}{65,171,93}
\definecolor{modMIX}{RGB}{150,150,150}
\definecolor{tagint}{RGB}{212,239,236}
\definecolor{tagacc}{RGB}{214,232,248}
\definecolor{tagagr}{RGB}{253,229,204}
\definecolor{tagabl}{RGB}{220,240,218}
\definecolor{tagfeat}{RGB}{232,224,245}
\definecolor{tagunc}{RGB}{250,220,222}
\definecolor{tagreg}{RGB}{228,228,228}
\newcommand{\mCT}{\cellcolor{modCT}}
\newcommand{\mCBCT}{\cellcolor{modCBCT}}
\newcommand{\mMRI}{\cellcolor{modMRI}}
\newcommand{\mMIX}{\cellcolor{modMIX}}
\newcommand{\hit}[2]{\cellcolor{#1}#2}
\newcommand{\colref}[1]{{\scriptsize\color{regiongray}#1}}
\DeclareRobustCommand{\modbox}[1]{\textcolor{#1}{\rule{1.6ex}{1.6ex}}}
\caption{Datasets and their use. Colour bar: imaging modality (\modbox{modCT}~CT, \modbox{modCBCT}~CBCT, \modbox{modMRI}~MRI, \modbox{modMIX}~multimodal pairs). Training-pool cohorts are used for training, validation and testing in the five-fold cross-validation; each fold draws 624/100/57 training/validation/test volumes per modality, in proportion to cohort size. Vol.: volumes, before the eligibility criteria for external cohorts (Section~\ref{app:benchmarks}). Paired: image of the same patient in another modality, voxel-aligned by construction ($\equiv$) or registered ($\rightleftarrows$). Use in evaluation (shaded cells; numbers are evaluated cases, or image pairs for Feat.): Int., internal test set formed by the union of the five test partitions, scored out of fold on all 126 structures (Figs.~\ref{fig:internal_groups}, \ref{fig:internal_labels}--\ref{fig:internal}); Acc., benchmark accuracy on the labels shared with the compared methods (Table~\ref{tab:main_benchmarks}; Pancreatic-CT-CBCT-SEG: Table~\ref{tab:pancreas_cbct}); VISCERAL CT combines 13 Gold and 45 Silver Corpus cases; Agr., cross-modality prediction agreement and CT-to-CBCT domain shift (Figs.~\ref{fig:domain_gap}, \ref{fig:artefacts}, \ref{fig:agreement3d}); Abl., training-data ablation (Table~\ref{tab:ablation_data}); Feat., feature consistency (Table~\ref{tab:feature_consistency}); Unc., failure detection and five-fold ensembles (Table~\ref{tab:uncertainty_detection}); Reg., Learn2Reg registration (Table~\ref{tab:learn2reg}). Benchmark cases taken from the training pool are scored out of fold (Section~\ref{sec:evaluation}).}
\label{tab:data_overview}
\setlength{\tabcolsep}{4pt}
\renewcommand{\arraystretch}{1.18}
\fittable{%
\begin{tabular}{@{}p{1.4pt}@{\hspace{5pt}}l>{\color{regiongray}}lrl*{7}{>{\centering\arraybackslash}p{8mm}}@{}}
\toprule
& & & & & \multicolumn{7}{c}{\textbf{Use in evaluation}}\\
\cmidrule(l{2pt}){6-12}
& \textbf{Cohort} & \textbf{\color{black}Region} & \textbf{Vol.} & \textbf{Paired} & \textbf{Int.} & \textbf{Acc.} & \textbf{Agr.} & \textbf{Abl.} & \textbf{Feat.} & \textbf{Unc.} & \textbf{Reg.}\\
& & & & & \colref{Fig.~\ref{fig:internal}} & \colref{Tab.~\ref{tab:main_benchmarks}} & \colref{Fig.~\ref{fig:domain_gap}} & \colref{Tab.~\ref{tab:ablation_data}} & \colref{Tab.~\ref{tab:feature_consistency}} & \colref{Tab.~\ref{tab:uncertainty_detection}} & \colref{Tab.~\ref{tab:learn2reg}}\\
\midrule
\mCT & \multicolumn{11}{l}{\cellcolor{grouprow}\rule{0pt}{2.7ex}\textbf{Training pool \textperiodcentered{} CT \textperiodcentered{} 15 cohorts, 4\,366 volumes}}\\
\mCT & TotalSegmentator v2~\citep{wasserthal2023totalsegmentator} & Whole body & 955 & Sim.\ CBCT $\equiv$ & \hit{tagint}{$\bullet$} &  & \hit{tagagr}{$\bullet$} & \hit{tagabl}{$\bullet$} & \hit{tagfeat}{48} &  & \\
\rowcolor{zebra}\mCT & AbdomenCT-1K~\citep{ma2022abdomenct1k} & Abdomen & 812 & -- & \hit{tagint}{$\bullet$} &  &  &  &  &  & \\
\mCT & SAROS~\citep{koitka2024saros} & Whole body & 677 & -- & \hit{tagint}{$\bullet$} &  &  &  &  &  & \\
\rowcolor{zebra}\mCT & MSD (CT tasks)~\citep{simpson2019msddata,antonelli2022msd} & Thorax, abdomen & 643 & -- & \hit{tagint}{$\bullet$} &  &  &  &  &  & \\
\mCT & AMOS CT~\citep{ji2022amos} & Abdomen & 271 & -- & \hit{tagint}{$\bullet$} &  &  &  &  &  & \\
\rowcolor{zebra}\mCT & KiTS21~\citep{heller2023kits21} & Abdomen & 253 & -- & \hit{tagint}{$\bullet$} &  &  &  &  &  & \\
\mCT & VerSe~\citep{sekuboyina2021verse} & Spine & 247 & -- & \hit{tagint}{$\bullet$} &  &  &  &  &  & \\
\rowcolor{zebra}\mCT & LiTS~\citep{bilic2023lits} & Abdomen & 128 & -- & \hit{tagint}{$\bullet$} &  &  &  &  &  & \\
\mCT & CT-ORG~\citep{rister2020ctorg} & Thorax to pelvis & 96 & -- & \hit{tagint}{$\bullet$} &  &  &  &  &  & \\
\rowcolor{zebra}\mCT & VISCERAL Silver Corpus~\citep{jimenez2016visceral} & Trunk & 87 & -- & \hit{tagint}{$\bullet$} & \hit{tagacc}{45} &  & \hit{tagabl}{$\bullet$} &  &  & \\
\mCT & VISCERAL Gold Corpus~\citep{jimenez2016visceral} & Trunk & 32 & -- & \hit{tagint}{$\bullet$} & \hit{tagacc}{13} &  & \hit{tagabl}{$\bullet$} &  &  & \\
\rowcolor{zebra}\mCT & Pancreas-CT~\citep{roth2016pancreasct,roth2015deeporgan} & Abdomen & 58 & -- & \hit{tagint}{$\bullet$} &  &  &  &  &  & \\
\mCT & BCV Abdomen~\citep{landman2015bcv} & Abdomen & 40 & -- & \hit{tagint}{$\bullet$} &  &  &  &  &  & \\
\rowcolor{zebra}\mCT & HaN-Seg~\citep{podobnik2023hanseg} & Head and neck & 35 & -- & \hit{tagint}{$\bullet$} &  &  &  &  &  & \\
\mCT & CHAOS CT~\citep{kavur2021chaos} & Abdomen & 32 & -- & \hit{tagint}{$\bullet$} &  &  &  &  &  & \\
\mCBCT & \multicolumn{11}{l}{\cellcolor{grouprow}\rule{0pt}{2.7ex}\textbf{Training pool \textperiodcentered{} CBCT \textperiodcentered{} 2 cohorts, 3\,143 volumes}}\\
\mCBCT & Simulated (RTK)~\citep{rit2014rtk} & Whole body & 2\,864 & CT $\equiv$ & \hit{tagint}{$\bullet$} & \hit{tagacc}{497} & \hit{tagagr}{$\bullet$} & \hit{tagabl}{$\bullet$} & \hit{tagfeat}{48} &  & \\
\rowcolor{zebra}\mCBCT & SynthRAD 2023/2025~\citep{thummerer2023synthrad,rogowski2026generating} & Brain to pelvis & 279 & CT $\rightleftarrows$ & \hit{tagint}{$\bullet$} & \hit{tagacc}{139} & \hit{tagagr}{$\bullet$} & \hit{tagabl}{$\bullet$} & \hit{tagfeat}{57} &  & \\
\mMRI & \multicolumn{11}{l}{\cellcolor{grouprow}\rule{0pt}{2.7ex}\textbf{Training pool \textperiodcentered{} MRI \textperiodcentered{} 4 cohorts, 781 volumes \textperiodcentered{} test partitions resampled for each fold}}\\
\mMRI & TotalSegmentator MRI~\citep{dantonoli2025totalsegmentatormri} & Whole body & 374 & -- & \hit{tagint}{$\bullet$} & \hit{tagacc}{120} &  &  &  &  & \\
\rowcolor{zebra}\mMRI & SynthRAD 2023/2025~\citep{thummerer2023synthrad,rogowski2026generating} & Brain to pelvis & 229 & CT $\rightleftarrows$ & \hit{tagint}{$\bullet$} & \hit{tagacc}{52} & \hit{tagagr}{$\bullet$} &  & \hit{tagfeat}{11} &  & \\
\mMRI & MTBD~\citep{hinge2026multimodal} & Whole body & 120 & CT $\rightleftarrows$ & \hit{tagint}{$\bullet$} &  &  &  &  &  & \\
\rowcolor{zebra}\mMRI & AMOS MR~\citep{ji2022amos} & Abdomen & 58 & -- & \hit{tagint}{$\bullet$} & \hit{tagacc}{20} &  &  &  &  & \\
\mCT & \multicolumn{11}{l}{\cellcolor{grouprow}\rule{0pt}{2.7ex}\textbf{Paired evaluation images \textperiodcentered{} never used for training or validation}}\\
\mCT & SynthRAD planning CT~\citep{thummerer2023synthrad,rogowski2026generating} & Thorax to pelvis & 296 & CBCT, MRI $\rightleftarrows$ &  &  & \hit{tagagr}{$\bullet$} &  & \hit{tagfeat}{$\bullet$} &  & \\
\mMIX & \multicolumn{11}{l}{\cellcolor{grouprow}\rule{0pt}{2.7ex}\textbf{External cohorts \textperiodcentered{} never used for training or validation}}\\
\mCT & TotalSeg.\ v3 CT (new)~\citep{wasserthal2023totalsegmentator} & Whole body & 712 & -- &  & \hit{tagacc}{247} &  & \hit{tagabl}{$\bullet$} &  & \hit{tagunc}{$\bullet$} & \\
\rowcolor{zebra}\mMRI & TotalSeg.\ v3 MRI (new)~\citep{dantonoli2025totalsegmentatormri} & Whole body & 680 & -- &  & \hit{tagacc}{253} &  &  &  & \hit{tagunc}{$\bullet$} & \\
\mCBCT & Pancreatic-CT-CBCT-SEG~\citep{hong2022pancreatic} & Abdomen & 120 & CT, 2 CBCTs &  & \hit{tagacc}{80} &  &  &  &  & \\
\rowcolor{zebra}\mMIX & Learn2Reg AbdomenMRCT~\citep{hering2023learn2reg} & Abdomen & -- & MRI--CT &  &  &  &  &  &  & \hit{tagreg}{$\bullet$}\\
\mMIX & Learn2Reg ThoraxCBCT~\citep{hering2023learn2reg} & Thorax & -- & CT--CBCT &  &  &  &  &  &  & \hit{tagreg}{$\bullet$}\\
\bottomrule
\end{tabular}
}
\end{table*}

All volumes were reoriented to RAS, and image-label geometry, intensity scaling, laterality, anatomical numbering, and label-volume outliers were audited. Flagged cases were corrected or excluded when their metadata could not be trusted. Detailed geometry checks and spacing criteria are given in Section~\ref{app:implementation}.

\textbf{Shared label space.} All cohorts, whether CT, MRI, real CBCT, or simulated CBCT, are annotated in one label space. It follows the label definitions of CADS \citep{xu2025cads}, from which 131 classes were retained, 126 of which are evaluated (Section~\ref{sec:training}). Because each voxel carries a single label, classes that enclose finer ones were removed: the whole heart, for example, is represented by the four cardiac chambers and the myocardium, the larynx by the glottis and supraglottis, and the peritoneal cavity by the abdominal organs it contains. The few voxels still claimed by two labels were assigned by a fixed priority between the CADS sub-models from which the labels originate. The label map covers abdominal and pelvic organs, bowel, lung lobes, cardiac chambers and major vessels, individually numbered vertebrae and ribs, major bones and muscles, and head-and-neck organs at risk (Figure~\ref{fig:dense}). Training, stored references, and evaluation use the same label convention.

\subsection{Reference Annotation Pipeline} \label{sec:annotations}

Each source dataset annotates only a subset of the 126 structures. Directly combining these datasets would therefore introduce incomplete and inconsistent supervision, with unannotated structures treated as background. To address this, we used model-assisted pseudo-labeling to complete the references in a shared label space, followed by iterative refinement to reduce missing or anatomically implausible annotations.

(i) Source annotations were mapped to the shared label space and kept unchanged in all later steps. 

(ii) Voxels without a source label were filled with the predictions of an existing modality-specific model: CADS \citep{xu2025cads} for CT, and NV-Segment-CTMR \citep{nvidia2025nvsegmentctmr}, with selected labels such as the vertebrae taken from TotalSegmentator MRI \citep{dantonoli2025totalsegmentatormri}, for MRI. Simulated CBCTs inherit the completed reference of their source CT. For SynthRAD (\href{https://huggingface.co/datasets/VBoussot/synthrad2025-impact-registration}{registration files}) and MTBD, whose MRIs and CBCTs are paired with a CT of the same patient, the reference was built on that CT and transferred on the shared grid to the registered CBCT or MRI; selected MRI labels were then replaced by TotalSegmentator MRI predictions on the MRI itself. 

(iii) The initial pseudo-labels were further refined over successive training generations. At each generation, an independent network was trained on the current references, and its predictions were used to complete missing structures and replace implausible automatically generated annotations within anatomically plausible regions. The refined references were then used to train the next generation, progressively updating the pseudo-labels while preserving the original source annotations.

(iv) All completed references were visually reviewed, and erroneous structures were corrected manually or interactively with nnInteractive \citep{isensee2025nninteractive}. MRI required the most correction: \amosmrmanual{} of \amosmrtotal{} AMOS-MR cases and \totalsmrmanual{} of \totalsmrtotal{} TotalSegmentator-MRI cases were edited, mostly for structures missing from the automatic predictions. All methods were evaluated against the same references.

\subsection{CBCT Simulation} \label{sec:cbct_simulation}

\begin{figure*}[!tbp]
\centering
\includegraphics[width=\textwidth]{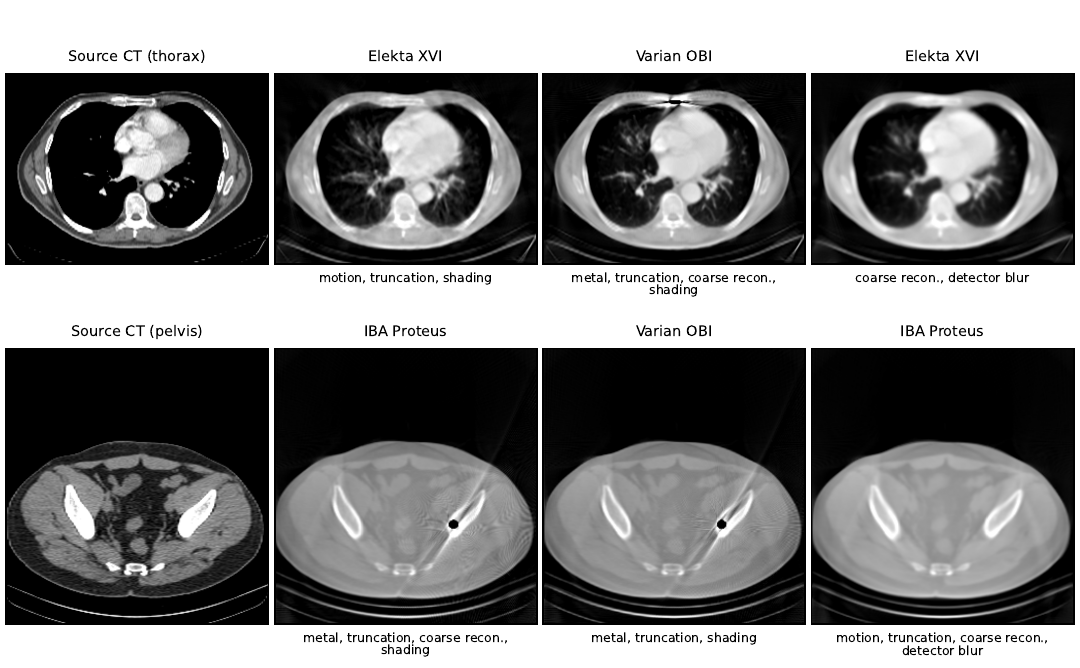}
\caption{Simulated CBCTs. For two patients, the source CT and three simulated CBCT realizations on the same axial slice, with the scanner geometry and the artefacts simulated for each realization. Implants appear on the displayed slices: sternal wires (thorax, Varian OBI) and a hip prosthesis (pelvis). The CBCTs are reconstructed back onto the CT grid and inherit its labels voxel by voxel.}
\label{fig:cbct_simulation}
\end{figure*}

CBCTs were simulated from 955 TotalSegmentator CT volumes with the Reconstruction Toolkit (RTK) \citep{rit2014rtk}. CT-derived attenuation volumes were projected with circular cone-beam geometries representative of on-board imagers and reconstructed with FDK onto the source CT grid, preserving voxel-wise label correspondence (Figure~\ref{fig:cbct_simulation}).  
CBCT simulations used five scanner profiles derived from the acquisition parameters of the SynthRAD2025 centres (three Elekta XVI configurations, one IBA Proteus, and one Varian TrueBeam on-board imager). Each profile defined the acquisition geometry, tube current, exposure time, and detector characteristics. Three realizations were generated per CT, each with an independently sampled scanner profile and randomly enabled artefacts of varying severity. These included noise, scatter, detector imperfections, shading, truncation, miscentring, respiratory motion, and metal artefacts representative of radiotherapy CBCT \citep{aapm104}. Projections used a monoenergetic attenuation model, without explicit beam-hardening simulation. Further implementation details are provided in Section~\ref{app:implementation}.

\subsection{Networks and Training} \label{sec:training}

The 2.5D and 3D residual-encoder U-Nets were implemented in KonfAI \citep{boussot2025konfai} and trained jointly on all modalities, without modality-specific heads or fine-tuning. Images were resampled to $1.5\times1.5\times2.0$~mm and standardized to zero mean and unit variance over the volume. The networks predict 126 evaluated structures plus background. Five additional output classes (the cochleae, optic chiasm, and optic nerves) were not reliably learned at this resolution and were suppressed in every evaluation.

The 2.5D network uses five adjacent axial slices to predict the central $288\times288$ patch, with batch normalization and a batch size of 16. The 3D network uses single-channel $128^3$ patches, instance normalization, and a batch size of one. Both were trained in mixed precision for $2\,000\,000$ SGD steps with Nesterov momentum 0.99, weight decay $3\times10^{-5}$, and an initial learning rate of 0.01 decayed polynomially with exponent 0.9. Encoder specifications and augmentations are detailed in Section~\ref{app:implementation}.

The training loss combines a weighted cross-entropy with soft Dice terms:
\begin{equation}
\mathcal{L}_{\mathrm{seg}}=\mathcal{L}_{\mathrm{CE}}+\mathcal{D}_{\mathrm{pres}}+0.2\,\mathcal{D}_{\mathrm{all}}+0.2\,\mathcal{D}_{\mathrm{pres}}^{(1/2)}+0.1\,\mathcal{D}_{\mathrm{pres}}^{(1/4)}.
\label{eq:seg_loss}
\end{equation}
In $\mathcal{L}_{\mathrm{CE}}$, the background class is weighted 0.2 and all other classes 1. Label smoothing of 0.02 is applied: instead of a one-hot target, 2\% of the target probability is spread uniformly over all classes, which discourages over-confident predictions.

The Dice terms are computed on each training patch. With $p_c(v)=p_\theta(c\mid x,v)$ and $y_c(v)=1$ if $y(v)=c$ and 0 otherwise, the soft Dice of class $c$ is
\begin{equation}
D_c=\frac{2\sum_{v}p_c(v)\,y_c(v)+\varepsilon}{\sum_{v}p_c(v)+\sum_{v}y_c(v)+\varepsilon},
\label{eq:soft_dice}
\end{equation}
where the sums run over the voxels of the patch and $\varepsilon=10^{-6}$ avoids division by zero. For a set of classes $\mathcal{A}$, the Dice loss averages $1-D_c$ over $c\in\mathcal{A}$ and over the patches of the batch. $\mathcal{D}_{\mathrm{pres}}$ uses the foreground classes present in the reference of the patch, and $\mathcal{D}_{\mathrm{all}}$ uses all 126 foreground output classes, including those absent from the patch. The decoder also outputs segmentations at one half and one quarter of the final resolution (deep supervision). These auxiliary outputs, $\mathcal{D}_{\mathrm{pres}}^{(1/2)}$ and $\mathcal{D}_{\mathrm{pres}}^{(1/4)}$, are compared with references downsampled by nearest-neighbour interpolation, using present classes only.

Voxels without a foreground annotation are treated as background, with no ignore label or source-specific validity mask. Incomplete source annotations are addressed upstream through reference completion and correction (Section~\ref{sec:annotations}), and the reduced background weight in $\mathcal{L}_{\mathrm{CE}}$ attenuates the penalty on structures left unannotated. $\mathcal{D}_{\mathrm{all}}$ additionally penalizes residual probabilities on classes absent from the patch.
Foreground-centred sampling was used with probability 0.7; spatial flips were omitted to preserve laterality. Training one fold required approximately six days for 2.5D and nine days for 3D on an RTX A6000. 

\textbf{Training-data ablation.} Four additional 2.5D networks, trained on the fold-1 partition (Section~\ref{sec:splits}), assessed the training modalities using a common, shorter AdamW training schedule and a fixed checkpoint at 200\,000 steps (Section~\ref{app:implementation}). Training sets comprised CT only (624 volumes), CT plus real CBCT (763), CT plus real and simulated CBCT (1248, including 485 simulations), and all modalities (1872, adding 624 MRI volumes). Slices were sampled uniformly, yielding CBCT shares of 10\% with real CBCT alone and 45\% with simulations. Validation used only the included modalities; all variants were evaluated on identical fold-1 test and benchmark patient-label pairs, with each configuration compared against its predecessor.

\subsection{Data Splits, Model Selection and Inference}
\label{sec:splits}

Five patient-stratified folds were constructed with 781 cases per modality (CT, CBCT, MRI) per fold with 624 for training, 100 for validation, and 57 for testing. Thus, each training fold contained 1872 volumes equally balanced across the three modalities. CT and CBCT cases were rotated across folds, whereas for MRI, test cases were randomly resampled from the available training pool for each fold due to limited data availability. Not all cases were utilized by the splits, therefore those left-over cases are used for further evaluations (see \ref{sec:evaluation}). Each network (2.5D and 3D) was trained once per fold, giving five models per network; the model trained on the training partition of fold $k$ is referred to as the fold-$k$ model. \\

CT sampling was proportional to source-cohort size to retain dataset diversity; CBCT selection covered anatomical regions and noise conditions while excluding simulated variants of TotalSegmentator test patients. All acquisitions and simulated variants of a patient remained in the same partition within a fold. %CT and CBCT cohorts rotate across folds; evaluation therefore uses the patient-exclusion procedure below. 
Validation loss was measured every $100\,000$ steps on the 300 validation cases, and the checkpoint with the lowest loss was retained.

Inference uses the training preprocessing, patch size, and 50\% overlap. At each voxel, the label of the most confident patch is retained. The label map is restored to the native grid by nearest-neighbour interpolation rather than by resampling the class probabilities, which keeps inference memory-bounded but limits boundary precision at native resolution.
Each case uses one eligible fold model. On external TotalSegmentator v3 cohorts, an additional ensemble averages the five folds' softmax predictions before patch assembly. Section~\ref{sec:results_reconstruction} compares confidence-based assembly with Gaussian-weighted probability averaging. 

\subsection{Evaluation Protocol} \label{sec:evaluation}

\textbf{Out-of-fold evaluation.} As CT and CBCT rotated across folds, trained models were evaluated on the hold-out test set. Cases were segmented by a fold model if the case was present in the respective test set. Left over cases from the split were assigned to a fold model and encoded with a fixed hash.
For simulated CBCT, the patient is identified by its source CT, so that no model is evaluated on a CBCT derived from a training CT.

\textbf{Benchmarks and compared methods.} CADS \citep{xu2025cads}, TotalSegmentator \citep{wasserthal2023totalsegmentator}, and NV-Segment-CTMR \citep{nvidia2025nvsegmentctmr} were evaluated on CT and CBCT; MRSegmentator \citep{hantze2025mrsegmentator}, TotalSegmentator MRI \citep{dantonoli2025totalsegmentatormri}, and NV-Segment-CTMR on MRI. As MRSegmentator and TotalSegmentator MRI also support CT segmentation, both were applied to CT and MRI for the cross-modality prediction agreement analysis. Fixed software and weight versions are listed in Table~\ref{tab:comparator_versions}; released inference settings and verified shared-label mappings were used. CADS was excluded from VISCERAL because of training overlap. Competing CT frameworks had access to the TotalSegmentator source cohort of the simulated CBCT benchmark, whereas IMPACT-Seg excluded the evaluated patients from training and validation. External evaluation used cases added in version~3 of the TotalSegmentator datasets (712 CT and 680 MRI volumes absent from version~2). The evaluated TotalSegmentator weights predate this release, but training overlap cannot be excluded (Section~\ref{app:implementation}). Shared eligibility and prediction-availability criteria yielded 247 CT and 253 MRI volumes (Section~\ref{app:benchmarks}).

\textbf{External radiotherapy CBCT.} Independent validation used the Pancreatic-CT-CBCT-SEG collection \citep{hong2022pancreatic}: 40 patients treated with ablative radiotherapy for locally advanced pancreatic cancer at Memorial Sloan Kettering Cancer Center, each with a contrast-enhanced planning CT and two non-contrast treatment CBCTs (Varian TrueBeam or OBI on-board). All images were acquired in deep-inspiration breath-hold verified with an external respiratory monitor. Expert contours cover the small bowel and stomach-duodenum within the low-dose planning target volume plus a 1~cm margin; ten CBCTs have two independent observers. Stomach and duodenum predictions were merged. Their union with the small bowel was evaluated as a single gastrointestinal (GI) region to reduce sensitivity to differing duodenal label definitions. Rigidly propagated planning-CT contours and inter-observer agreement on the ten CBCTs annotated by two observers were included as clinical comparators.

\textbf{Metrics.} Dice, 95th-percentile Hausdorff distance (HD95), and average symmetric surface distance (ASD) were computed for the shared reference structures; distances are reported in millimetres. For comparisons with existing methods, only structures annotated in the original source datasets were evaluated, excluding automatically generated annotations to avoid favouring any of the models used for reference completion. Structures absent from the reference were not scored because the annotations do not exhaustively cover every field of view. All methods in a paired comparison were evaluated on the same cases and patient-label pairs.

\textbf{Statistics.} Metrics were averaged across evaluated structures within each case and then summarized across cases, with mean $\pm$ standard deviation values provided in the Supplementary Material. The primary endpoint was per-case mean Dice. Two-sided paired Wilcoxon signed-rank tests compared each framework with each IMPACT-Seg network and the two networks with one another, with Holm correction across comparisons within each benchmark. Mean Dice confidence intervals were obtained by patient-level bootstrap with $10\,000$ resamples and are reported for IMPACT-Seg in Table~\ref{tab:dice_confidence_intervals}. HD95 and ASD were secondary endpoints, outside this family of tests. Unless stated otherwise, $p$-values reported in the Results compare the 2.5D network with each competitor, and 3D comparisons are given when their conclusion differs. HD95, ASD, and label-wise results are reported in the supplementary material. Statistical significance was interpreted alongside the magnitude and anatomical relevance of each difference.

\textbf{Domain shift and consistency.}
Cross-modality robustness was assessed using two complementary metrics. Dice degradation was defined as the difference between segmentation accuracy on CT and CBCT for the same patient, computed over the same eligible structures:
\begin{equation}
\Delta_{\mathrm{Dice}} =
\mathrm{Dice}_{\mathrm{CT}} -
\mathrm{Dice}_{\mathrm{CBCT}}.
\end{equation}
Positive values indicate lower segmentation accuracy on CBCT.

Prediction agreement was defined as the mean per-structure Dice between predictions obtained independently on two modalities after bringing them into a common coordinate system (Eq.~\ref{eq:agreement}). Unlike accuracy, agreement does not measure correctness against a reference, as two predictions may agree while sharing the same errors.

Simulated CBCTs were reconstructed on the grid of their source CT and shared its reference, allowing both metrics to be evaluated without inter-acquisition anatomical changes. For real CT-CBCT and CT-MRI pairs, the images were aligned to the planning CT with the IMPACT-Reg pipeline using TotalSegmentator features \citet{boussot2026misalignment}. All methods used the same registered pairs and shared labels. Residual registration errors and anatomical differences may affect the measured consistency. CT-side accuracy against CT-derived references was therefore not interpreted as independent validation on these real pairs.

\subsection{Feature-Level Consistency Analysis} \label{sec:feature_methods}

Training constrains only the predicted label maps, without explicitly enforcing feature correspondence across modalities. We investigate whether shared segmentation supervision nevertheless produces representations that are both consistent at anatomically corresponding positions and spatially discriminative. These properties are distinct: features encoding only anatomical identity, as the output probabilities do, may be similar across modalities yet unable to distinguish positions within the same structure. Deformable registration requires both: features must agree at corresponding positions and provide spatially localized similarity.

To formalize these properties, we consider the local similarity landscape around an anatomical correspondence. For a position $v$ in the CT, let $F^\ell(x)(v)$ denote the feature vector $F_\theta^\ell(x)(v)$ at network depth $\ell$, normalized to unit length. Given a spatially aligned image $x_m$ of another modality, the cosine similarity at an in-plane displacement $u$ and the most similar displacement within a radius $R$ are
\begin{equation}
S_v^\ell(u)=\bigl\langle F^\ell(x_{\mathrm{CT}})(v),\,F^\ell(x_m)(v+u)\bigr\rangle,
\qquad
\hat u_v^\ell=\arg\max_{\|u\|\le R} S_v^\ell(u).
\label{eq:feature_landscape}
\end{equation}
Ideally, the similarity is high at the corresponding position ($u=0$ after alignment) and decreases at competing positions. A well-localized maximum provides an unambiguous correspondence, whereas broad plateaus or secondary maxima make localization less reliable. This landscape characterizes the local correspondence properties of the features rather than the complete registration objective, which also includes other similarity terms and regularization (Section~\ref{sec:registration_methods}).

We assess these properties using complementary metrics. Matched cosine similarity $S_v^\ell(0)$ measures point-wise cross-modality agreement, while linear centred kernel alignment (CKA) compares the representational structure of matched feature sets. Matched-versus-negative discrimination (AUC) evaluates whether corresponding features are more similar than features at non-corresponding positions, using negatives from another structure, which test anatomical identity, or from the same structure at least 15~mm away, which test localization within a structure. Finally, retrieval error $\|\hat u_v^\ell\|$ measures whether the highest local similarity identifies the corresponding position, jointly reflecting cross-modality consistency and spatial discrimination.

\subsection{Application to Multimodal Registration} \label{sec:registration_methods}

Decoder features were evaluated as a similarity representation for deformable registration on the Learn2Reg AbdomenMRCT and ThoraxCBCT tasks \citep{hering2023learn2reg}. We built upon our previous IMPACT-Reg work \citep{boussot2025impact}, which demonstrated the utility of pretrained segmentation features from TotalSegmentator as a semantic similarity measure for multimodal registration within a multi-resolution Elastix framework. For AbdomenMRCT, the previous approach used deep features from the TotalSegmentator M730 model, extracted in static mode and compared using an L1 distance after affine initialization. For ThoraxCBCT, early features from the M730/M731 models were used in Jacobian mode, allowing gradients to propagate through the feature extractor. In the present work, the same feature-based registration principle was retained, while the similarity representation was replaced by features from the jointly trained IMPACT-Seg network.

For AbdomenMRCT, ConvexAdam \citep{siebert2024convexadam} initialized the deformation from the predicted liver, spleen, and kidney segmentations. Subsequent refinement minimized the L1 distance between the fixed and warped moving features from the last IMPACT-Seg decoder layer, with diffusion regularization: 160 Adam iterations were followed by 180 iterations on a control grid with two-voxel spacing. Segmentation masks were used for initialization only.

For ThoraxCBCT, the previous Elastix-IMPACT pipeline \citep{boussot2025impact} was retained for initialization. Refinement combined last-decoder IMPACT-Seg feature similarity, MIND-SSC descriptors \citep{heinrich2012mind}, and diffusion regularization, with 60 iterations at approximately 4~mm and 100 iterations at 2~mm.

\subsection{Uncertainty Estimation and Failure Detection}
\label{sec:uncertainty_methods}

We investigated whether segmentation failures could be detected without reference annotations at inference time. A failure was defined as a structure with a reference Dice below 0.7, with 0.5 used in a secondary analysis. Evaluation was performed on the external TotalSegmentator v3 CT and MRI cohorts, unseen by all fold models. Three voxel-wise uncertainty signals were compared to identify structures at risk of failure.

With $p(c\mid v)=p_\theta(c\mid x,v)$ denoting the predicted class probabilities, normalized softmax entropy is
\begin{equation}
H(v)=-\frac{1}{\log|\mathcal{C}|}\sum_{c\in\mathcal{C}}p(c\mid v)\log p(c\mid v)\in[0,1],
\label{eq:entropy}
\end{equation}
ranging from 0 for a one-hot prediction to 1 for a uniform distribution. It was computed over the $|\mathcal{C}|=127$ evaluated classes, after removing the five suppressed outputs and renormalizing. Given the predicted label maps $\hat y^{(1)},\ldots,\hat y^{(F)}$ of the $F=5$ fold models, vote disagreement measures the fraction of models not voting for the majority label:
\begin{equation}
U_{\mathrm{fold}}(v)=1-\max_{c\in\mathcal{C}}\frac{1}{F}\sum_{f=1}^{F}\mathbf{1}\bigl[\hat y^{(f)}(v)=c\bigr],
\label{eq:vote}
\end{equation}
which is 0 when all models agree. Network disagreement $U_{\mathrm{net}}(v)=\mathbf{1}\bigl[\hat y^{\mathrm{2.5D}}(v)\neq\hat y^{\mathrm{3D}}(v)\bigr]$ indicates differences between the 2.5D and 3D predictions.

Each signal $u$ was summarized per predicted structure. For structure $c$, let $R_c=\{v:\widehat y(v)=c\}$ be its predicted region and $B_c=\delta^2(R_c)\setminus\epsilon^2(R_c)$ its boundary band, where $\delta^2$ and $\epsilon^2$ denote binary dilation and erosion by two voxels. The structure-level scores are the means $\bar u_c=|R_c|^{-1}\sum_{v\in R_c}u(v)$ inside the structure and $\bar u_c^{B}=|B_c|^{-1}\sum_{v\in B_c}u(v)$ in its boundary band. Two baselines were included: the mean confidence deficit $1-|R_c|^{-1}\sum_{v\in R_c}\max_{c'}p(c'\mid v)$, and the predicted volume $|R_c|$, with smaller structures ranked as more likely to fail.

Detection was assessed using Spearman correlation, AUROC, and AUPRC, with patient-level bootstrap confidence intervals ($10\,000$ resamples). Sensitivity and specificity were reported at the threshold maximizing Youden's index on the evaluated cohort; these are descriptive operating points rather than independently validated deployment thresholds. Triage analyses measured the fraction of failures captured by reviewing the 5, 10, 20, or 30\% most uncertain structures. Voxel-level confidence calibration was assessed with reliability diagrams and expected calibration error. Entirely missed structures have no predicted region from which to derive a score; they were excluded from score-based detection metrics and reported separately as undetectable failures.

\section{Results} \label{sec:results}

\subsection{Consistent Anatomy Across Modalities} \label{sec:results_gap}

We first assess prediction consistency on voxel-aligned CT-CBCT pairs, then on registered real CT-CBCT and CT-MRI pairs. The simulated pairs isolate appearance changes, whereas real pairs also retain uncertainty from registration and anatomical variation.

\begin{figure*}[!tbp]
\centering
\includegraphics[width=\textwidth]{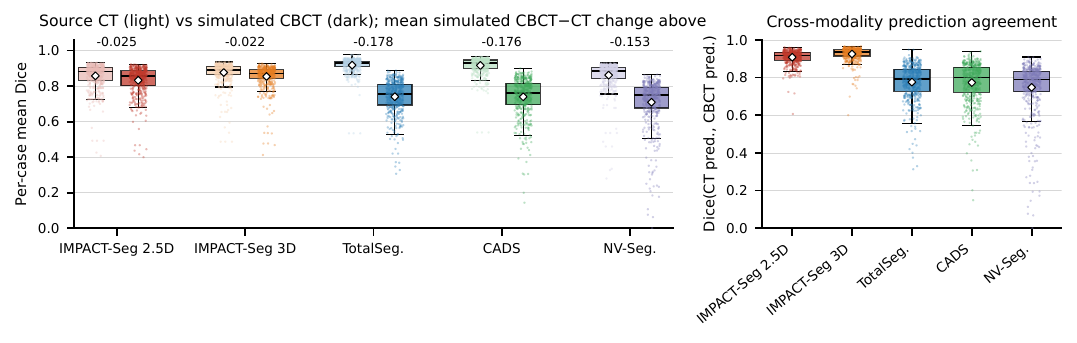}\\[2pt]
{\small (a) Simulated CBCT (TS\_RTK): Dice on the source CT and on the voxel-aligned CBCT (left, mean change above) and Dice between the CT and CBCT predictions of the same patient (right).}\\[6pt]
\includegraphics[width=\textwidth]{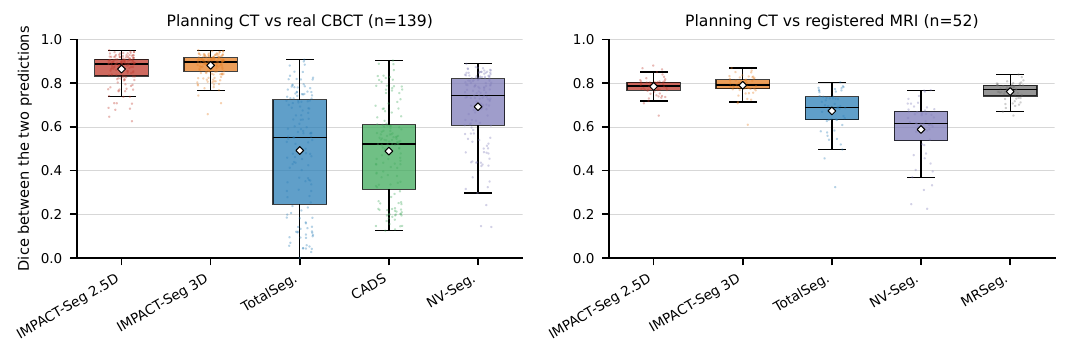}\\[2pt]
{\small (b) Real paired data (SynthRAD): Dice between the planning-CT prediction and the prediction on the registered CBCT (left) or MRI (right).}
\caption{Cross-modality consistency. (a) CT-to-CBCT domain shift on voxel-aligned simulated pairs. (b) Reference-free agreement between the predictions of the planning CT and of the deformably registered real CBCT or MRI of the same patient, on the patient-label pairs of the benchmark tables.}
\label{fig:domain_gap}
\end{figure*}

\textbf{Voxel-aligned simulated pairs.} On \res{gap.cases} TS\_RTK pairs (Figure~\ref{fig:domain_gap}a), the CT-to-CBCT Dice decrease was \resabs{gap.ims25.gap} for IMPACT-Seg 2.5D and \resabs{gap.ims3d.gap} for 3D, compared with \resabs{gap.ts.gap} for TotalSegmentator, \resabs{gap.nv.gap} for NV-Segment-CTMR, and \resabs{gap.cads.gap} for CADS (all \res{gap.ts.gapp}).

The source CTs belong to cohorts used to train TotalSegmentator and CADS, whereas IMPACT-Seg was evaluated out of fold. On the simulated CBCTs, TotalSegmentator and CADS achieved lower Dice (\res{gap.ts.cbct} and \res{gap.cads.cbct}) than IMPACT-Seg (\res{gap.ims25.cbct} and \res{gap.ims3d.cbct}). 
CT-CBCT prediction agreement reached \res{gap.ims25.agree} and \res{gap.ims3d.agree} for IMPACT-Seg, compared with \res{gap.ts.agree}, \res{gap.nv.agree}, and \res{gap.cads.agree}, respectively. Detector blur and respiratory motion were associated with the largest Dice decreases for the compared frameworks; corresponding decreases were approximately three to four times smaller for IMPACT-Seg (Figure~\ref{fig:artefacts}).

\textbf{Real paired data.} On \res{conscbct.cases} registered SynthRAD CT-CBCT pairs, agreement was \res{conscbct.ims25.agree} and \res{conscbct.ims3d.agree} for IMPACT-Seg 2.5D and 3D, compared with \res{conscbct.nv.agree} for NV-Segment-CTMR, \res{conscbct.ts.agree} for TotalSegmentator, and \res{conscbct.cads.agree} for CADS. On \res{consmri.cases} CT-MRI pairs, the corresponding values were \res{consmri.ims25.agree} and \res{consmri.ims3d.agree}, compared with \res{consmri.mrseg.agree} for MRSegmentator, \res{consmri.ts.agree} for TotalSegmentator, and \res{consmri.nv.agree} for NV-Segment-CTMR. IMPACT-Seg therefore achieved the highest agreement in both settings. These scores combine prediction consistency with residual registration uncertainty and do not measure correctness against an independent reference. Figure~\ref{fig:agreement3d} provides surface-based examples.

\subsection{Segmentation Accuracy on CBCT, CT and MRI} \label{sec:results_accuracy}

\begin{figure*}[!tbp]
\centering
\includegraphics[width=\textwidth]{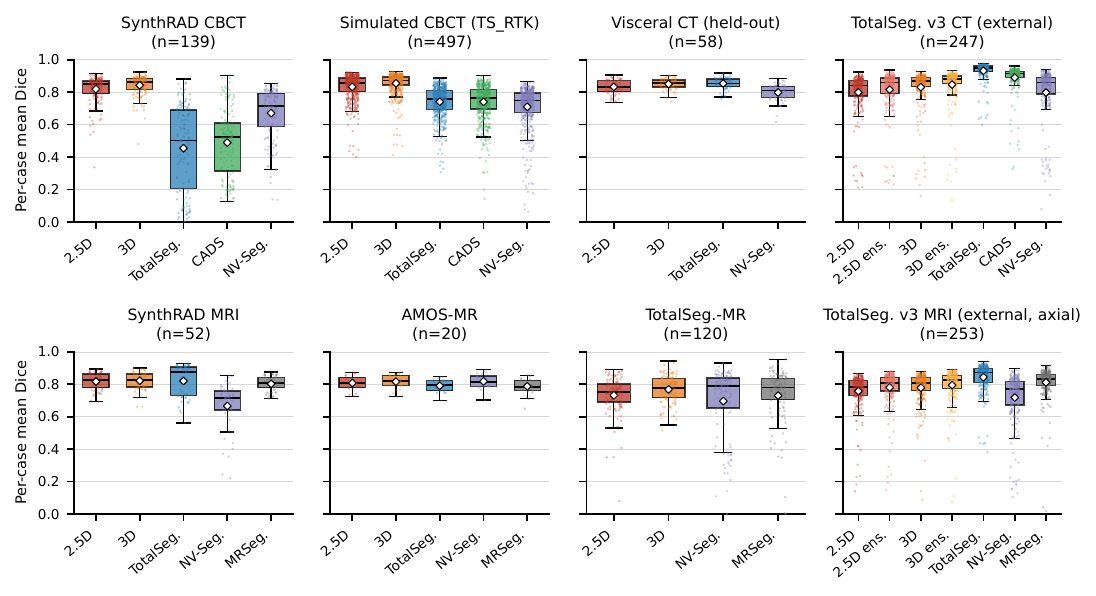}
\caption{Per-case mean Dice of IMPACT-Seg (2.5D and 3D networks; ens.: five-fold ensemble on the external cohorts) and the compared frameworks on every benchmark (boxes: interquartile range; whiskers: 1.5 IQR; diamonds: means; dots: cases). A method trained on the images of a benchmark is not shown on it (Section~\ref{sec:evaluation}).}
\label{fig:benchmarks}
\end{figure*}

Table~\ref{tab:main_benchmarks} summarizes all benchmarks; Figure~\ref{fig:benchmarks} shows the per-case Dice distributions. Figure~\ref{fig:failure} illustrates failures. Dense anatomical outputs and representative cases selected by explicit rules are shown in Section~\ref{app:qualitative}.

\begin{table}[!tbp]
\centering
\footnotesize
\caption{Mean Dice / HD95 (mm) of every method on the eight benchmarks (per-case means over the evaluated patient--label pairs, out-of-fold predictions for IMPACT-Seg; best Dice per benchmark in bold). $^{\dagger}$/$^{\ddagger}$: difference with IMPACT-Seg 2.5D/3D significant (paired Wilcoxon test on the per-case Dice, Holm-corrected, $p<0.05$). overlap: not evaluated, the method was trained on the images of this benchmark; --: no model for this modality or not applicable. On MRI, TotalSeg.\ is the MRI model of TotalSegmentator (release 2.4.0). Five-fold ensembles are reported on the external cohorts only, which no fold has seen. Standard deviations and ASD are given in Supplementary Tables~\ref{tab:synthrad_cbct_framework_overview}--\ref{tab:ts_v3_mri_framework_overview}; IMPACT-Seg Dice confidence intervals appear in Table~\ref{tab:dice_confidence_intervals}.}
\label{tab:main_benchmarks}
\setlength{\tabcolsep}{3.5pt}
\fittable{%
\begin{tabular}{lrcccccc}
\toprule
Benchmark & Cases & IMPACT-Seg 2.5D & IMPACT-Seg 3D & TotalSeg. & CADS & NV-Seg. & MRSeg.\\
\midrule
\multicolumn{8}{l}{\textit{CBCT}}\\
SynthRAD CBCT (real) & 139 & 0.819$^{\ddagger}$ / 6.8 & \textbf{0.842}$^{\dagger}$ / 5.6 & 0.454$^{\dagger\ddagger}$ / 31.4 & 0.489$^{\dagger\ddagger}$ / 19.2 & 0.671$^{\dagger\ddagger}$ / 30.8 & --\\
Simulated CBCT (TS\_RTK) & 497 & 0.833$^{\ddagger}$ / 6.1 & \textbf{0.855}$^{\dagger}$ / 4.6 & 0.741$^{\dagger\ddagger}$ / 11.7 & 0.740$^{\dagger\ddagger}$ / 10.8 & 0.710$^{\dagger\ddagger}$ / 14.3 & --\\
\midrule
\multicolumn{8}{l}{\textit{CT}}\\
VISCERAL CT & 58 & 0.833$^{\ddagger}$ / 7.4 & 0.850$^{\dagger}$ / 6.1 & \textbf{0.853}$^{\dagger}$ / 7.0 & overlap & 0.798$^{\dagger\ddagger}$ / 41.7 & --\\
TotalSeg.\ v3 CT (external) & 247 & 0.799$^{\ddagger}$ / 11.4 & 0.831$^{\dagger}$ / 7.8 & \textbf{0.932}$^{\dagger\ddagger}$ / 2.8 & 0.890$^{\dagger\ddagger}$ / 4.7 & 0.798$^{\dagger\ddagger}$ / 32.9 & --\\
\quad five-fold ensembles & & 0.816 / 8.6 & 0.846 / 5.9 & -- & -- & -- & --\\
\midrule
\multicolumn{8}{l}{\textit{MRI}}\\
SynthRAD MRI & 52 & 0.817 / 7.3 & \textbf{0.822} / 6.8 & 0.820 / 10.0 & -- & 0.667$^{\dagger\ddagger}$ / 37.6 & 0.803$^{\ddagger}$ / 11.0\\
AMOS-MR & 20 & 0.808 / 6.3 & 0.816 / 5.5 & 0.791$^{\dagger\ddagger}$ / 6.5 & -- & \textbf{0.818} / 7.4 & 0.787$^{\dagger\ddagger}$ / 7.7\\
TotalSeg.-MR & 120 & 0.732$^{\ddagger}$ / 13.8 & \textbf{0.768}$^{\dagger}$ / 10.6 & overlap & -- & 0.696 / 22.8 & 0.730 / 15.0\\
TotalSeg.\ v3 MRI (external) & 253 & 0.758$^{\ddagger}$ / 15.0 & 0.778$^{\dagger}$ / 11.7 & \textbf{0.843}$^{\dagger\ddagger}$ / 8.8 & -- & 0.718$^{\dagger\ddagger}$ / 45.4 & 0.812$^{\dagger\ddagger}$ / 9.9\\
\quad five-fold ensembles & & 0.780 / 11.7 & 0.795 / 10.0 & -- & -- & -- & --\\
\bottomrule
\end{tabular}
}
\end{table}

\subsubsection{CBCT}

\textbf{Real CBCT.} On \res{synthradcbct.cases} SynthRAD CBCT scans (\res{synthradcbct.labels} labels), IMPACT-Seg achieved mean Dice of \res{synthradcbct.ims25.dice} (2.5D) and \res{synthradcbct.ims3d.dice} (3D), compared with \res{synthradcbct.nv.dice} for the strongest competitor, NV-Segment-CTMR (\res{synthradcbct.nv.p}). IMPACT-Seg also had the lowest surface distances. The advantage held across anatomical regions and was largest in the pelvis; fewer partial failures were reflected in a narrower Dice distribution (Figure~\ref{fig:benchmarks}).

\textbf{Simulated CBCT.} On \res{tsrtk.cases} TS\_RTK volumes (\res{tsrtk.labels} labels), Dice reached \res{tsrtk.ims25.dice} (2.5D) and \res{tsrtk.ims3d.dice} (3D), compared with \res{tsrtk.ts.dice} for TotalSegmentator, the strongest competitor. Although competitors had seen the source CTs, IMPACT-Seg retained its advantage across structure-size strata on both CBCT benchmarks (Figure~\ref{fig:structure_size}).

\textbf{External radiotherapy CBCT.} Table~\ref{tab:pancreas_cbct} summarizes the results on the pancreatic radiotherapy cohort. On planning CT, automatic methods achieved comparable GI Dice (0.828 for both IMPACT-Seg 3D and TotalSegmentator, and 0.817 for NV-Segment-CTMR; $p\geq0.47$). On daily CBCT, evaluated on identical voxels across methods, IMPACT-Seg 3D achieved a GI Dice of 0.735 [0.70, 0.77] and an HD95 of 9.5~mm, outperforming TotalSegmentator (0.634, 14.5~mm) and NV-Segment-CTMR (0.536, 20.7~mm; both $p<0.001$). The GI Dice decrease from planning CT to CBCT was 0.093 for IMPACT-Seg 3D, compared with 0.194 and 0.281 for the respective competitors, although these differences also reflect inter-fraction anatomical changes and independent delineations. Rigid propagation of planning contours achieved a lower GI Dice of 0.690 ($p<0.001$) with a similar HD95 of 9.8~mm. However, propagation performed better on the individual small-bowel and stomach-duodenum structures (Dice 0.579 and 0.702, versus 0.477 and 0.690 for IMPACT-Seg 3D), partly reflecting the dataset's duodenal delineation convention. On the ten CBCTs independently contoured by two observers, IMPACT-Seg 3D achieved a GI Dice of 0.795 against the observer contours, below the inter-observer agreement of 0.847 ($p=0.01$).

\begin{table}[!tbp]
\centering
\footnotesize
\caption{External clinical CBCT (MSKCC pancreatic radiotherapy, TCIA Pancreatic-CT-CBCT-SEG), scored against the expert contours of each image inside the volume of interest and the CBCT field of view. GI: union of the small bowel and the stomach with duodenum, independent of the duodenal boundary, which differs between the dataset (proximal duodenum with the stomach, distal duodenum with the small bowel) and the compared label spaces (whole duodenum); Dice with patient-level bootstrap 95\% CI and HD95. SB, SD: Dice of the small bowel and of the stomach with duodenum as defined in the dataset. Planning CT: the planning CT of the same patient scored on the same voxels as each CBCT; drop: planning-CT minus CBCT GI Dice. Observer CBCT: raw CBCTs contoured by two independent observers, each method compared with both. Best automatic method in bold.}
\label{tab:pancreas_cbct}
\setlength{\tabcolsep}{3.5pt}
\fittable{%
\begin{tabular}{lccccccc}
\toprule
 & \multicolumn{4}{c}{Daily CBCT (80)} & \multicolumn{2}{c}{Planning CT (80)} & Observer CBCT (10)\\
\cmidrule(lr){2-5}\cmidrule(lr){6-7}\cmidrule(lr){8-8}
Method & GI Dice [95\% CI] & GI HD95 (mm) & SB & SD & GI Dice & Drop & GI Dice\\
\midrule
IMPACT-Seg 2.5D & 0.713 [0.68, 0.74] & 9.8 & 0.450 & 0.677 & 0.806 & 0.094 & 0.762\\
IMPACT-Seg 3D & 0.735 [0.70, 0.77] & 9.5 & 0.477 & 0.690 & 0.828 & 0.093 & 0.795\\
IMPACT-Seg 2.5D, ensemble & 0.719 [0.68, 0.75] & 9.8 & 0.459 & 0.683 & 0.811 & 0.092 & 0.772\\
IMPACT-Seg 3D, ensemble & \textbf{0.742} [0.71, 0.77] & \textbf{9.2} & 0.483 & 0.700 & \textbf{0.833} & 0.091 & \textbf{0.795}\\
TotalSegmentator & 0.634 [0.59, 0.67] & 14.5 & 0.374 & 0.599 & 0.828 & 0.194 & 0.748\\
NV-Segment-CTMR & 0.536 [0.48, 0.59] & 20.7 & 0.280 & 0.535 & 0.817 & 0.281 & 0.652\\
\midrule
Planning-CT contours, rigidly propagated & 0.690 [0.67, 0.71] & 9.8 & 0.579 & 0.702 & -- & -- & --\\
Second observer & -- & -- & -- & -- & -- & -- & 0.847\\
\bottomrule
\end{tabular}
}
\end{table}

\subsubsection{CT}

\textbf{Held-out VISCERAL CT.} On \res{visceral.cases} cases, Dice was \res{visceral.ims25.dice} (2.5D) and \res{visceral.ims3d.dice} (3D), compared with \res{visceral.ts.dice} for TotalSegmentator and \res{visceral.nv.dice} for NV-Segment-CTMR. The 3D network did not differ significantly from TotalSegmentator (\res{visceral.ts.pthreed}); label-wise results are provided in Section~\ref{app:labels}.

\textbf{Vertebral and rib identification.} Errors in the numbering of ribs and vertebrae explain a substantial part of the CT difference: merged-vertebra Dice on VISCERAL was \res{ident.ims25.vertebrae.merged} and \res{ident.ims3d.vertebrae.merged}, compared with \res{ident.ts.vertebrae.merged} for TotalSegmentator, whereas correct level identification was \res{ident.ims25.vertebrae.correct}, \res{ident.ims3d.vertebrae.correct}, and \res{ident.ts.vertebrae.correct}, respectively. Similar patterns occurred for ribs, simulated CBCT, and external CT. Excluding vertebrae and ribs narrowed the CT gap while preserving the CBCT advantage (Table~\ref{tab:novertrib}).

\textbf{External TotalSegmentator v3 CT.} On \res{tsvthreect.cases} cases, IMPACT-Seg Dice was \res{tsvthreect.ims25.dice} (2.5D) and \res{tsvthreect.ims3d.dice} (3D), below TotalSegmentator (\res{tsvthreect.ts.dice}) and CADS (\res{tsvthreect.cads.dice}; both \res{tsvthreect.ts.p}). The 3D ensemble improved Dice to \res{tsvthreect.ims3dens.dice}; deficits were largest below 10~ml. TotalSegmentator Dice exceeded 0.99 for 9.7\% of patient-structure pairs (1714/17\,740), versus 0.5\% for CADS and none for IMPACT-Seg. This suggests possible retention of automatic outputs in the references, potentially favouring their generating framework, but does not establish provenance or explain the full gap. Potential training overlap also limits independence.

\subsubsection{MRI}

\textbf{SynthRAD MRI.} IMPACT-Seg achieved Dice of \res{synthradmri.ims25.dice} (2.5D) and \res{synthradmri.ims3d.dice} (3D), comparable to TotalSegmentator (\res{synthradmri.ts.dice}) and MRSegmentator (\res{synthradmri.mrseg.dice}), and above NV-Segment-CTMR (Table~\ref{tab:main_benchmarks}).

\textbf{AMOS-MR and TotalSegmentator-MR.} Dice reached \res{amos.ims25.dice}/\res{amos.ims3d.dice} (2.5D/3D) on AMOS-MR and \res{tsmr.ims25.dice}/\res{tsmr.ims3d.dice} on TotalSegmentator-MR. IMPACT-Seg was competitive on AMOS-MR and exceeded the evaluated competitors on TotalSegmentator-MR; TotalSegmentator MRI was excluded from its own training cohort.

\textbf{External TotalSegmentator v3 MRI.} On \res{tsvthreemri.cases} axial volumes, Dice was \res{tsvthreemri.ims25.dice} (2.5D) and \res{tsvthreemri.ims3d.dice} (3D), below TotalSegmentator~2.4 (\res{tsvthreemri.ts.dice}) and MRSegmentator (\res{tsvthreemri.mrseg.dice}). The 3D ensemble reached \res{tsvthreemri.ims3dens.dice}. 
The evaluated volumes combine 137 whole-body examinations from public collections of the NCI Imaging Data Commons and 116 examinations from University Hospital Basel, the source of the TotalSegmentator MRI dataset. Whole-body acquisitions are under-represented in our MRI training sources, whereas MRSegmentator was trained on a large whole-body Dixon cohort. Deficits were concentrated in small structures and a minority of partially failed cases (Figure~\ref{fig:structure_size}).

\subsubsection{All 126 Structures}

\begin{figure*}[!tbp]
\centering
\includegraphics[width=\textwidth]{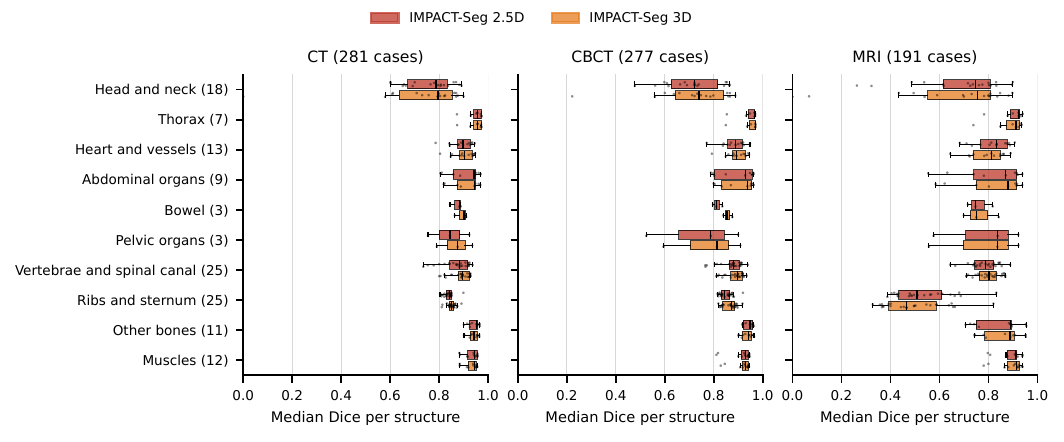}
\caption{Internal out-of-fold Dice of the 126 structures grouped by anatomical region (number of structures in parentheses), per modality and network: each dot is the median Dice of one structure over the test cases whose reference contains it (a missed structure counts as 0), and boxes summarize the structures of a group. For MRI, 50 of the 140 TotalSegmentator-MR test cases are drawn so that this cohort does not dominate. The Dice of every structure is given in Figures~\ref{fig:internal_labels}-\ref{fig:internal_labels_skeleton}.}
\label{fig:internal_groups}
\end{figure*}

The internal out-of-fold evaluation scores every structure present in the reference, including small structures and the individually numbered ribs, rather than only the labels shared with the compared frameworks. Across 838 cases, per-case Dice was 0.845/0.863 on CT, 0.803/0.824 on CBCT, and 0.682/0.704 on MRI (2.5D/3D). Of the 126 structures, 94 (2.5D) and 102 (3D) reached a median Dice of at least 0.8, and 41 and 45 at least 0.9; only the thin anterior eye segments remained below 0.5. Figure~\ref{fig:internal_groups} summarizes performance by region; Figures~\ref{fig:internal_labels}-\ref{fig:internal_labels_skeleton} provide all structures. Simulated CBCT remained close to source CT, whereas real CBCT and MRI were more challenging (cohort details in Figure~\ref{fig:internal}).

% Source: make_paper_figures.py figure_internal_groups (paper_v2/results/internal_oof_{2p5d,3d}/per_label.csv).

\subsubsection{2.5D versus 3D Networks} \label{sec:results_ablation}

On identical out-of-fold cases and patient-label pairs, the 3D network achieved higher mean Dice and lower HD95 on every benchmark (Table~\ref{tab:dim_ablation}). The largest Dice gains occurred on real CBCT, external CT, and TotalSegmentator-MR (all $p<0.001$ after Holm correction); differences on SynthRAD and AMOS MRI were not significant. Volumetric context particularly improved level identification: correct vertebral numbering rose from \res{ident.ims25.vertebrae.correct} to \res{ident.ims3d.vertebrae.correct} on VISCERAL and from \res{identvthreect.ims25.vertebrae.correct} to \res{identvthreect.ims3d.vertebrae.correct} on external CT. CT-CBCT degradation and CT-MRI agreement remained similar between networks. The 3D network therefore favours accuracy, whereas 2.5D offers approximately twice the CT inference speed (Section~\ref{sec:results_cost}).

\subsubsection{Training Data Ablation} \label{sec:results_data_ablation}

We assessed the contribution of each training modality to segmentation accuracy and cross-modality prediction consistency by progressively expanding the training set.

The all-modality ablation variant scored up to 0.03 Dice below the main fold-1 model (Table~\ref{tab:ablation_data}), which was trained using a longer SGD schedule.

\textbf{Simulated CBCT and transfer to real CBCT.}
Adding real CBCT improved segmentation accuracy on clinical CBCT, while simulated CBCT provided further gains in accuracy and CT-CBCT prediction agreement (Table~\ref{tab:ablation_data}). Improvements were larger on simulated CBCT than on real CBCT, although the latter supports transfer beyond the simulation domain. Since adding simulations also increased CBCT exposure during training, these effects cannot be attributed to simulation alone. Notably, the CT-only variant already outperformed the compared frameworks on real SynthRAD CBCT, indicating that the overall advantage of IMPACT-Seg does not rely solely on CBCT-specific training.

\textbf{Joint training preserves CT and CBCT performance.}
Adding MRI was essential for MRI segmentation (Dice $\leq 0.22$ without MRI), while reducing CT and CBCT Dice by at most 0.012 and leaving CT-to-CBCT degradation nearly unchanged. Thus, joint training enabled segmentation across all three modalities with minimal loss of CT and CBCT performance.

% Generated by paper_v2/experiments/ablation_data/summarize_ablation.py -- do not edit.
\begin{table}[!tbp]
\centering
\small
\caption{Training-data ablation (2.5D network, fold~1, AdamW; checkpoint at 200\,000 steps): training sets of increasing composition, from CT only to the full multimodal set (number of training volumes and share of CBCT among the training slices, which are sampled uniformly), evaluated on the fold-1 cases of each benchmark (same patient--label pairs as Table~\ref{tab:main_benchmarks}) and of the internal test set (126 structures). Dice drop: CT minus CBCT Dice of the same patient; agreement: Dice between the CT and CBCT predictions. $^{*}$: different from the previous column (paired Wilcoxon test, Holm-corrected, $p<0.05$).}
\label{tab:ablation_data}
\setlength{\tabcolsep}{4pt}
\begin{tabular}{lcccc}
\toprule
Metric & CT only & + real CBCT & + sim.\ CBCT & + MRI (full)\\
Training volumes & 624 & 763 & 1248 & 1872\\
CBCT share of training slices & 0\% & 10\% & 45\% & 30\%\\
\midrule
\multicolumn{5}{l}{\textit{CBCT}}\\
SynthRAD CBCT (real), Dice & 0.775 & 0.798$^{*}$ & 0.810$^{*}$ & 0.808$^{*}$\\
Internal real CBCT, Dice & 0.649 & 0.723$^{*}$ & 0.733 & 0.724\\
Simulated CBCT (TS\_RTK), Dice & 0.750 & 0.756$^{*}$ & 0.830$^{*}$ & 0.821$^{*}$\\
Internal simulated CBCT, Dice & 0.688 & 0.701$^{*}$ & 0.781$^{*}$ & 0.770$^{*}$\\
TS\_RTK CT-to-CBCT Dice drop & 0.108 & 0.096$^{*}$ & 0.034$^{*}$ & 0.033\\
TS\_RTK CT--CBCT agreement & 0.808 & 0.823$^{*}$ & 0.902$^{*}$ & 0.897$^{*}$\\
SynthRAD CT-to-CBCT Dice drop & 0.103 & 0.085$^{*}$ & 0.072$^{*}$ & 0.071\\
SynthRAD CT--CBCT agreement & 0.813 & 0.837$^{*}$ & 0.854$^{*}$ & 0.851$^{*}$\\
\midrule
\multicolumn{5}{l}{\textit{CT}}\\
VISCERAL CT, Dice & 0.817 & 0.816 & 0.823 & 0.817$^{*}$\\
TotalSeg.\ v3 CT, Dice & 0.765 & 0.766 & 0.782$^{*}$ & 0.773$^{*}$\\
Internal CT, Dice & 0.821 & 0.822 & 0.825 & 0.818$^{*}$\\
\bottomrule
\end{tabular}
\end{table}

\subsection{Feature-Level Consistency and Downstream Registration}
\label{sec:results_features}

We first examined whether the learned representations capture anatomical correspondence across modalities, and then assessed their utility beyond segmentation through a downstream deformable registration task.

\subsubsection{Cross-Modality Feature Correspondence}
\label{sec:Cross_Modality_results_features}

\begin{figure*}[!tbp]
\centering
\includegraphics[width=\textwidth]{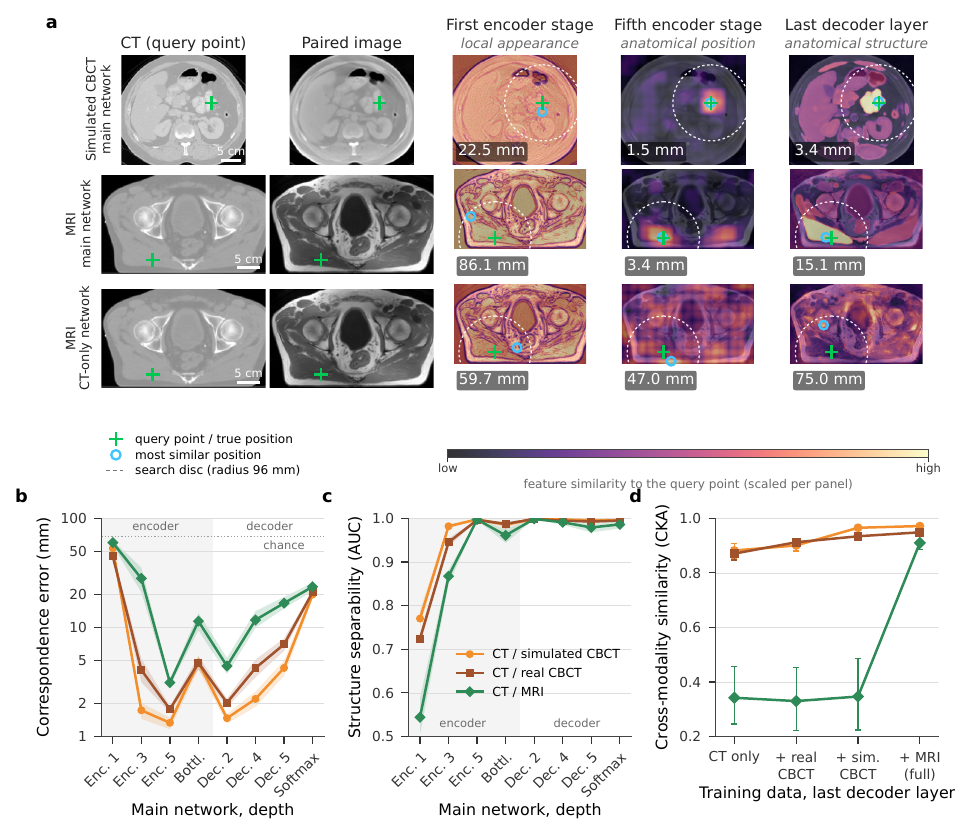}
\caption{Feature-level consistency of the 2.5D networks on paired volumes from patients excluded from all compared models. (a) For a query point on the CT (green cross), similarity of its feature to the features of every position of the same slice of the paired image, at three depths (scaled per panel). The cyan circle marks the most similar position within the 96~mm search disc (dashed), and the number its distance to the true position. Top two rows: main fold-1 network; bottom row: network of the ablation trained on CT only, same query point. The query point was selected from the slice's reference points based on proximity to the networks' cohort-level retrieval statistics at the fifth encoder stage and last decoder layer. (b, c) Main fold-1 network: mean of the patient-wise median retrieval errors within a 96~mm disc (logarithmic scale; dotted: chance, 68~mm) and discrimination between features from different structures (AUC), as a function of depth (shaded: encoder). (d) Similarity of the last-decoder features across modalities (linear CKA) for the networks of the training-data ablation. Bands and error bars: patient-level bootstrap 95\% confidence intervals; simulated CBCT 48, real CBCT 57 and MRI 11 patients.}
\label{fig:features}
\end{figure*}

\textbf{Variation across depth.}
Early encoder features were sensitive to local appearance and poorly localized anatomical correspondences, with mean patient-wise median retrieval errors of 45-60~mm at the first stage against a chance level of approximately 68~mm. The fifth encoder stage was most spatially selective: the main fold-1 model achieved mean patient-wise median retrieval errors of 1.3~mm on simulated CBCT, 1.8~mm on real CBCT, and 3.1~mm on MRI (Figure~\ref{fig:features}). Last-decoder features showed high cross-modality cosine similarity (0.96-0.99) but broader responses within individual structures, with retrieval errors of 4.2, 7.0, and 16.7~mm, respectively.

\textbf{Association with training data.}
CT-only encoder retrieval was already accurate for CBCT ($2.0\pm0.6$ and $2.3\pm0.7$~mm for simulated and real CBCT; mean $\pm$ standard deviation across patients), but far from the corresponding position for MRI ($44.0\pm23.9$~mm). Adding simulated CBCT increased last-decoder CT-simulated-CBCT CKA from $0.900\pm0.076$ to $0.965\pm0.036$ and reduced retrieval error from $9.7\pm4.1$ to $4.9\pm2.2$~mm (Table~\ref{tab:feature_consistency}). The same addition improved real-CBCT CKA from $0.912\pm0.053$ to $0.934\pm0.048$ and retrieval from $10.4\pm4.1$ to $8.1\pm3.6$~mm (both $p<0.05$), supporting transfer beyond simulated appearance. Adding MRI increased CT-MRI CKA from $0.346\pm0.235$ to $0.910\pm0.044$ and reduced fifth-encoder retrieval error from $36.8\pm22.8$ to $3.2\pm0.5$~mm; last-decoder retrieval remained coarser at $17.2\pm3.3$~mm. Thus, supervision improved both representational similarity and localization, with a persistent distinction between spatially selective encoder features and broader decoder responses.

\subsubsection{Downstream Task: Multimodal Registration}
\label{sec:results_registration}

To evaluate whether the learned representations are useful beyond segmentation and local feature correspondence, we applied IMPACT-Seg features to deformable multimodal registration on two Learn2Reg benchmarks. The resulting pipelines ranked first on both AbdomenMRCT (Dice 0.917) and ThoraxCBCT (Dice 0.762; Table~\ref{tab:learn2reg}), with the lowest or tied-lowest HD95. Compared with our previous Elastix-IMPACT submissions \citep{boussot2025impact}, Dice increased by 0.04 and 0.11, respectively. ThoraxCBCT landmark TRE remained within the range of the other top entries, while larger SDlogJ values indicated greater variability in local volume change.

These results demonstrate that the multimodal representations learned through segmentation supervision can serve as a semantic similarity measure for a downstream registration task.

\begin{table}[!tbp]
\centering
\caption{Official Learn2Reg leaderboards (retrieved October 2026): our entries (team BreizhReg, highlighted), the next best entries and the organizers' baseline; the previous BreizhReg submission used pretrained-model IMPACT features in Elastix. Mean $\pm$ standard deviation as reported by the platform. TRE: target registration error on landmarks; SDlogJ: standard deviation of the log-Jacobian determinant (lower indicates less variability in local volume change).}
\label{tab:learn2reg}
\small
\fittable{%
\begin{tabular}{llcccc}
\toprule
Task & Team (method) & Dice $\uparrow$ & HD95 (mm) $\downarrow$ & TRE (mm) $\downarrow$ & SDlogJ $\downarrow$\\
\midrule
AbdomenMRCT & \cellcolor{oursblue}\textbf{BreizhReg (IMPACT-Seg, ours), rank 1} & \cellcolor{oursblue}\textbf{0.917 $\pm$ 0.013} & \cellcolor{oursblue}\textbf{2.24} & \cellcolor{oursblue}- & \cellcolor{oursblue}0.201\\
 & Orange, rank 2 & 0.915 $\pm$ 0.002 & 2.73 & - & 0.097\\
 & UUReg, rank 3 & 0.912 $\pm$ 0.013 & \textbf{2.24} & - & \textbf{0.066}\\
 & corrField, organizers & 0.876 $\pm$ 0.017 & 4.40 & - & 0.096\\
 & \cellcolor{oursblue}BreizhReg (Elastix + IMPACT, ours, previous) & \cellcolor{oursblue}0.875 $\pm$ 0.007 & \cellcolor{oursblue}3.11 & \cellcolor{oursblue}- & \cellcolor{oursblue}0.173\\
\midrule
ThoraxCBCT & \cellcolor{oursblue}\textbf{BreizhReg (IMPACT-Seg, ours), rank 1} & \cellcolor{oursblue}\textbf{0.762 $\pm$ 0.033} & \cellcolor{oursblue}\textbf{9.10 $\pm$ 3.76} & \cellcolor{oursblue}13.22 $\pm$ 2.84 & \cellcolor{oursblue}0.458\\
 & Individual participant, rank 2 & 0.732 $\pm$ 0.026 & 26.26 $\pm$ 3.13 & 13.81 $\pm$ 2.59 & \textbf{0.177}\\
 & CARL, rank 3 & 0.732 $\pm$ 0.044 & 11.25 $\pm$ 2.83 & 17.49 $\pm$ 2.76 & 0.732\\
 & \cellcolor{oursblue}BreizhReg (Elastix + IMPACT, ours, previous) & \cellcolor{oursblue}0.652 $\pm$ 0.032 & \cellcolor{oursblue}39.75 $\pm$ 8.20 & \cellcolor{oursblue}13.04 $\pm$ 2.81 & \cellcolor{oursblue}0.182\\
\bottomrule
\end{tabular}
}
\end{table}

\subsection{Detecting Failures Without a Reference} \label{sec:results_uncertainty}

On external CT and MRI datasets, disagreement between the five fold models was the most effective indicator of segmentation failures (AUROC 0.92–0.94, AUPRC 0.73–0.83; Table~\ref{tab:uncertainty_detection}). This performance was not solely explained by structure size, as fold disagreement maintained high AUROC values (0.89–0.95) when evaluated separately across structure-volume categories, compared with 0.57–0.70 for volume alone (Table~\ref{tab:uncertainty_volume}). In contrast, single-model entropy was less effective, particularly for the 3D network (Table~\ref{tab:uncertainty_detection}).

Reviewing the 20\% most uncertain structures captured 75\% and 81\% of failures for 2.5D and 3D, respectively (Table~\ref{tab:uncertainty_triage}). Entirely missed structures remained undetectable: 75 failures for 2.5D and 88 for 3D. Both networks were overconfident at voxel level (Figure~\ref{fig:uncertainty_calibration}); Figure~\ref{fig:uncertainty_examples} illustrates flagged failures. These results support prioritizing review on the evaluated diagnostic cohorts; treatment-CBCT performance and deployment thresholds remain unvalidated.

\begin{table}[!tbp]
\centering
\footnotesize
\caption{Detection of structures with Dice $<$ 0.7 from uncertainty on the external TotalSegmentator v3 cohorts: AUROC and AUPRC with patient-level bootstrap 95\% confidence intervals (10\,000 resamples). Random baseline: AUROC 0.5 and AUPRC equal to the failure prevalence. Structures not predicted at all are excluded (they carry no uncertainty). The volume baseline ranks structures by their predicted volume only (smaller is riskier).}
\label{tab:uncertainty_detection}
\setlength{\tabcolsep}{4pt}
\begin{tabular}{lcccc}
\toprule
 & \multicolumn{2}{c}{CT} & \multicolumn{2}{c}{MRI}\\
\cmidrule(lr){2-3}\cmidrule(lr){4-5}
Signal & 2.5D & 3D & 2.5D & 3D\\
\midrule
Structures ($n$) & 17695 & 17687 & 7922 & 7917\\
Failure prevalence & 16.3\% & 10.5\% & 22.7\% & 18.2\%\\
\midrule
\multicolumn{5}{l}{\textit{AUROC}}\\
Entropy (inside) & 0.83 [0.80, 0.85] & 0.73 [0.68, 0.76] & 0.83 [0.81, 0.84] & 0.60 [0.57, 0.62]\\
Entropy (boundary band) & 0.64 [0.60, 0.67] & 0.55 [0.50, 0.61] & 0.60 [0.57, 0.63] & 0.40 [0.37, 0.43]\\
Fold disagreement (inside) & \textbf{0.93 [0.92, 0.94]} & \textbf{0.94 [0.93, 0.95]} & \textbf{0.93 [0.93, 0.94]} & \textbf{0.92 [0.92, 0.93]}\\
Fold disagreement (band) & 0.85 [0.84, 0.87] & 0.87 [0.85, 0.89] & 0.83 [0.81, 0.84] & 0.83 [0.81, 0.84]\\
2.5D/3D disagreement (inside) & 0.92 [0.91, 0.93] & 0.88 [0.87, 0.89] & 0.90 [0.89, 0.92] & 0.84 [0.83, 0.86]\\
2.5D/3D disagreement (band) & 0.88 [0.87, 0.89] & 0.84 [0.82, 0.86] & 0.84 [0.83, 0.86] & 0.82 [0.81, 0.84]\\
1 $-$ confidence & 0.85 [0.82, 0.87] & 0.72 [0.68, 0.76] & 0.85 [0.83, 0.87] & 0.61 [0.59, 0.64]\\
Volume only (baseline) & 0.77 [0.75, 0.79] & 0.80 [0.78, 0.82] & 0.84 [0.83, 0.86] & 0.85 [0.84, 0.86]\\
\midrule
\multicolumn{5}{l}{\textit{AUPRC}}\\
Entropy (inside) & 0.64 [0.58, 0.70] & 0.47 [0.37, 0.56] & 0.71 [0.67, 0.74] & 0.39 [0.35, 0.43]\\
Entropy (boundary band) & 0.39 [0.32, 0.46] & 0.29 [0.20, 0.38] & 0.41 [0.36, 0.47] & 0.21 [0.18, 0.25]\\
Fold disagreement (inside) & 0.75 [0.71, 0.80] & \textbf{0.73 [0.67, 0.78]} & \textbf{0.83 [0.81, 0.85]} & \textbf{0.76 [0.74, 0.79]}\\
Fold disagreement (band) & 0.60 [0.55, 0.65] & 0.59 [0.52, 0.65] & 0.61 [0.57, 0.64] & 0.54 [0.51, 0.57]\\
2.5D/3D disagreement (inside) & \textbf{0.78 [0.75, 0.81]} & 0.53 [0.45, 0.60] & 0.81 [0.79, 0.83] & 0.61 [0.58, 0.64]\\
2.5D/3D disagreement (band) & 0.67 [0.62, 0.71] & 0.45 [0.39, 0.50] & 0.65 [0.61, 0.69] & 0.53 [0.50, 0.56]\\
1 $-$ confidence & 0.66 [0.61, 0.72] & 0.45 [0.35, 0.54] & 0.74 [0.71, 0.77] & 0.39 [0.35, 0.43]\\
Volume only (baseline) & 0.46 [0.40, 0.52] & 0.44 [0.37, 0.51] & 0.68 [0.65, 0.71] & 0.63 [0.60, 0.66]\\
\bottomrule
\end{tabular}
\end{table}

\FloatBarrier

\subsection{Computational Efficiency}
\label{sec:results_cost}

IMPACT-Seg is implemented with KonfAI, which enables fast, memory-efficient, out-of-core inference and supports deployment across a wide range of hardware configurations, including memory-constrained systems. End-to-end inference on an RTX A6000 required 8.0~s for a whole-body CT with 2.5D and 14.8~s with 3D; CBCT and MRI required approximately 2~s (Table~\ref{tab:runtime}). On CT, the evaluated competitor configurations required 82~s for NV-Segment-CTMR, 328~s for TotalSegmentator, and 114~s for one CADS part model or 555~s for all nine. These measurements include preprocessing, patch inference, reconstruction, and output writing, using 30 volumes per modality with the models loaded once per run. Peak IMPACT-Seg GPU memory remained below 6~GB for one fold and 9~GB for the five-fold ensemble. Detailed network-only timings and competitor memory use are reported in Table~\ref{tab:runtime}.

\begin{table}[!tbp]
\centering
\small
\caption{Inference cost per volume on one NVIDIA RTX A6000 (48~GB), Intel Xeon Gold 6254 @ 3.10GHz (72 threads), 376~GB RAM, for lots of 30 whole-body CT (VISCERAL), 30 real CBCT and 30 MRI volumes (SynthRAD), each processed in one run with the model loaded once. Total: wall time of the run divided by the number of volumes (loading, pre-processing, inference, post-processing and writing); network: sliding-window inference only (patches and their fusion); GB: peak GPU memory of the method's processes. IMPACT-Seg predicts all 126 structures with one network per fold (2.5D with an inference batch of 32; ensemble: the five folds, softmax averaged); the competitors run in the configuration of the evaluation (shared labels), TotalSegmentator and MRSegmentator chaining several nnU-Net models; CADS is also reported with all nine part models, as needed for its full label set.}
\label{tab:runtime}
\fittable{%
\begin{tabular}{lrrrrrrrrr}
\toprule
 & \multicolumn{3}{c}{CT} & \multicolumn{3}{c}{CBCT} & \multicolumn{3}{c}{MRI}\\
\cmidrule(lr){2-4}\cmidrule(lr){5-7}\cmidrule(lr){8-10}
Method & Total (s) & Network (s) & GB & Total (s) & Network (s) & GB & Total (s) & Network (s) & GB\\
\midrule
IMPACT-Seg 2.5D & 8.0 & 4.4 & 5.7 & 1.8 & 0.3 & 5.6 & 2.0 & 0.6 & 5.6\\
IMPACT-Seg 3D & 14.8 & 11.9 & 4.4 & 2.3 & 0.8 & 4.2 & 2.6 & 1.1 & 4.2\\
IMPACT-Seg 2.5D, ensemble & 32.9 & 29.6 & 8.5 & 3.4 & 1.8 & 8.3 & 5.1 & 3.5 & 8.3\\
IMPACT-Seg 3D, ensemble & 54.5 & 51.5 & 8.2 & 6.1 & 4.3 & 7.8 & 7.5 & 5.9 & 7.8\\
TotalSegmentator & 327.9 & 14.4 & 19.6 & 67.2 & 4.7 & 3.4 & 52.3 & 3.6 & 3.7\\
CADS, one part model & 114.1 & 7.8 & 18.6 & 15.0 & 1.0 & 7.9 & -- & -- & --\\
CADS, nine part models & 554.5 & 72.1 & 28.0 & 77.4 & 12.9 & 10.7 & -- & -- & --\\
NV-Segment-CTMR & 82.4 & 15.4 & 42.4 & 9.1 & 1.9 & 12.8 & 4.6 & 2.4 & 11.7\\
MRSegmentator & -- & -- & -- & -- & -- & -- & 24.4 & 5.3 & 6.7\\
\bottomrule
\end{tabular}
}
\end{table}

\FloatBarrier
\section{Discussion} \label{sec:discussion}

IMPACT-Seg combines harmonized anatomical labels, joint multimodal training, and physics-based CBCT simulation to enable dense segmentation across CT, CBCT, and MRI. Our results demonstrate improved robustness to CT-to-CBCT appearance changes and greater consistency between predictions across modalities compared with existing methods. Feature analyses further reveal that this shared anatomical supervision promotes anatomical correspondence in the learned representations, without explicit feature-alignment objectives. These representations were also successfully integrated into IMPACT-Reg, supporting improved performance on multimodal deformable registration benchmarks.

\begin{figure*}[!tbp]
\centering
\includegraphics[width=\textwidth]{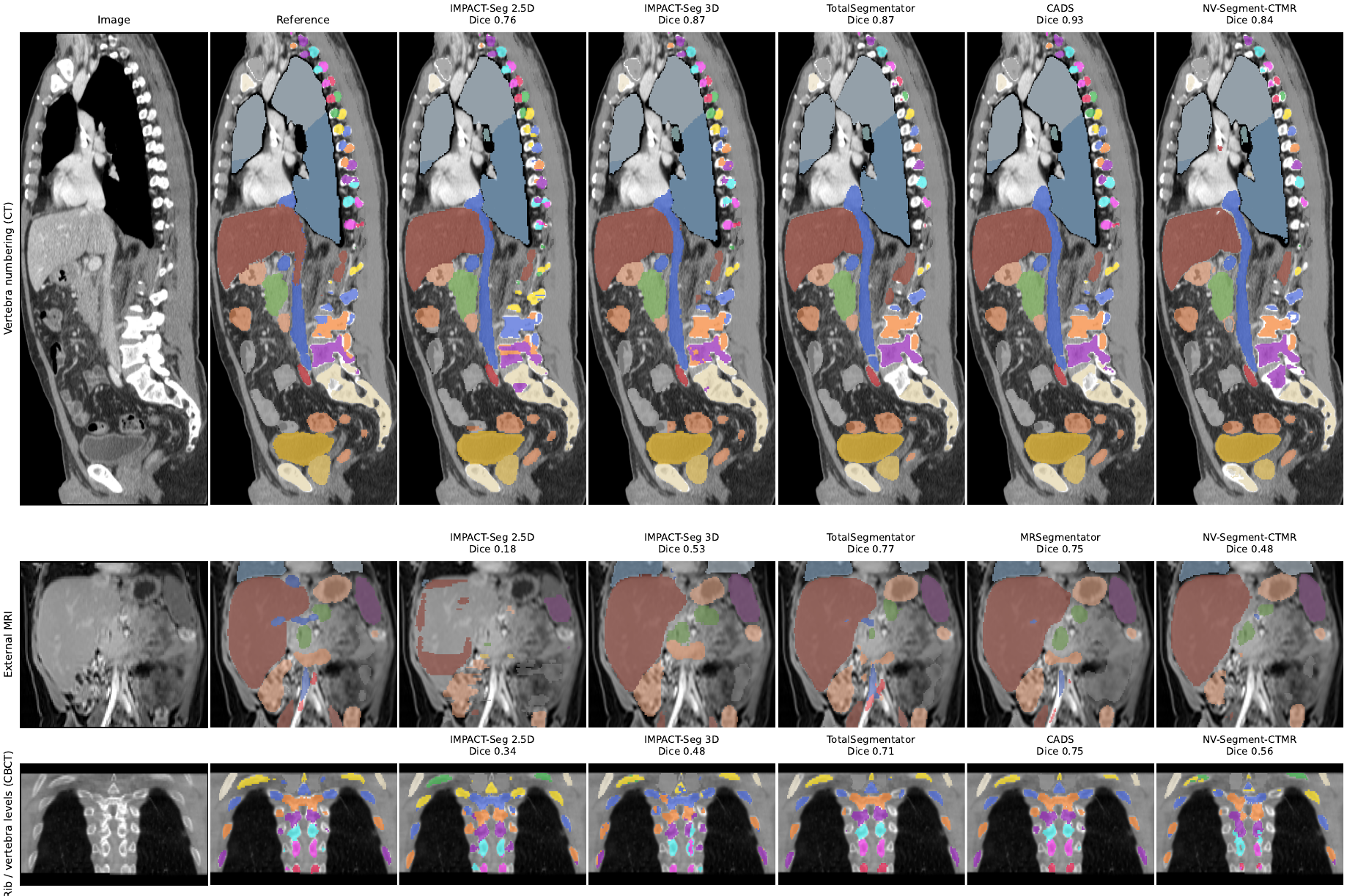}
\caption{Failure cases, selected on all five folds by the mean of the 2.5D and 3D networks. Top: held-out CT with the lowest rate of correctly identified vertebral levels; vertebrae are coloured by level, so that a level shift appears as a colour change: the 2.5D network shifts the lumbar levels, which the 3D network identifies correctly. Middle: external MRI with the lowest Dice (partial liver miss of the 2.5D network, recovered by the 3D network). Bottom: real thoracic CBCT with the lowest Dice; ribs and vertebrae are delineated but shifted by one level by both networks. Titles give the mean Dice of the case.}
\label{fig:failure}
\end{figure*}

\textbf{External radiotherapy validation.} On an independent pancreatic radiotherapy cohort, IMPACT-Seg 3D performed comparably to other automatic methods on planning CT but showed substantially less degradation on daily CBCT, with approximately half the GI Dice decrease observed for TotalSegmentator. It also outperformed rigid contour propagation for the combined gastrointestinal region, despite breath-hold acquisitions favouring propagation. However, propagation achieved higher Dice for individual structures, partly reflecting the dataset's duodenal delineation convention, while IMPACT-Seg agreement with expert contours remained below inter-observer agreement. These findings support automatic segmentation as a complement to contour propagation in adaptive radiotherapy, with expert review remaining necessary.

\textbf{Anatomical consistency without explicit feature alignment.} The feature analysis distinguishes cross-modality similarity from spatial discrimination. Deep encoder features identify corresponding positions more precisely, whereas last-decoder features exhibit high cross-modality similarity but broader responses within anatomical structures. This distinction is particularly relevant for registration, where features must provide both consistent anatomical correspondence and sufficiently localized similarity to guide deformation. Joint multimodal supervision improved both properties, with MRI correspondence particularly dependent on MRI exposure. The downstream registration experiments further demonstrated that these representations can support multimodal deformable registration, extending their utility beyond segmentation. However, the evaluated pipelines used last-decoder features as a semantic similarity objective, and the relative contribution of the learned features and other registration components was not isolated. The sharper localization of deep encoder features also makes them promising candidates for future registration objectives. Together, these findings complement appearance-robust pretraining in Anatomix \citep{dey2025learning} and semantic registration losses in IMPACT \citep{boussot2025impact}, showing that useful multimodal representations can emerge from dense segmentation supervision without an explicit feature-alignment objective.

\textbf{Dense anatomical coverage.} The shared label space enables consistent delineation of 126 anatomical structures across CT, CBCT, and MRI, including cardiac substructures, muscles, and head-and-neck organs at risk (Figure~\ref{fig:coverage}). While comparisons with existing frameworks are limited to the structures supported by each method (39–104 labels), our internal evaluation covers all 126 structures, providing a more comprehensive assessment of the framework's anatomical coverage.

\textbf{Improved confidence measures for uncertainty-guided review.} Disagreement between the five fold models proved effective in identifying segmentation failures, even when accounting for differences in anatomical structure size. This approach could help prioritize structures requiring manual review and correction~\citep{zenk2025failure,wahid2024uncertainty}, although it requires predictions from multiple models. In contrast, uncertainty estimates derived from single-model confidence were less reliable. Further validation in clinical radiotherapy workflows is needed to determine whether uncertainty-guided review can improve efficiency and segmentation reliability.

\subsection{Limitations and Future Work}

\textbf{External MRI generalization.} Small structures and partially failed cases account for much of the external MRI gap. Limited protocol coverage is a plausible contributor, although protocol effects were not isolated. Broader sampling across sequences, field strengths, and institutions, together with non-axial slab augmentation, merits evaluation.

\textbf{Vertebral and rib numbering.} Most of the residual CT gap with TotalSegmentator arises from level identification rather than boundary delineation. Counting vertebrae and ribs requires cranio-caudal context that the 2.5D input does not provide; the 3D network reduces this error, and an explicit spine-ordering post-processing step is a natural next improvement.

\textbf{Reference quality and evaluation bias.}
Despite iterative refinement and manual review, the pseudo-labeled references may retain systematic errors inherited from the generating models. Their quality and completeness vary across cohorts and structures, potentially introducing biases in comparisons with existing segmentation frameworks. Future validation should include a stratified expert audit and report performance separately for original annotations, pseudo-labeled structures, and manually corrected structures.

\textbf{Clinical translation.} The paper quantifies segmentation accuracy, cross-modality consistency and registration performance, but not the clinical consequence of residual segmentation errors. A prospective or retrospective radiotherapy study should now measure manual correction time, contour acceptability and dosimetric impact in adaptive workflows.

\FloatBarrier
\section{Conclusion}
\label{sec:conclusion}

IMPACT-Seg combines harmonized anatomical supervision, joint multimodal training, and physics-based CBCT simulation to segment 126 anatomical structures across CT, CBCT, and MRI using shared network parameters. Our results demonstrate improved robustness to CBCT appearance changes and greater consistency of anatomical predictions across modalities. Beyond segmentation, the learned representations capture anatomical correspondences that can be leveraged to improve multimodal deformable registration with IMPACT-Reg. IMPACT-Seg also provides a fast and memory-efficient segmentation tool, with the 2.5D model segmenting a whole-body CT in just 8 seconds. Uncertainty-based failure detection further supports its potential for targeted expert review in adaptive radiotherapy. The publicly available datasets, models, and code provide a foundation for further multimodal research, while future clinical studies should assess contour acceptability, manual correction time, and dosimetric impact.

\FloatBarrier
\section*{Data and Code Availability}
The IMPACT-Seg dataset (8171 CT, CBCT and MRI volumes with their reference label maps and label definitions) is publicly available at \url{https://huggingface.co/datasets/HCedric/IMPACTSeg}. The 119 VISCERAL cases in the training collection will not be redistributed, in accordance with their license. The trained models (2.5D and 3D networks, five folds each, with ready-to-run KonfAI prediction configurations) are available at \url{https://huggingface.co/HCedric/IMPACTSeg}. The training and inference code, the five data splits and the mapping of split case identifiers to released cases are available at \url{https://github.com/HEMONCedric/IMPACTSeg}. The source datasets retain their original licenses.

\section*{CRediT Authorship Contribution Statement}
\textbf{C{\'e}dric H{\'e}mon:} Conceptualization, Methodology, Software, Data curation, Formal analysis, Investigation, Validation, Visualization, Writing - original draft, Writing - review \& editing.

\textbf{Valentin Boussot:} Software, Investigation, Validation, Writing - original draft, Writing - review \& editing.

\textbf{Matthew McCormick:} Supervision, Validation, Writing - original draft, Writing - review \& editing.

\textbf{Lukas Zimmermann:}  Investigation, Validation, Writing - original draft, Writing - review \& editing.

\textbf{Christopher Kurz:} Supervision, Validation, Writing - original draft, Writing - review \& editing.

\textbf{Guillaume Landry:} Supervision, Validation, Writing - original draft, Writing - review \& editing.

\section*{Declaration of Competing Interest}
The authors declare that they have no known competing financial interests or personal relationships that could have appeared to influence the work reported in this paper. 

\section*{Acknowledgment}
This work was supported by the Bavarian Cancer Research Center (BZKF) Lighthouse ‘Image guidance in local therapies’.

\section*{Declaration of Generative AI and AI-Assisted Technologies in the Writing Process}
During the preparation of this work, the authors used ChatGPT (OpenAI) to improve the structure and language of the manuscript and to assist in writing analysis code. After using these tools, the authors reviewed and edited the content as needed and take full responsibility for the content of the publication.

\bibliographystyle{elsarticle-num-names}
\begingroup\small\setlength{\bibsep}{2pt}\interlinepenalty=10000
\bibliography{biblio}
\endgroup
\clearpage
\setcounter{section}{0}\setcounter{subsection}{0}\setcounter{table}{0}\setcounter{figure}{0}
\renewcommand{\thesection}{S\arabic{section}}
\renewcommand{\thetable}{S\arabic{table}}
\renewcommand{\thefigure}{S\arabic{figure}}
\renewcommand{\theHsection}{S\arabic{section}}
\renewcommand{\theHtable}{S\arabic{table}}
\renewcommand{\theHfigure}{S\arabic{figure}}
\begin{center}
{\Large\bfseries Supplementary Material}\\[4pt]
{\large IMPACT-Seg: consistent dense anatomical segmentation across CT, CBCT and MRI for radiotherapy}
\end{center}
\raggedbottom
\suppressfloats[t]  % keep the first supplementary tables below the title

\begingroup
\raggedbottom
\FloatBarrier
\Needspace{12\baselineskip}
\section{Benchmark Construction} \label{app:benchmarks}

\begin{table}[H]
\centering
\caption{Software and pretrained weights used in the comparisons. Code versions and weight releases are reported separately. Linked commit and model revisions identify the evaluated implementations.}
\label{tab:comparator_versions}
\small
\setlength{\tabcolsep}{4pt}
\begin{tabularx}{\linewidth}{@{}>{\raggedright\arraybackslash}p{0.22\linewidth}>{\raggedright\arraybackslash}p{0.29\linewidth}>{\raggedright\arraybackslash}X@{}}
\toprule
Framework & Code & Pretrained weights\\
\midrule
CADS & 1.0.2; commit \href{https://github.com/murong-xu/CADS/commit/29d90ca04216cd2ea8782a5dfb9e4893b54ba829}{\texttt{29d90ca04216}} & \href{https://github.com/murong-xu/CADS/releases/tag/cads-model_v1.0.0}{\texttt{cads-model\_v1.0.0}}\\
MRSegmentator & 2.0.0 & Base 1.2 (release v1.2.0); five-fold ensemble, \texttt{checkpoint\_final.pth}\\
NV-Segment-CTMR & Commit \href{https://github.com/NVIDIA-Medtech/NV-Segment-CTMR/commit/e04af885a0d39c939b5b550115f23ac8feca95d2}{\texttt{e04af885a0d3}}; MONAI bundle 0.0.1s & \texttt{model.pt}, model revision \href{https://huggingface.co/nvidia/NV-Segment-CTMR/tree/4fb8b4a6b253}{\texttt{4fb8b4a6b253}}\\
TotalSegmentator CT/CBCT & 2.18.0 & CT weights unchanged since 2.0.0; \texttt{total} and \texttt{head\_glands\_cavities} tasks\\
TotalSegmentator MRI & 2.4.0 (separate environment) & MRI weights distributed with 2.4.0\\
\bottomrule
\end{tabularx}
\begin{minipage}{\linewidth}
\footnotesize\vspace{3pt}
CADS CT logs document tasks 551--555; the nine-model runtime includes tasks 551--559. CADS weights are identified by the release hard-coded in the recorded commit; the local checkpoint files were unavailable for hashing. Full commits, available SHA-256 identifiers, and inference environments are recorded in \texttt{COMPETITOR\_VERSIONS.md} in the source archive.
\end{minipage}
\end{table}

\textbf{External TotalSegmentator v3 cohorts.} MRI volumes acquired in coronal or sagittal slabs (largest voxel dimension not along the cranio-caudal axis) were excluded for every method, leaving 391 axial or isotropic MRI volumes. Because IMPACT-Seg targets adult radiotherapy imaging from the head and neck to the pelvis, examinations outside this scope were excluded using identical metadata- and reference-based criteria for all methods: paediatric patients (age below 14 years or paediatric protocol), limb and joint examinations, volumes whose reference contains less than 50~ml of structures other than limb bones and muscles, and skull-only volumes in which the brain makes up at least half of the reference volume (without a shared brain label, these volumes would otherwise be scored only on the upper cervical spine). Final comparisons also required an available output from every compared method. This yielded 247 of 712 CT and 253 of 391 MRI volumes (CT: 264 paediatric, 36 limb, 165 skull-only; MRI: 119 paediatric, 11 limb-only, 3 skull-only, 5 without NV-Segment-CTMR output).

\clearpage
\endgroup
\section{Implementation Details} \label{app:implementation}

\textbf{Geometry audits.} In addition to checking image-label geometry, cohort-level heuristics flagged suspicious direction matrices, rare orientation codes, unexpected header scaling, and plausible permutations of the spatial axes. Such checks cannot detect an image and mask that share the same erroneous geometry. Volumes with slices thinner than 0.5~mm were excluded; those thicker then 0.5~mm and thinner than 1.5~mm were resampled to 1.5~mm along the cranio-caudal axis. Anatomical numbering and label-volume outliers was used to select cases for manual review.

\textbf{CBCT simulation.} CT Hounsfield units were converted to linear attenuation coefficients before projection. After FDK reconstruction, intensities were mapped back to a HU-like scale and resampled onto the source CT grid. Acquisition geometries were sampled from priors representative of common on-board devices. Physics-based degradations simulated during projection generation and image reconstruction were Poisson noise, scatter, detector saturation and blur, low-frequency shading, field-of-view truncation, and isocentre miscentring, with occasionally coarser reconstruction spacing. Parameters remained within clinically plausible ranges to avoid sparse-view aliasing \citep{bian2010sparseview}. Without 4D CT, respiratory motion was synthesized by displacing the lungs, heart, and upper abdominal organs of the static CT, mainly cranio-caudally (up to about 11~mm), using smoothed reference masks. This yielded four to eight phases around the source CT. Projections were interpolated between phases along a quasi-periodic breathing signal. Dental, sternal, hip, and spinal implants were positioned using anatomical masks to generate metal artefacts.

\textbf{Architecture and augmentation.} The 2.5D encoder has six stages with 32, 64, 128, 256, 320, and 512 features and 1, 3, 4, 6, 6, and 6 residual blocks, respectively. Brightness, contrast, Gaussian noise, blur, and low-resolution augmentations were each applied with probability 0.15. Blur was two-dimensional or three-dimensional according to the network. Foreground-centered sampling and omission of spatial flips are specified in Section~\ref{sec:training}.

\textbf{Training-data ablation.} The four 2.5D fold-1 variants used AdamW with a learning rate warmed up to $5\times10^{-4}$ over 25\,000 steps, followed by cosine decay scheduled over 500\,000 steps; other settings matched the main training configuration. The CT-only and full-multimodal variants reached their lowest validation loss at 200\,000 steps, with subsequent validation Dice changes below 0.01. The other variants were therefore evaluated at the same 200\,000-step checkpoint, following the same learning-rate trajectory. Validation included only the modalities present in each training set. Uniform slice sampling changed modality exposure as data were added; this is not a control at fixed sample count or modality frequency.

\textbf{Comparator provenance.} The TotalSegmentator CT model was trained on 1559 subjects and its weights are unchanged since release 2.0.0. The 1559 training subjects exceed the 1228 public version~2 cases, but cohort sizes alone cannot establish or bound case-level overlap with the new CT cohort. MRI used release 2.4.0 weights trained on 495 subjects; later releases trained on 1088 subjects (2.5.0 onwards) were excluded because of potential overlap. Software, checkpoint, and environment identifiers are recorded in Table~\ref{tab:comparator_versions}.

\clearpage
\section{Detailed Benchmark Results} \label{app:detailed}

Tables~\ref{tab:synthrad_cbct_framework_overview}--\ref{tab:ts_v3_mri_framework_overview} detail the comparison of Table~\ref{tab:main_benchmarks} with all metrics, and Table~\ref{tab:novertrib} repeats the CT and CBCT comparison without the vertebra and rib labels.

\begin{table}[H]
\centering
\small
\caption{Framework-level comparison on the curated SynthRADSegCBCT subset (139 cases, 89 evaluated labels). IMPACT-Seg is evaluated out of fold: each case is predicted by a fold model that never saw the patient. Mean $\pm$ standard deviation of case-level means over the evaluated reference structures (Section~\ref{sec:evaluation}). Best values per group in bold; $^{\dagger}$/$^{\ddagger}$ overall difference with IMPACT-Seg 2.5D/3D significant (paired Wilcoxon, Holm-corrected $p<0.05$).}
\label{tab:synthrad_cbct_framework_overview}
\fittable{%
\begin{tabular}{lccc}
\toprule
Method & Dice & ASD (mm) & HD95 (mm)\\
\multicolumn{4}{l}{\textit{Abdomen}}\\
CADS & 0.634 ± 0.231 & 4.352 ± 3.791 & 15.259 ± 11.630\\
NV-Segment-CTMR & 0.627 ± 0.151 & 9.362 ± 18.084 & 25.244 ± 29.838\\
TotalSegmentator & 0.588 ± 0.263 & 6.608 ± 6.742 & 19.361 ± 15.989\\
IMPACT-Seg 2.5D & 0.763 ± 0.076 & 2.570 ± 1.168 & 9.484 ± 4.251\\
IMPACT-Seg 3D & \textbf{0.791 ± 0.059} & \textbf{2.138 ± 0.818} & \textbf{8.084 ± 3.744}\\
\midrule
\multicolumn{4}{l}{\textit{Thorax}}\\
CADS & 0.616 ± 0.204 & 7.141 ± 11.913 & 21.700 ± 27.851\\
NV-Segment-CTMR & 0.642 ± 0.094 & 3.996 ± 2.778 & 15.468 ± 12.125\\
TotalSegmentator & 0.534 ± 0.211 & 8.638 ± 13.342 & 24.854 ± 30.211\\
IMPACT-Seg 2.5D & 0.694 ± 0.132 & 2.885 ± 2.446 & 10.109 ± 6.780\\
IMPACT-Seg 3D & \textbf{0.764 ± 0.089} & \textbf{1.929 ± 1.757} & \textbf{7.146 ± 5.726}\\
\midrule
\multicolumn{4}{l}{\textit{Pelvis}}\\
CADS & 0.414 ± 0.157 & 6.401 ± 3.266 & 20.234 ± 9.256\\
NV-Segment-CTMR & 0.692 ± 0.165 & 11.304 ± 17.580 & 35.515 ± 42.540\\
TotalSegmentator & 0.391 ± 0.245 & 15.538 ± 15.028 & 37.036 ± 30.306\\
IMPACT-Seg 2.5D & 0.862 ± 0.024 & 1.531 ± 0.345 & 5.237 ± 1.493\\
IMPACT-Seg 3D & \textbf{0.875 ± 0.023} & \textbf{1.294 ± 0.303} & \textbf{4.382 ± 1.371}\\
\midrule
\multicolumn{4}{l}{\textit{Overall}}\\
CADS & 0.489 ± 0.208$^{\dagger}$$^{\ddagger}$ & 5.999 ± 5.152$^{\dagger}$$^{\ddagger}$ & 19.221 ± 13.269$^{\dagger}$$^{\ddagger}$\\
NV-Segment-CTMR & 0.671 ± 0.157$^{\dagger}$$^{\ddagger}$ & 10.001 ± 16.774$^{\dagger}$$^{\ddagger}$ & 30.769 ± 37.930$^{\dagger}$$^{\ddagger}$\\
TotalSegmentator & 0.454 ± 0.259$^{\dagger}$$^{\ddagger}$ & 12.581 ± 13.818$^{\dagger}$$^{\ddagger}$ & 31.355 ± 28.445$^{\dagger}$$^{\ddagger}$\\
IMPACT-Seg 2.5D & 0.819 ± 0.086$^{\ddagger}$ & 1.933 ± 1.162$^{\ddagger}$ & 6.806 ± 3.900$^{\ddagger}$\\
IMPACT-Seg 3D & \textbf{0.842 ± 0.063$^{\dagger}$} & \textbf{1.568 ± 0.832$^{\dagger}$} & \textbf{5.579 ± 3.278$^{\dagger}$}\\
\bottomrule
\end{tabular}%
}
\end{table}

\begin{table}[H]
\centering
\small
\caption{Framework-level comparison on the TS\_RTK simulated CBCT benchmark (\res{tsrtk.cases} cases, 93 evaluated labels). IMPACT-Seg is evaluated out of fold: each case is predicted by a fold model that never saw the patient. Mean $\pm$ standard deviation of case-level means over the evaluated reference structures (Section~\ref{sec:evaluation}). Best values per group in bold; $^{\dagger}$/$^{\ddagger}$ overall difference with IMPACT-Seg 2.5D/3D significant (paired Wilcoxon, Holm-corrected $p<0.05$).}
\label{tab:cbct_sim_framework_overview}
\fittable{%
\begin{tabular}{lccc}
\toprule
Method & Dice & ASD (mm) & HD95 (mm)\\
\multicolumn{4}{l}{\textit{Abdomen}}\\
CADS & 0.765 ± 0.099 & 2.765 ± 2.387 & 9.845 ± 7.758\\
NV-Segment-CTMR & 0.721 ± 0.137 & 4.347 ± 3.929 & 15.049 ± 11.909\\
TotalSegmentator & 0.761 ± 0.088 & 3.078 ± 2.624 & 10.819 ± 8.014\\
IMPACT-Seg 2.5D & 0.843 ± 0.083 & 1.794 ± 1.381 & 6.155 ± 3.905\\
IMPACT-Seg 3D & \textbf{0.861 ± 0.067} & \textbf{1.470 ± 1.145} & \textbf{4.678 ± 3.113}\\
\midrule
\multicolumn{4}{l}{\textit{Thorax}}\\
CADS & 0.686 ± 0.107 & 3.456 ± 1.824 & 12.757 ± 6.140\\
NV-Segment-CTMR & 0.683 ± 0.129 & 3.558 ± 2.662 & 12.661 ± 6.437\\
TotalSegmentator & 0.695 ± 0.091 & 3.868 ± 1.948 & 13.777 ± 5.802\\
IMPACT-Seg 2.5D & 0.812 ± 0.067 & 1.725 ± 0.964 & 5.989 ± 3.079\\
IMPACT-Seg 3D & \textbf{0.842 ± 0.062} & \textbf{1.312 ± 0.905} & \textbf{4.386 ± 3.120}\\
\midrule
\multicolumn{4}{l}{\textit{Overall}}\\
CADS & 0.740 ± 0.108$^{\dagger}$$^{\ddagger}$ & 2.980 ± 2.248$^{\dagger}$$^{\ddagger}$ & 10.753 ± 7.410$^{\dagger}$$^{\ddagger}$\\
NV-Segment-CTMR & 0.710 ± 0.135$^{\dagger}$$^{\ddagger}$ & 4.101 ± 3.598$^{\dagger}$$^{\ddagger}$ & 14.304 ± 10.564$^{\dagger}$$^{\ddagger}$\\
TotalSegmentator & 0.741 ± 0.094$^{\dagger}$$^{\ddagger}$ & 3.324 ± 2.459$^{\dagger}$$^{\ddagger}$ & 11.741 ± 7.516$^{\dagger}$$^{\ddagger}$\\
IMPACT-Seg 2.5D & 0.833 ± 0.079$^{\ddagger}$ & 1.773 ± 1.265$^{\ddagger}$ & 6.103 ± 3.665$^{\ddagger}$\\
IMPACT-Seg 3D & \textbf{0.855 ± 0.066$^{\dagger}$} & \textbf{1.420 ± 1.077$^{\dagger}$} & \textbf{4.587 ± 3.115$^{\dagger}$}\\
\bottomrule
\end{tabular}%
}
\end{table}

\begin{table}[H]
\centering
\scriptsize
\caption{Framework-level comparison on the held-out VISCERAL CT benchmark (unseen patients from the VisceralGC and VisceralGCsilver cohorts) (58 cases, 103 evaluated labels). IMPACT-Seg is evaluated out of fold: each case is predicted by a fold model that never saw the patient. Mean $\pm$ standard deviation of case-level means over the evaluated reference structures (Section~\ref{sec:evaluation}). Best values per group in bold; $^{\dagger}$/$^{\ddagger}$ overall difference with IMPACT-Seg 2.5D/3D significant (paired Wilcoxon, Holm-corrected $p<0.05$).}
\label{tab:ct_eval_framework_overview}
\fittable{%
\begin{tabular}{lccc}
\toprule
Method & Dice & ASD & HD95\\
\midrule
NV-Segment-CTMR & 0.798 ± 0.053$^{\dagger}$$^{\ddagger}$ & 12.025 ± 8.477$^{\dagger}$$^{\ddagger}$ & 41.709 ± 23.997$^{\dagger}$$^{\ddagger}$\\
TotalSegmentator & \textbf{0.853 ± 0.037$^{\dagger}$} & 1.619 ± 0.744$^{\dagger}$ & 7.014 ± 3.418$^{\ddagger}$\\
IMPACT-Seg 2.5D & 0.833 ± 0.043$^{\ddagger}$ & 1.875 ± 0.898$^{\ddagger}$ & 7.439 ± 4.273$^{\ddagger}$\\
IMPACT-Seg 3D & 0.850 ± 0.035$^{\dagger}$ & \textbf{1.523 ± 0.590$^{\dagger}$} & \textbf{6.149 ± 2.962$^{\dagger}$}\\
\bottomrule
\end{tabular}%
}
\end{table}

\begin{table}[H]
\centering
\scriptsize
\caption{Framework-level comparison on the external TotalSegmentator v3 CT cases (absent from v2) (247 cases, 93 evaluated labels). IMPACT-Seg is evaluated out of fold: each case is predicted by a fold model that never saw the patient. Mean $\pm$ standard deviation of case-level means over the evaluated reference structures (Section~\ref{sec:evaluation}). Best values per group in bold; $^{\dagger}$/$^{\ddagger}$ overall difference with IMPACT-Seg 2.5D/3D significant (paired Wilcoxon, Holm-corrected $p<0.05$).}
\label{tab:ts_v3_ct_framework_overview}
\fittable{%
\begin{tabular}{lccc}
\toprule
Method & Dice & ASD & HD95\\
\midrule
CADS & 0.890 ± 0.091$^{\dagger}$$^{\ddagger}$ & 1.146 ± 0.866$^{\dagger}$$^{\ddagger}$ & 4.700 ± 2.937$^{\dagger}$$^{\ddagger}$\\
NV-Segment-CTMR & 0.798 ± 0.171$^{\dagger}$$^{\ddagger}$ & 8.679 ± 11.536$^{\dagger}$$^{\ddagger}$ & 32.923 ± 38.721$^{\dagger}$$^{\ddagger}$\\
TotalSegmentator & \textbf{0.932 ± 0.056$^{\dagger}$$^{\ddagger}$} & \textbf{0.709 ± 0.585$^{\dagger}$$^{\ddagger}$} & \textbf{2.796 ± 2.023$^{\dagger}$$^{\ddagger}$}\\
IMPACT-Seg 2.5D & 0.799 ± 0.128$^{\ddagger}$ & 3.102 ± 2.609$^{\ddagger}$ & 11.365 ± 7.498$^{\ddagger}$\\
IMPACT-Seg 2.5D (5-fold ens.) & 0.816 ± 0.130$^{\dagger}$$^{\ddagger}$ & 2.399 ± 2.268$^{\dagger}$ & 8.553 ± 6.255$^{\dagger}$$^{\ddagger}$\\
IMPACT-Seg 3D & 0.831 ± 0.122$^{\dagger}$ & 2.295 ± 2.276$^{\dagger}$ & 7.850 ± 5.375$^{\dagger}$\\
IMPACT-Seg 3D (5-fold ens.) & 0.846 ± 0.119$^{\dagger}$$^{\ddagger}$ & 1.767 ± 1.832$^{\dagger}$$^{\ddagger}$ & 5.903 ± 4.096$^{\dagger}$$^{\ddagger}$\\
\bottomrule
\end{tabular}%
}
\end{table}

\clearpage
\begin{table}[H]
\centering
\small
\caption{Framework-level comparison on the curated SynthRAD MRI subset (52 cases, 39 evaluated labels). IMPACT-Seg is evaluated out of fold: each case is predicted by a fold model that never saw the patient. Mean $\pm$ standard deviation of case-level means over the evaluated reference structures (Section~\ref{sec:evaluation}). Best values per group in bold; $^{\dagger}$/$^{\ddagger}$ overall difference with IMPACT-Seg 2.5D/3D significant (paired Wilcoxon, Holm-corrected $p<0.05$).}
\label{tab:mri_framework_overview}
\fittable{%
\begin{tabular}{lccc}
\toprule
Method & Dice & ASD (mm) & HD95 (mm)\\
\multicolumn{4}{l}{\textit{Abdomen}}\\
MRSegmentator & \textbf{0.783 ± 0.035} & \textbf{2.308 ± 0.699} & \textbf{8.148 ± 3.072}\\
NV-Segment-CTMR & 0.679 ± 0.127 & 5.804 ± 7.568 & 18.269 ± 17.062\\
TotalSegmentator & 0.775 ± 0.083 & 2.690 ± 1.438 & 10.356 ± 5.078\\
IMPACT-Seg 2.5D & 0.764 ± 0.037 & 2.555 ± 0.654 & 8.726 ± 2.480\\
IMPACT-Seg 3D & 0.767 ± 0.048 & 2.506 ± 0.892 & 8.334 ± 2.958\\
\midrule
\multicolumn{4}{l}{\textit{Thorax}}\\
MRSegmentator & 0.757 ± 0.032 & 5.172 ± 2.337 & 22.412 ± 8.812\\
NV-Segment-CTMR & 0.706 ± 0.041 & 4.380 ± 1.692 & 17.065 ± 6.626\\
TotalSegmentator & 0.696 ± 0.059 & 5.280 ± 2.215 & 21.067 ± 8.115\\
IMPACT-Seg 2.5D & 0.822 ± 0.032 & \textbf{1.820 ± 0.323} & \textbf{5.727 ± 1.214}\\
IMPACT-Seg 3D & \textbf{0.823 ± 0.031} & 1.847 ± 0.415 & 5.969 ± 1.539\\
\midrule
\multicolumn{4}{l}{\textit{Pelvis}}\\
MRSegmentator & 0.835 ± 0.031 & 2.078 ± 0.541 & 8.317 ± 4.205\\
NV-Segment-CTMR & 0.642 ± 0.183 & 17.123 ± 17.758 & 58.893 ± 43.678\\
TotalSegmentator & \textbf{0.901 ± 0.036} & \textbf{1.272 ± 0.531} & \textbf{5.318 ± 4.148}\\
IMPACT-Seg 2.5D & 0.852 ± 0.034 & 1.656 ± 0.454 & 6.942 ± 3.779\\
IMPACT-Seg 3D & 0.859 ± 0.034 & 1.571 ± 0.469 & 6.067 ± 3.685\\
\midrule
\multicolumn{4}{l}{\textit{Overall}}\\
MRSegmentator & 0.803 ± 0.045$^{\ddagger}$ & 2.748 ± 1.641$^{\dagger}$$^{\ddagger}$ & 10.972 ± 7.534$^{\dagger}$$^{\ddagger}$\\
NV-Segment-CTMR & 0.667 ± 0.147$^{\dagger}$$^{\ddagger}$ & 10.972 ± 14.242$^{\dagger}$$^{\ddagger}$ & 37.568 ± 37.767$^{\dagger}$$^{\ddagger}$\\
TotalSegmentator & 0.820 ± 0.101 & 2.506 ± 1.978 & 9.993 ± 7.913\\
IMPACT-Seg 2.5D & 0.817 ± 0.052 & 1.981 ± 0.645 & 7.291 ± 3.183\\
IMPACT-Seg 3D & \textbf{0.822 ± 0.056} & \textbf{1.929 ± 0.748} & \textbf{6.789 ± 3.277}\\
\bottomrule
\end{tabular}%
}
\end{table}

\begin{table}[H]
\centering
\small
\caption{Framework-level comparison on the AMOS22 MRI test cases (20 cases, 36 evaluated labels). IMPACT-Seg is evaluated out of fold: each case is predicted by a fold model that never saw the patient. Mean $\pm$ standard deviation of case-level means over the evaluated reference structures (Section~\ref{sec:evaluation}). Best values per group in bold; $^{\dagger}$/$^{\ddagger}$ overall difference with IMPACT-Seg 2.5D/3D significant (paired Wilcoxon, Holm-corrected $p<0.05$).}
\label{tab:mri_amos_framework_overview}
\fittable{%
\begin{tabular}{lccc}
\toprule
Method & Dice & ASD (mm) & HD95 (mm)\\
\multicolumn{4}{l}{\textit{AMOS22 MRI}}\\
MRSegmentator & 0.787 ± 0.050 & 2.156 ± 0.709 & 7.689 ± 2.264\\
NV-Segment-CTMR & \textbf{0.818 ± 0.048} & 1.969 ± 1.181 & 7.449 ± 6.342\\
TotalSegmentator & 0.791 ± 0.040 & 1.942 ± 0.370 & 6.533 ± 1.232\\
IMPACT-Seg 2.5D & 0.808 ± 0.045 & 1.773 ± 0.475 & 6.330 ± 1.561\\
IMPACT-Seg 3D & 0.816 ± 0.043 & \textbf{1.601 ± 0.473} & \textbf{5.509 ± 2.076}\\
\midrule
\multicolumn{4}{l}{\textit{Overall}}\\
MRSegmentator & 0.787 ± 0.050$^{\dagger}$$^{\ddagger}$ & 2.156 ± 0.709$^{\dagger}$$^{\ddagger}$ & 7.689 ± 2.264$^{\dagger}$$^{\ddagger}$\\
NV-Segment-CTMR & \textbf{0.818 ± 0.048} & 1.969 ± 1.181 & 7.449 ± 6.342\\
TotalSegmentator & 0.791 ± 0.040$^{\dagger}$$^{\ddagger}$ & 1.942 ± 0.370$^{\ddagger}$ & 6.533 ± 1.232$^{\ddagger}$\\
IMPACT-Seg 2.5D & 0.808 ± 0.045 & 1.773 ± 0.475 & 6.330 ± 1.561\\
IMPACT-Seg 3D & 0.816 ± 0.043 & \textbf{1.601 ± 0.473} & \textbf{5.509 ± 2.076}\\
\bottomrule
\end{tabular}%
}
\end{table}

\begin{table}[H]
\centering
\small
\caption{Framework-level comparison on the curated TotalSegmentator MRI test cases (\res{tsmr.cases} cases, 39 evaluated labels). IMPACT-Seg is evaluated out of fold: each case is predicted by a fold model that never saw the patient. Mean $\pm$ standard deviation of case-level means over the evaluated reference structures (Section~\ref{sec:evaluation}). Best values per group in bold; $^{\dagger}$/$^{\ddagger}$ overall difference with IMPACT-Seg 2.5D/3D significant (paired Wilcoxon, Holm-corrected $p<0.05$).}
\label{tab:mri_tsmr_framework_overview}
\fittable{%
\begin{tabular}{lccc}
\toprule
Method & Dice & ASD (mm) & HD95 (mm)\\
\multicolumn{4}{l}{\textit{TotalSeg MRI}}\\
MRSegmentator & 0.730 ± 0.181 & 4.320 ± 5.876 & 14.992 ± 16.808\\
NV-Segment-CTMR & 0.696 ± 0.216 & 8.480 ± 18.382 & 22.815 ± 37.937\\
IMPACT-Seg 2.5D & 0.732 ± 0.115 & 3.671 ± 3.204 & 13.805 ± 10.925\\
IMPACT-Seg 3D & \textbf{0.768 ± 0.085} & \textbf{2.851 ± 2.015} & \textbf{10.650 ± 7.276}\\
\midrule
\multicolumn{4}{l}{\textit{Overall}}\\
MRSegmentator & 0.730 ± 0.181 & 4.320 ± 5.876 & 14.992 ± 16.808$^{\ddagger}$\\
NV-Segment-CTMR & 0.696 ± 0.216 & 8.480 ± 18.382$^{\ddagger}$ & 22.815 ± 37.937$^{\ddagger}$\\
IMPACT-Seg 2.5D & 0.732 ± 0.115$^{\ddagger}$ & 3.671 ± 3.204$^{\ddagger}$ & 13.805 ± 10.925$^{\ddagger}$\\
IMPACT-Seg 3D & \textbf{0.768 ± 0.085$^{\dagger}$} & \textbf{2.851 ± 2.015$^{\dagger}$} & \textbf{10.650 ± 7.276$^{\dagger}$}\\
\bottomrule
\end{tabular}%
}
\end{table}

\begin{table}[H]
\centering
\scriptsize
\caption{Framework-level comparison on the external TotalSegmentator v3 axial MRI cases (absent from v2) (253 cases, 39 evaluated labels). IMPACT-Seg is evaluated out of fold: each case is predicted by a fold model that never saw the patient. Mean $\pm$ standard deviation of case-level means over the evaluated reference structures (Section~\ref{sec:evaluation}). Best values per group in bold; $^{\dagger}$/$^{\ddagger}$ overall difference with IMPACT-Seg 2.5D/3D significant (paired Wilcoxon, Holm-corrected $p<0.05$).}
\label{tab:ts_v3_mri_framework_overview}
\fittable{%
\begin{tabular}{lccc}
\toprule
Method & Dice & ASD & HD95\\
\midrule
MRSegmentator & 0.812 ± 0.098$^{\dagger}$$^{\ddagger}$ & 2.509 ± 2.167$^{\dagger}$$^{\ddagger}$ & 9.872 ± 8.533$^{\dagger}$$^{\ddagger}$\\
NV-Segment-CTMR & 0.718 ± 0.155$^{\dagger}$$^{\ddagger}$ & 12.043 ± 14.320$^{\dagger}$$^{\ddagger}$ & 45.411 ± 33.073$^{\dagger}$$^{\ddagger}$\\
TotalSegmentator & \textbf{0.843 ± 0.087$^{\dagger}$$^{\ddagger}$} & \textbf{2.242 ± 2.384$^{\dagger}$$^{\ddagger}$} & \textbf{8.777 ± 8.102$^{\dagger}$$^{\ddagger}$}\\
IMPACT-Seg 2.5D & 0.758 ± 0.109$^{\ddagger}$ & 4.133 ± 3.302$^{\ddagger}$ & 15.038 ± 9.904$^{\ddagger}$\\
IMPACT-Seg 2.5D (5-fold ens.) & 0.780 ± 0.109$^{\dagger}$ & 3.202 ± 2.057$^{\dagger}$ & 11.709 ± 6.755$^{\dagger}$\\
IMPACT-Seg 3D & 0.778 ± 0.102$^{\dagger}$ & 3.302 ± 2.277$^{\dagger}$ & 11.748 ± 6.994$^{\dagger}$\\
IMPACT-Seg 3D (5-fold ens.) & 0.795 ± 0.108$^{\dagger}$$^{\ddagger}$ & 2.852 ± 2.215$^{\dagger}$$^{\ddagger}$ & 9.963 ± 6.101$^{\dagger}$$^{\ddagger}$\\
\bottomrule
\end{tabular}%
}
\end{table}

\begin{table}[H]
\centering
\caption{Mean Dice with all evaluated labels and without the vertebra and rib labels (level identification excluded), same evaluated cases and labels as the main tables. CADS is not evaluated on VISCERAL (training overlap).}
\label{tab:novertrib}
\small
\fittable{%
\begin{tabular}{lcccccccccc}
\toprule
 & \multicolumn{2}{c}{IMPACT-Seg 2.5D} & \multicolumn{2}{c}{IMPACT-Seg 3D} & \multicolumn{2}{c}{TotalSeg.} & \multicolumn{2}{c}{CADS} & \multicolumn{2}{c}{NV-Seg.}\\
\cmidrule(lr){2-3}\cmidrule(lr){4-5}\cmidrule(lr){6-7}\cmidrule(lr){8-9}\cmidrule(lr){10-11}
Benchmark & all & w/o V\&R & all & w/o V\&R & all & w/o V\&R & all & w/o V\&R & all & w/o V\&R\\
\midrule
SynthRAD CBCT & \res{synthradcbct.ims25.dice} & \res{nvrsynthradcbct.ims25.dice} & \res{synthradcbct.ims3d.dice} & \res{nvrsynthradcbct.ims3d.dice} & \res{synthradcbct.ts.dice} & \res{nvrsynthradcbct.ts.dice} & \res{synthradcbct.cads.dice} & \res{nvrsynthradcbct.cads.dice} & \res{synthradcbct.nv.dice} & \res{nvrsynthradcbct.nv.dice}\\
Simulated CBCT & \res{tsrtk.ims25.dice} & \res{nvrtsrtk.ims25.dice} & \res{tsrtk.ims3d.dice} & \res{nvrtsrtk.ims3d.dice} & \res{tsrtk.ts.dice} & \res{nvrtsrtk.ts.dice} & \res{tsrtk.cads.dice} & \res{nvrtsrtk.cads.dice} & \res{tsrtk.nv.dice} & \res{nvrtsrtk.nv.dice}\\
VISCERAL CT & \res{visceral.ims25.dice} & \res{nvrvisceral.ims25.dice} & \res{visceral.ims3d.dice} & \res{nvrvisceral.ims3d.dice} & \res{visceral.ts.dice} & \res{nvrvisceral.ts.dice} & -- & -- & \res{visceral.nv.dice} & \res{nvrvisceral.nv.dice}\\
TotalSeg.\ v3 CT & \res{tsvthreect.ims25.dice} & \res{nvrtsvthreect.ims25.dice} & \res{tsvthreect.ims3d.dice} & \res{nvrtsvthreect.ims3d.dice} & \res{tsvthreect.ts.dice} & \res{nvrtsvthreect.ts.dice} & \res{tsvthreect.cads.dice} & \res{nvrtsvthreect.cads.dice} & \res{tsvthreect.nv.dice} & \res{nvrtsvthreect.nv.dice}\\
\bottomrule
\end{tabular}
}
\end{table}

\clearpage
\FloatBarrier
\Needspace{12\baselineskip}
\section{2.5D versus 3D Networks} \label{app:dim}

% Source: make_results.sh NETWORKS="2_5D 3D" (paper_v2/results); internal: internal_oof_{2p5d,3d}; p: paired Wilcoxon on the per-case Dice, Holm-corrected within each benchmark.
\begin{table}[H]
\centering
\caption{2.5D versus 3D IMPACT-Seg (five folds each, out-of-fold predictions), on the same cases and patient--label pairs as Tables~\ref{tab:synthrad_cbct_framework_overview}--\ref{tab:ts_v3_mri_framework_overview}: per-case mean Dice / HD95 (mm). $p$: paired Wilcoxon test on the per-case Dice, Holm-corrected. Bottom rows: five-fold probability ensembles on the external cases, which no fold has seen.}
\label{tab:dim_ablation}
\small
\fittable{%
\begin{tabular}{lrccc}
\toprule
Benchmark & Cases & 2.5D & 3D & $p$ \\
\midrule
SynthRAD CBCT & \res{synthradcbct.cases} & \res{synthradcbct.ims25.dice} / \res{synthradcbct.ims25.hd95} & \textbf{\res{synthradcbct.ims3d.dice}} / \res{synthradcbct.ims3d.hd95} & \res{synthradcbct.ims3d.p} \\
Simulated CBCT (TS\_RTK) & \res{tsrtk.cases} & \res{tsrtk.ims25.dice} / \res{tsrtk.ims25.hd95} & \textbf{\res{tsrtk.ims3d.dice}} / \res{tsrtk.ims3d.hd95} & \res{tsrtk.ims3d.p} \\
VISCERAL CT & \res{visceral.cases} & \res{visceral.ims25.dice} / \res{visceral.ims25.hd95} & \textbf{\res{visceral.ims3d.dice}} / \res{visceral.ims3d.hd95} & \res{visceral.ims3d.p} \\
TotalSegmentator v3 CT & \res{tsvthreect.cases} & \res{tsvthreect.ims25.dice} / \res{tsvthreect.ims25.hd95} & \textbf{\res{tsvthreect.ims3d.dice}} / \res{tsvthreect.ims3d.hd95} & \res{tsvthreect.ims3d.p} \\
SynthRAD MRI & \res{synthradmri.cases} & \res{synthradmri.ims25.dice} / \res{synthradmri.ims25.hd95} & \textbf{\res{synthradmri.ims3d.dice}} / \res{synthradmri.ims3d.hd95} & \res{synthradmri.ims3d.p} \\
AMOS-MR & \res{amos.cases} & \res{amos.ims25.dice} / \res{amos.ims25.hd95} & \textbf{\res{amos.ims3d.dice}} / \res{amos.ims3d.hd95} & \res{amos.ims3d.p} \\
TotalSegmentator-MR & \res{tsmr.cases} & \res{tsmr.ims25.dice} / \res{tsmr.ims25.hd95} & \textbf{\res{tsmr.ims3d.dice}} / \res{tsmr.ims3d.hd95} & \res{tsmr.ims3d.p} \\
TotalSegmentator v3 MRI & \res{tsvthreemri.cases} & \res{tsvthreemri.ims25.dice} / \res{tsvthreemri.ims25.hd95} & \textbf{\res{tsvthreemri.ims3d.dice}} / \res{tsvthreemri.ims3d.hd95} & \res{tsvthreemri.ims3d.p} \\
\midrule
v3 CT, 5-fold ensemble & \res{tsvthreect.cases} & \res{tsvthreect.ims25ens.dice} / \res{tsvthreect.ims25ens.hd95} & \textbf{\res{tsvthreect.ims3dens.dice}} / \res{tsvthreect.ims3dens.hd95} & -- \\
v3 MRI, 5-fold ensemble & \res{tsvthreemri.cases} & \res{tsvthreemri.ims25ens.dice} / \res{tsvthreemri.ims25ens.hd95} & \textbf{\res{tsvthreemri.ims3dens.dice}} / \res{tsvthreemri.ims3dens.hd95} & -- \\
\midrule
Internal test (126 structures) & \res{internal.ALL.cases} & \res{internal.ALL.dice} & \textbf{\res{internal3d.ALL.dice}} & -- \\
\bottomrule
\end{tabular}
}
\end{table}

\FloatBarrier
\Needspace{12\baselineskip}
\section{Feature-Level Consistency} \label{app:features}

Table~\ref{tab:feature_consistency} gives the feature-level measures of every network of the training-data ablation at three depths.

\begin{table}[H]
\centering
\footnotesize
\caption{Feature-level consistency of the 2.5D fold-1 networks (training-data ablation and main fold-1 model) on paired volumes of unseen patients, at the fifth encoder stage (Enc.~5), the second decoder stage (Dec.~2) and the last decoder layer used for registration (Dec.~5): correspondence retrieval error within a 96~mm disc (chance median 68~mm) and linear CKA of the matched features. Mean $\pm$ standard deviation across patients; retrieval is first summarized by the median error of each patient. $^{*}$: different from the previous training set (paired Wilcoxon test, Holm-corrected over the table, $p<0.05$; tests use the unrounded per-patient values).}
\label{tab:feature_consistency}
\setlength{\tabcolsep}{2.5pt}
\fittable{%
\begin{tabular}{llcccccc}
\toprule
 & & \multicolumn{2}{c}{Enc.~5} & \multicolumn{2}{c}{Dec.~2} & \multicolumn{2}{c}{Dec.~5}\\
\cmidrule(lr){3-4}\cmidrule(lr){5-6}\cmidrule(lr){7-8}
Pairs & Training set & Retr.\ (mm) & CKA & Retr.\ (mm) & CKA & Retr.\ (mm) & CKA\\
\midrule
CT / simulated CBCT (TS\_RTK) ($n=48$) & CT only & 2.0$\pm$0.6 & 0.90$\pm$0.05 & 2.1$\pm$0.7 & 0.96$\pm$0.03 & 10.2$\pm$4.3 & 0.88$\pm$0.10\\
 & CT + real CBCT & 1.9$\pm$0.6 & 0.91$\pm$0.04$^{*}$ & 2.0$\pm$0.8 & 0.96$\pm$0.03 & 9.7$\pm$4.1 & 0.90$\pm$0.08$^{*}$\\
 & CT + all CBCT & 1.3$\pm$0.6$^{*}$ & 0.95$\pm$0.02$^{*}$ & 1.4$\pm$0.5$^{*}$ & 0.98$\pm$0.01$^{*}$ & 4.9$\pm$2.2$^{*}$ & 0.97$\pm$0.04$^{*}$\\
 & Full & 1.3$\pm$0.6 & 0.96$\pm$0.02$^{*}$ & 1.3$\pm$0.6 & 0.98$\pm$0.01$^{*}$ & 4.4$\pm$2.0$^{*}$ & 0.97$\pm$0.03$^{*}$\\
 & Main (fold 1) & 1.3$\pm$0.6 & 0.96$\pm$0.02 & 1.5$\pm$0.5 & 0.98$\pm$0.01 & 4.2$\pm$2.4 & 0.97$\pm$0.03\\
\midrule
CT / real CBCT (SynthRAD) ($n=57$) & CT only & 2.3$\pm$0.7 & 0.90$\pm$0.05 & 2.5$\pm$1.1 & 0.95$\pm$0.04 & 11.7$\pm$4.6 & 0.87$\pm$0.09\\
 & CT + real CBCT & 2.0$\pm$0.6$^{*}$ & 0.91$\pm$0.05$^{*}$ & 2.3$\pm$1.0$^{*}$ & 0.96$\pm$0.03 & 10.4$\pm$4.1$^{*}$ & 0.91$\pm$0.05$^{*}$\\
 & CT + all CBCT & 1.7$\pm$0.5$^{*}$ & 0.93$\pm$0.04$^{*}$ & 1.8$\pm$0.6$^{*}$ & 0.97$\pm$0.03$^{*}$ & 8.1$\pm$3.6$^{*}$ & 0.93$\pm$0.05$^{*}$\\
 & Full & 1.7$\pm$0.4 & 0.94$\pm$0.04$^{*}$ & 1.8$\pm$0.6 & 0.97$\pm$0.02 & 7.4$\pm$3.2$^{*}$ & 0.95$\pm$0.03$^{*}$\\
 & Main (fold 1) & 1.8$\pm$0.4 & 0.94$\pm$0.04 & 2.0$\pm$0.7 & 0.97$\pm$0.03 & 7.0$\pm$3.7 & 0.95$\pm$0.04\\
\midrule
CT / MRI (SynthRAD) ($n=11$) & CT only & 44.0$\pm$23.9 & 0.63$\pm$0.07 & 47.3$\pm$19.9 & 0.53$\pm$0.12 & 55.9$\pm$12.9 & 0.34$\pm$0.19\\
 & CT + real CBCT & 37.4$\pm$21.5$^{*}$ & 0.63$\pm$0.07 & 43.0$\pm$18.1 & 0.55$\pm$0.14 & 52.3$\pm$12.6 & 0.33$\pm$0.21\\
 & CT + all CBCT & 36.8$\pm$22.8 & 0.66$\pm$0.06 & 41.1$\pm$22.5 & 0.61$\pm$0.15 & 50.5$\pm$16.0 & 0.35$\pm$0.23\\
 & Full & 3.2$\pm$0.5$^{*}$ & 0.89$\pm$0.04$^{*}$ & 3.7$\pm$0.9$^{*}$ & 0.95$\pm$0.02$^{*}$ & 17.2$\pm$3.3$^{*}$ & 0.91$\pm$0.04$^{*}$\\
 & Main (fold 1) & 3.1$\pm$0.4 & 0.90$\pm$0.04 & 4.4$\pm$1.3 & 0.96$\pm$0.02 & 16.7$\pm$3.6 & 0.91$\pm$0.06\\
\bottomrule
\end{tabular}}
\end{table}

\FloatBarrier
\Needspace{12\baselineskip}
\section{Uncertainty Analysis} \label{app:uncertainty}

\begin{figure}[!htbp]
\centering
\includegraphics[width=\textwidth]{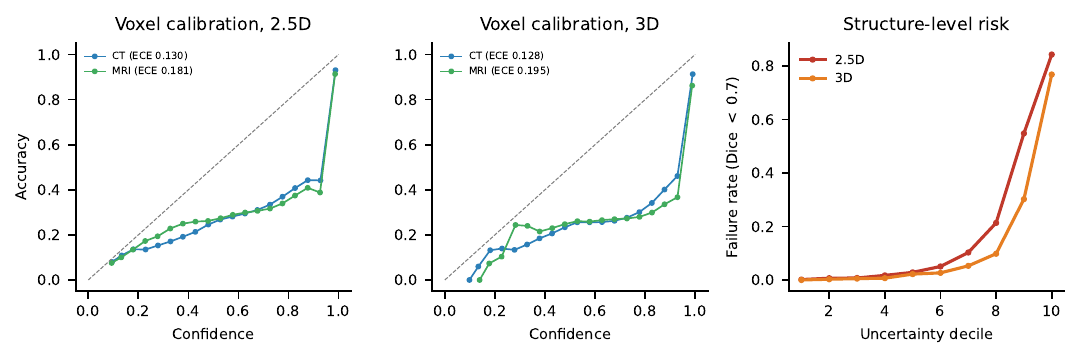}
\caption{Calibration on the external TotalSegmentator v3 CT and MRI cohorts. Left and middle: voxel-level reliability diagrams of the softmax confidence (accuracy of the predicted label against the reference, over the predicted and reference foreground) with the expected calibration error (ECE). Right: failure rate (Dice $<$ 0.7) of the structures per decile of fold disagreement, CT and MRI pooled.}
\label{fig:uncertainty_calibration}
\end{figure}

\begin{figure}[!htbp]
\centering
\includegraphics[width=\textwidth]{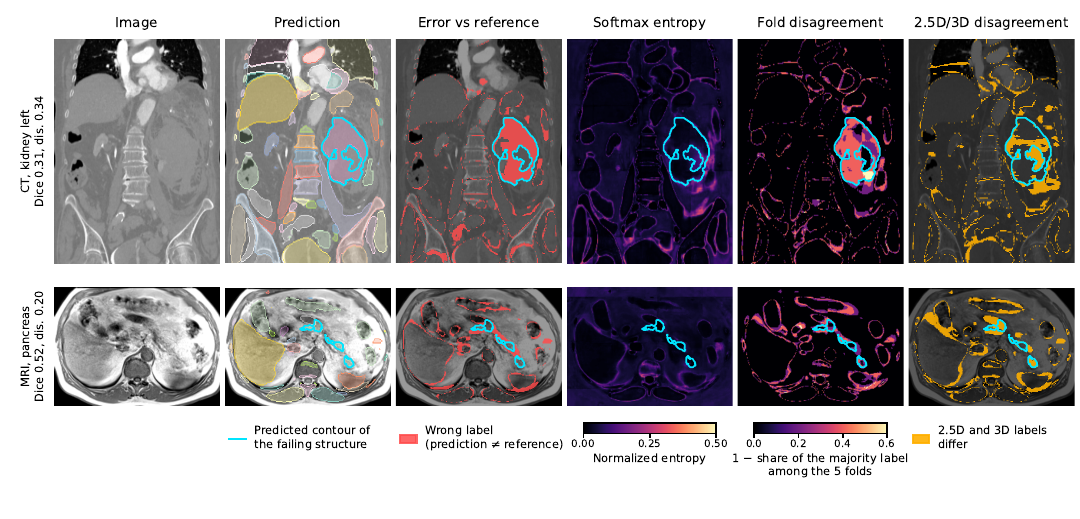}
\caption{Uncertainty maps of the 3D network on two failures flagged by the fold disagreement in the TotalSegmentator v3 cohorts, on the slice with the largest error of the structure. Top (CT, coronal): large mass at the left kidney, which the network partly labels as kidney (predicted 416 ml, reference 96 ml). Bottom (MRI, axial): under-segmented pancreas, of which the network recovers 26 of the 69 ml of the reference and leaves the rest unlabelled. From left to right: image, prediction, error against the reference on all the structures retained for the case (shown for evaluation only), normalized softmax entropy, fold disagreement (one minus the share of the majority label among the five fold models) and voxels where the 2.5D and 3D networks predict different labels. Cyan: predicted contour of the failing structure. In both cases, the entropy of the single model stays low on the failing structure whereas the fold models disagree.}
\label{fig:uncertainty_examples}
\end{figure}

Table~\ref{tab:uncertainty_summary} summarizes uncertainty per benchmark and network, Table~\ref{tab:uncertainty_volume} stratifies detection by structure volume, and Table~\ref{tab:uncertainty_triage} reports triage by fold disagreement. Figure~\ref{fig:uncertainty_calibration} shows voxel-level calibration and failure rates by uncertainty decile; Figure~\ref{fig:uncertainty_examples} illustrates detected failures. \ifextendedsupplement Table~\ref{tab:uncertainty_detection_dice05} additionally reports Dice $<$ 0.5 detection; Figures~\ref{fig:uncertainty_vs_dice} and~\ref{fig:uncertainty_roc_pr} provide structure-level association and ROC/precision-recall curves.\fi

\begin{table}[H]
\centering
\footnotesize
\caption{Uncertainty of IMPACT-Seg per benchmark and network (out-of-fold, evaluated patient--label pairs of the main comparison): structures, failures (Dice $<$ 0.7), structures not predicted at all, mean normalized entropy inside the predicted structures, and failure detection (AUROC, Dice $<$ 0.7) of the entropy, the fold disagreement and the 2.5D/3D disagreement inside the predicted structure.}
\label{tab:uncertainty_summary}
\fittable{%
\begin{tabular}{llrrrrrrc}
\toprule
Benchmark & Network & Cases & Structures & Mean Dice & Failures & Missed & Entropy & AUROC entropy / fold / 2.5D-3D\\
\midrule
TotalSeg.\ v3 CT & 2.5D & 247 & 17740 & 0.810 & 16.5\% & 45 & 0.045 & 0.83 / 0.93 / 0.92\\
TotalSeg.\ v3 CT & 3D & 247 & 17740 & 0.843 & 10.8\% & 53 & 0.028 & 0.73 / 0.94 / 0.88\\
TotalSeg.\ v3 MRI & 2.5D & 253 & 7952 & 0.776 & 23.0\% & 30 & 0.058 & 0.83 / 0.93 / 0.90\\
TotalSeg.\ v3 MRI & 3D & 253 & 7952 & 0.798 & 18.6\% & 35 & 0.031 & 0.60 / 0.92 / 0.84\\
\bottomrule
\end{tabular}
}
\end{table}

\ifextendedsupplement
\begin{table}[H]
\centering
\footnotesize
\caption{Detection of structures with Dice $<$ 0.5 from uncertainty on the external TotalSegmentator v3 cohorts: AUROC and AUPRC with patient-level bootstrap 95\% confidence intervals (10\,000 resamples). Random baseline: AUROC 0.5 and AUPRC equal to the failure prevalence. Structures not predicted at all are excluded (they carry no uncertainty). The volume baseline ranks structures by their predicted volume only (smaller is riskier).}
\label{tab:uncertainty_detection_dice05}
\setlength{\tabcolsep}{4pt}
\begin{tabular}{lcccc}
\toprule
 & \multicolumn{2}{c}{CT} & \multicolumn{2}{c}{MRI}\\
\cmidrule(lr){2-3}\cmidrule(lr){4-5}
Signal & 2.5D & 3D & 2.5D & 3D\\
\midrule
Structures ($n$) & 17695 & 17687 & 7922 & 7917\\
Failure prevalence & 8.3\% & 5.0\% & 8.3\% & 5.7\%\\
\midrule
\multicolumn{5}{l}{\textit{AUROC}}\\
Entropy (inside) & 0.86 [0.83, 0.89] & 0.80 [0.75, 0.85] & 0.86 [0.83, 0.89] & 0.66 [0.62, 0.70]\\
Entropy (boundary band) & 0.68 [0.63, 0.73] & 0.64 [0.56, 0.71] & 0.69 [0.64, 0.73] & 0.46 [0.41, 0.50]\\
Fold disagreement (inside) & \textbf{0.93 [0.92, 0.94]} & \textbf{0.96 [0.94, 0.97]} & \textbf{0.94 [0.93, 0.95]} & \textbf{0.94 [0.92, 0.95]}\\
Fold disagreement (band) & 0.86 [0.84, 0.88] & 0.91 [0.89, 0.93] & 0.84 [0.83, 0.86] & 0.84 [0.82, 0.87]\\
2.5D/3D disagreement (inside) & 0.93 [0.92, 0.94] & 0.90 [0.88, 0.92] & 0.89 [0.87, 0.92] & 0.85 [0.82, 0.87]\\
2.5D/3D disagreement (band) & 0.90 [0.89, 0.92] & 0.88 [0.85, 0.90] & 0.86 [0.84, 0.88] & 0.85 [0.83, 0.88]\\
1 $-$ confidence & 0.87 [0.84, 0.90] & 0.80 [0.75, 0.84] & 0.87 [0.84, 0.89] & 0.66 [0.62, 0.69]\\
Volume only (baseline) & 0.81 [0.78, 0.83] & 0.85 [0.81, 0.87] & 0.88 [0.86, 0.89] & 0.89 [0.87, 0.90]\\
\midrule
\multicolumn{5}{l}{\textit{AUPRC}}\\
Entropy (inside) & 0.57 [0.48, 0.65] & 0.48 [0.36, 0.58] & 0.59 [0.52, 0.65] & 0.29 [0.24, 0.34]\\
Entropy (boundary band) & 0.33 [0.23, 0.42] & 0.30 [0.19, 0.40] & 0.31 [0.22, 0.40] & 0.13 [0.09, 0.17]\\
Fold disagreement (inside) & 0.63 [0.55, 0.70] & \textbf{0.67 [0.58, 0.74]} & \textbf{0.69 [0.64, 0.73]} & \textbf{0.56 [0.51, 0.61]}\\
Fold disagreement (band) & 0.45 [0.37, 0.52] & 0.51 [0.41, 0.59] & 0.40 [0.35, 0.45] & 0.32 [0.28, 0.37]\\
2.5D/3D disagreement (inside) & \textbf{0.69 [0.62, 0.74]} & 0.45 [0.34, 0.53] & 0.60 [0.53, 0.65] & 0.39 [0.34, 0.45]\\
2.5D/3D disagreement (band) & 0.55 [0.49, 0.61] & 0.33 [0.27, 0.39] & 0.45 [0.37, 0.52] & 0.34 [0.29, 0.38]\\
1 $-$ confidence & 0.59 [0.50, 0.67] & 0.47 [0.34, 0.57] & 0.61 [0.54, 0.67] & 0.27 [0.22, 0.33]\\
Volume only (baseline) & 0.39 [0.31, 0.47] & 0.39 [0.30, 0.48] & 0.52 [0.45, 0.57] & 0.49 [0.43, 0.54]\\
\bottomrule
\end{tabular}
\end{table}

\fi
\begin{table}[H]
\centering
\footnotesize
\caption{Failure detection (AUROC, Dice $<$ 0.7) per reference volume of the structure, TotalSegmentator v3 CT and MRI pooled. Within a volume range, the size of the structure alone carries little information, whereas the disagreement signals remain discriminative.}
\label{tab:uncertainty_volume}
\setlength{\tabcolsep}{3pt}
\begin{tabular}{llrccccc}
\toprule
Volume & Net. & Prev. & Entropy & Fold dis. & 2.5D/3D dis. & 1 $-$ conf. & Volume\\
\midrule
$<$10 ml & 2.5D & 46\% & 0.87 & 0.89 & 0.88 & 0.86 & 0.66\\
 & 3D & 39\% & 0.74 & 0.89 & 0.81 & 0.69 & 0.70\\
10--100 ml & 2.5D & 19\% & 0.80 & 0.90 & 0.90 & 0.81 & 0.61\\
 & 3D & 11\% & 0.69 & 0.92 & 0.85 & 0.64 & 0.59\\
$\geq$100 ml & 2.5D & 5\% & 0.93 & 0.95 & 0.91 & 0.92 & 0.57\\
 & 3D & 3\% & 0.85 & 0.95 & 0.88 & 0.80 & 0.63\\
\bottomrule
\end{tabular}
\end{table}

\begin{table}[H]
\centering
\footnotesize
\caption{Triage of the structures by fold disagreement (Dice $<$ 0.7, TotalSegmentator v3 CT and MRI): share of the failing structures captured when the given fraction of the most uncertain structures is reviewed, failures left unflagged (including the structures not predicted at all, which are never flagged). Random review captures, on average, the reviewed fraction.}
\label{tab:uncertainty_triage}
\begin{tabular}{llrrrr}
\toprule
Network & Signal & Reviewed & Failures captured & Unflagged failures & Workload saved\\
\midrule
2.5D & Fold disagreement (inside) & 5\% & 24\% (1143/4751) & 3608 & 95\%\\
2.5D & Fold disagreement (inside) & 10\% & 46\% (2165/4751) & 2586 & 90\%\\
2.5D & Fold disagreement (inside) & 20\% & 75\% (3572/4751) & 1179 & 80\%\\
2.5D & Fold disagreement (inside) & 30\% & 87\% (4116/4751) & 635 & 70\%\\
\midrule
3D & Fold disagreement (inside) & 5\% & 33\% (1107/3394) & 2287 & 95\%\\
3D & Fold disagreement (inside) & 10\% & 58\% (1974/3394) & 1420 & 90\%\\
3D & Fold disagreement (inside) & 20\% & 81\% (2744/3394) & 650 & 80\%\\
3D & Fold disagreement (inside) & 30\% & 88\% (2996/3394) & 398 & 70\%\\
\bottomrule
\end{tabular}
\end{table}

\ifextendedsupplement
\begin{figure}[!htbp]
\centering
\includegraphics[width=0.76\linewidth]{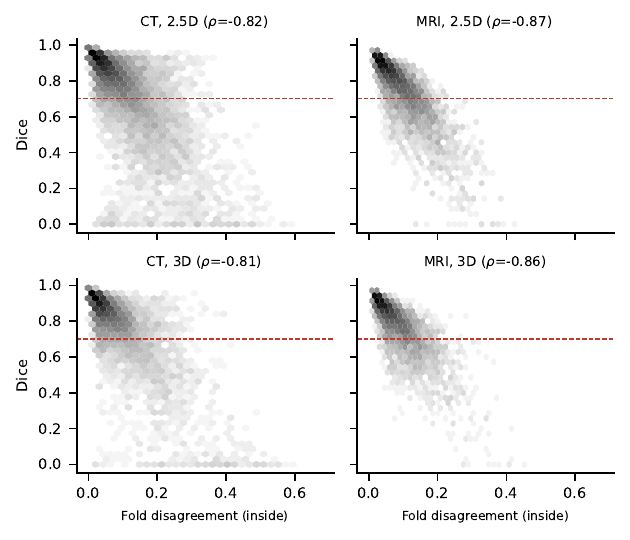}
\caption{Fold disagreement inside the predicted structure versus Dice of every evaluated structure of the TotalSegmentator v3 CT and MRI cohorts (logarithmic density), with the Spearman correlation. Dashed line: failure threshold (Dice 0.7).}
\label{fig:uncertainty_vs_dice}
\end{figure}

\begin{figure}[!htbp]
\centering
\includegraphics[width=0.76\linewidth]{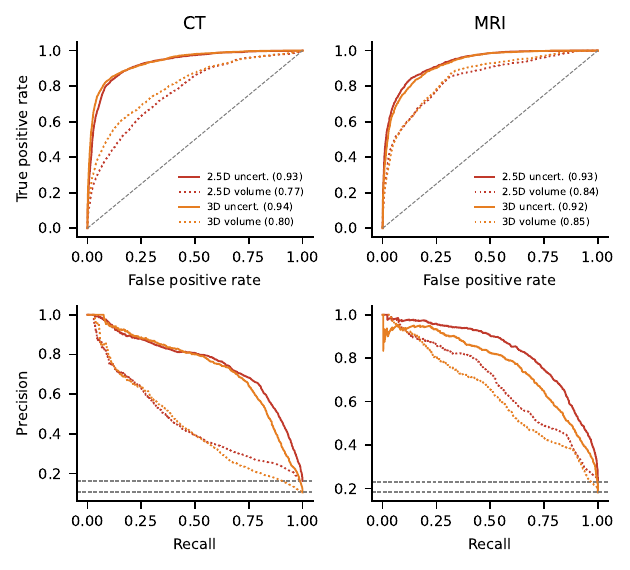}
\caption{ROC (top) and precision-recall (bottom) curves for the detection of structures with Dice $<$ 0.7 by the fold disagreement (solid) and by the predicted volume (dotted), with the AUROC in parentheses. Dashed horizontal lines: failure prevalence of each network.}
\label{fig:uncertainty_roc_pr}
\end{figure}
\fi

\FloatBarrier
\Needspace{12\baselineskip}
\section{Qualitative Results} \label{app:qualitative}

\begin{figure}[H]
\centering
\includegraphics[width=0.98\textwidth]{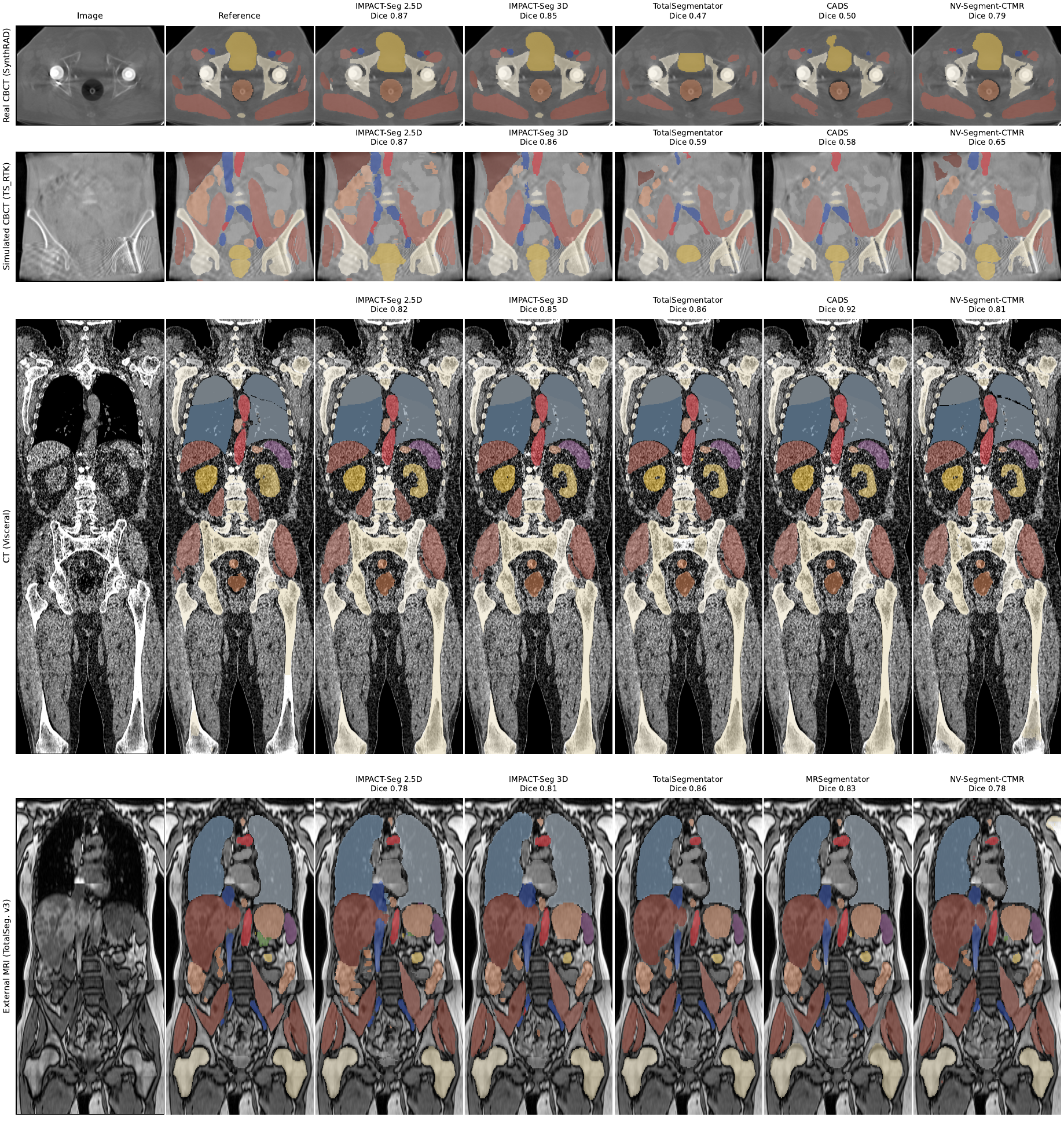}
\caption{Representative cases: for each benchmark, the case whose IMPACT-Seg Dice (mean of the 2.5D and 3D networks) is closest to the benchmark median, on the slice containing the most reference structures. Only the structures shared by the compared methods and evaluated in the benchmark are shown; titles give the mean Dice of the case over these structures.}
\label{fig:qualitative_median}
\end{figure}

\begin{figure}[!htbp]
\centering
\includegraphics[width=\textwidth]{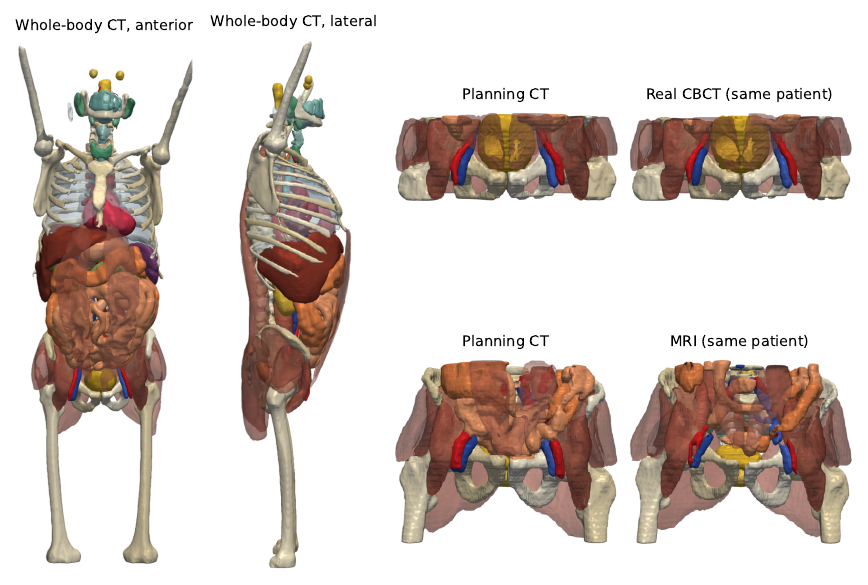}
\caption{Dense anatomical output of IMPACT-Seg (3D network). Left: surface rendering of the structures predicted on a held-out whole-body CT. Right: predictions on the planning CT and on the registered real CBCT (top) or MRI (bottom) of the same patient; lungs and large muscles are rendered translucent.}
\label{fig:dense}
\end{figure}

\begin{figure}[!htbp]
\centering
\includegraphics[width=\textwidth]{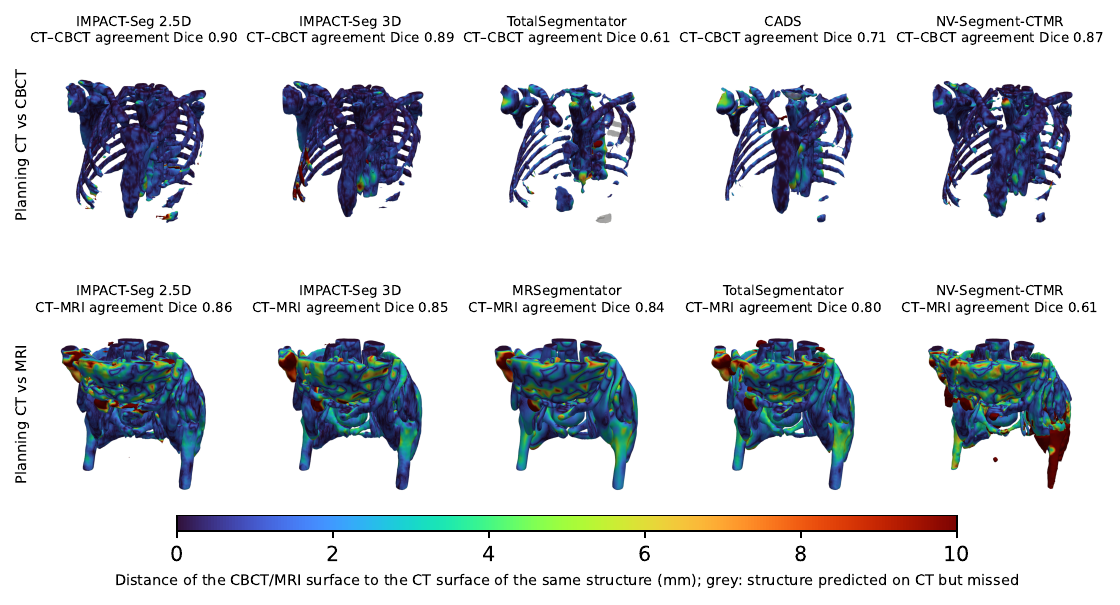}
\caption{Cross-modality agreement in 3D for one SynthRAD patient per cohort (a thoracic CBCT on which every method predicts most structures, and a pelvic MRI). Each method segments the planning CT and the registered CBCT (top) or MRI (bottom); the CBCT/MRI surface of every structure is coloured by its distance to the surface of the same structure predicted on the CT. Structures predicted on the CT but missed on the CBCT/MRI are shown in grey. Titles give the CT--CBCT/MRI agreement Dice of the case.}
\label{fig:agreement3d}
\end{figure}

Figures~\ref{fig:qualitative_median}--\ref{fig:agreement3d} illustrate representative predictions, dense anatomical coverage, and cross-modality agreement. Case-selection rules are given in the captions; illustrative Dice values are not benchmark means. Failure cases appear in Figure~\ref{fig:failure}.

\FloatBarrier
\Needspace{12\baselineskip}
\section{Additional Analyses} \label{app:analyses}

\begin{figure}[H]
\centering
\includegraphics[width=\textwidth]{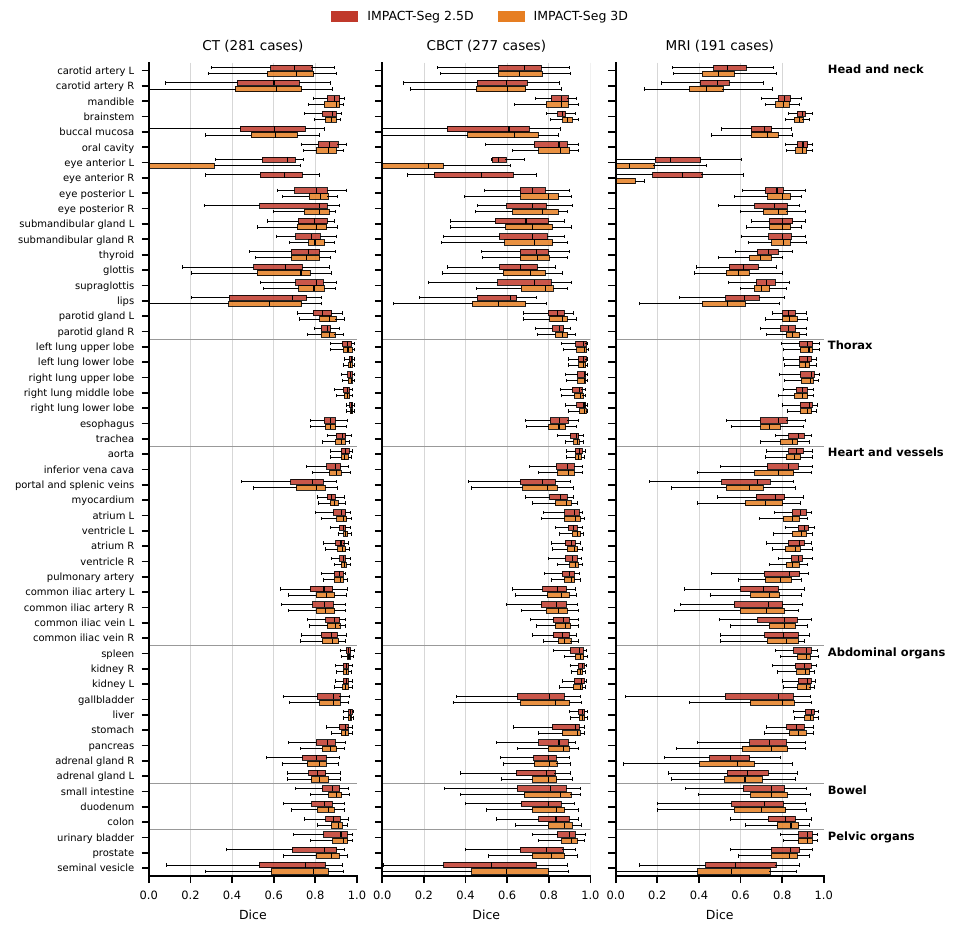}
\caption{Dice of each soft-tissue structure (head and neck, thorax, heart and vessels, abdominal organs, bowel and pelvic organs) on the internal out-of-fold test cases, per modality, for the 2.5D and 3D networks. Boxes give the quartiles over the cases whose reference contains the structure (a missed structure counts as 0), whiskers extend to 1.5 times the interquartile range; outliers are not drawn. MRI cases as in Figure~\ref{fig:internal_groups}.}
\label{fig:internal_labels}
\end{figure}

\begin{figure}[p]
\centering
\includegraphics[width=0.97\textwidth]{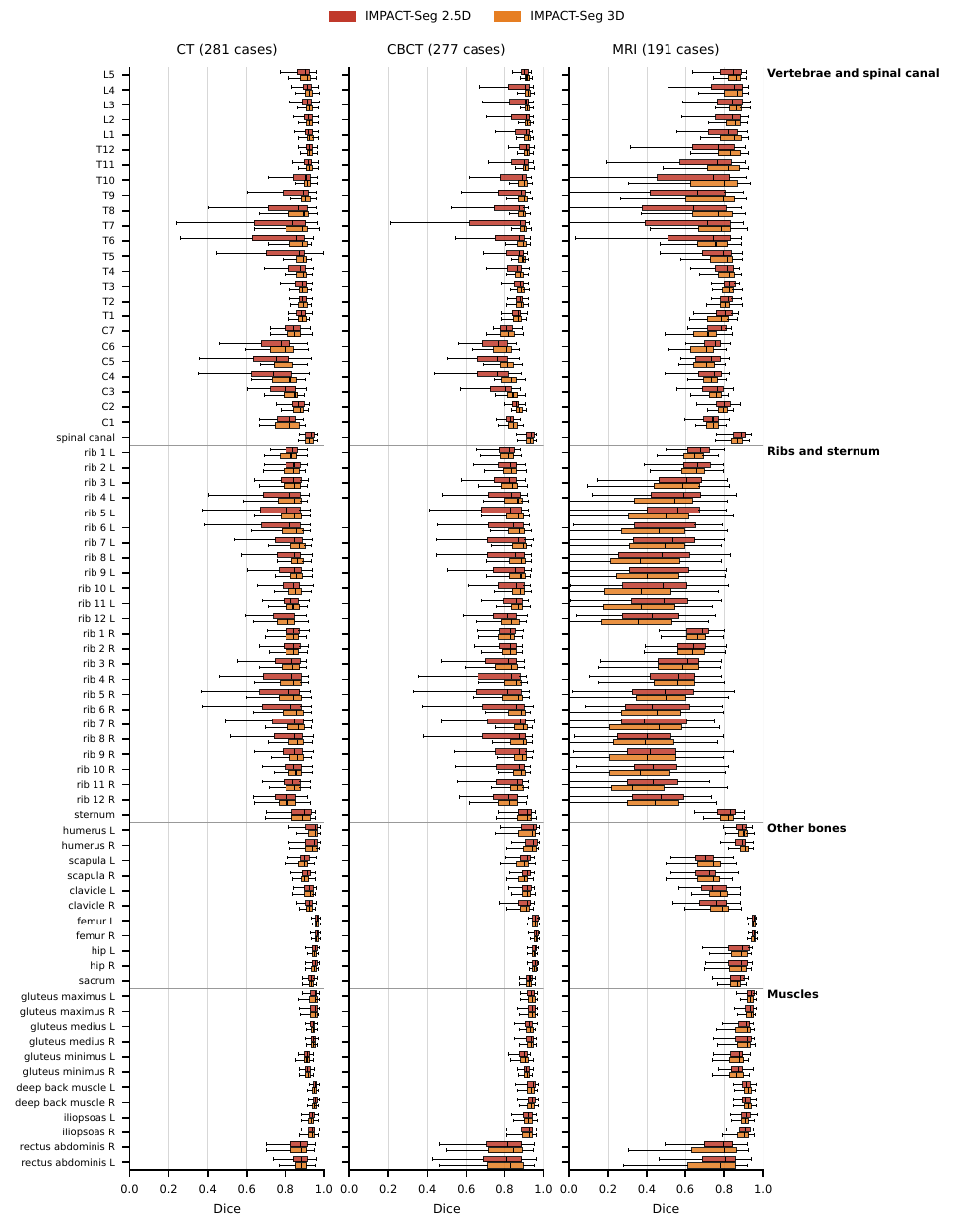}
\caption{Dice of each skeletal structure and muscle on the internal out-of-fold test cases, displayed as in Figure~\ref{fig:internal_labels}.}
\label{fig:internal_labels_skeleton}
\end{figure}

\begin{figure}[!htbp]
\centering
\includegraphics[width=\columnwidth]{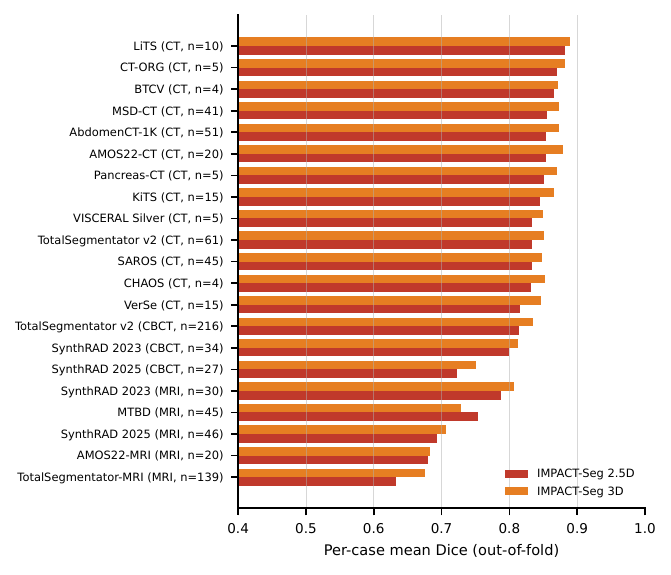}
\caption{Internal out-of-fold Dice of the 2.5D and 3D networks per source cohort and modality (126 structures, per-case mean over the labels present in the reference).}
\label{fig:internal}
\end{figure}

\begin{figure}[!htbp]
\centering
\includegraphics[width=\columnwidth]{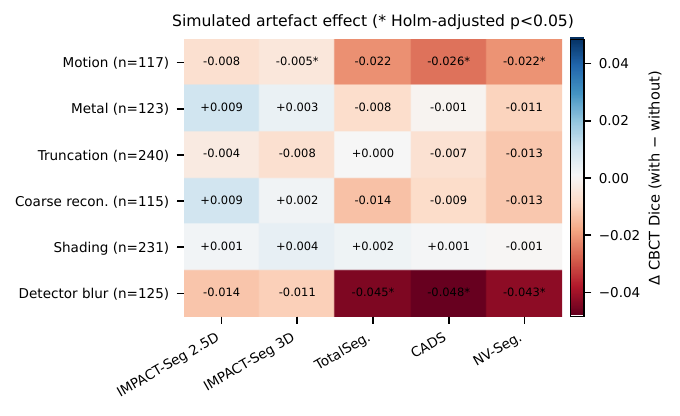}
\caption{Mean CBCT Dice difference between simulated CBCTs with and without each artefact, per method (Mann--Whitney test, Holm correction per method).}
\label{fig:artefacts}
\end{figure}

\begin{figure}[!htbp]
\centering
\includegraphics[width=\textwidth]{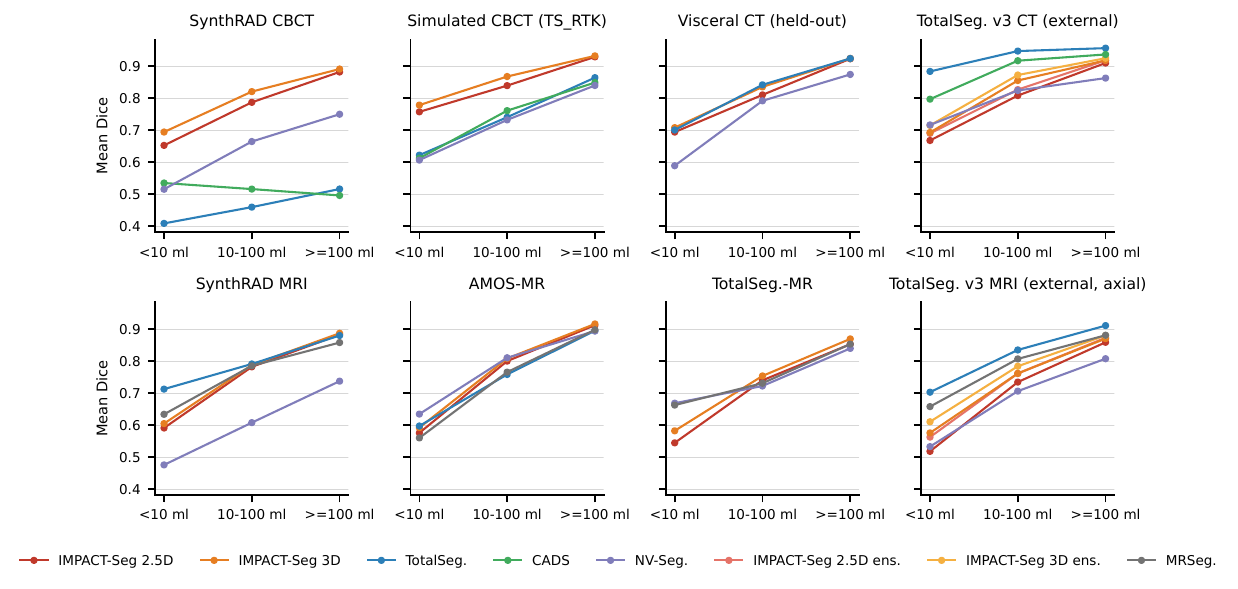}
\caption{Mean Dice by reference structure volume ($<$10~ml, 10--100~ml, $\geq$100~ml) on each benchmark.}
\label{fig:structure_size}
\end{figure}

\begin{figure}[!htbp]
\centering
\includegraphics[width=\textwidth]{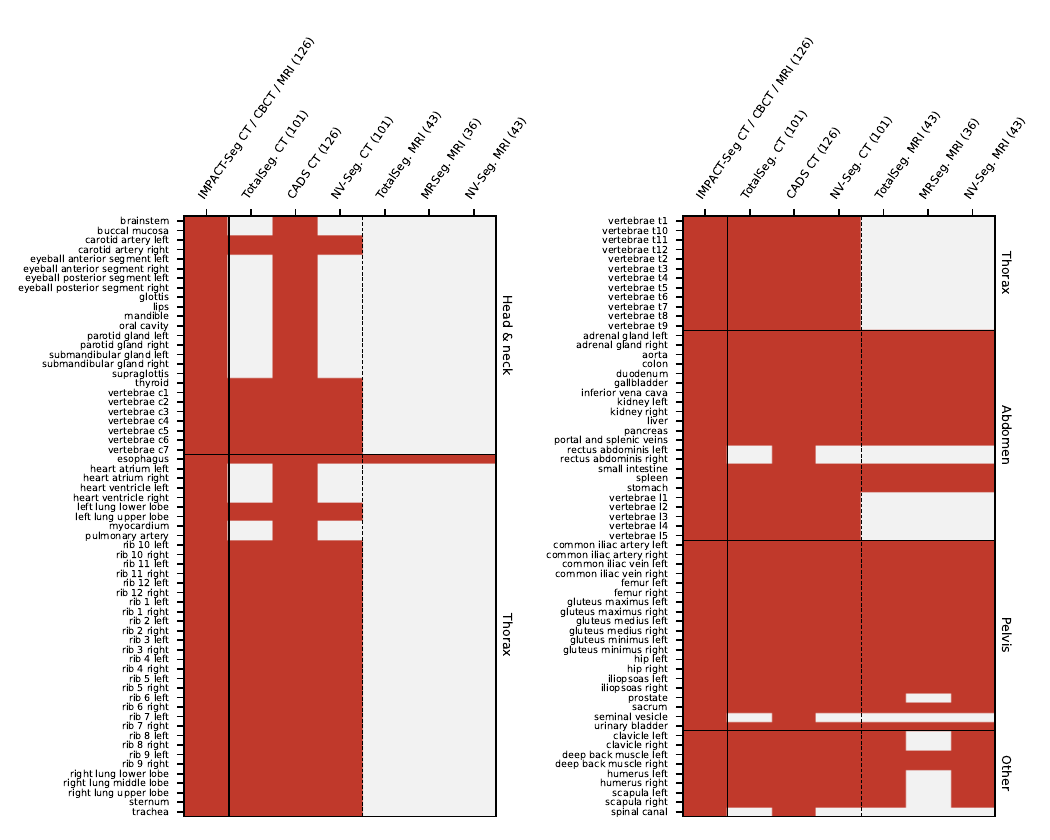}
\caption{Anatomical coverage. For each of the 126 structures of IMPACT-Seg, whether each framework outputs it as a structure of its own on CT and on MRI (red). IMPACT-Seg provides all of them with the same model on CT, CBCT and MRI. Its label space was derived from the CADS tasks, so CADS covers the same structures on CT; TotalSegmentator and NV-Segment-CTMR cover about 100 on CT and 43 on MRI, MRSegmentator 36 on MRI. Whole organs (e.g. heart, lungs) do not count as their substructures.}
\label{fig:coverage}
\end{figure}

Figures~\ref{fig:internal_labels}--\ref{fig:internal} detail internal performance by structure and cohort. Figures~\ref{fig:artefacts}--\ref{fig:coverage} analyse simulated artefacts, structure size, and anatomical coverage. \ifextendedsupplement Figure~\ref{fig:qualitative_reference} complements the median cases with references selected for readability.\fi

\ifextendedsupplement
\begin{figure}[!htbp]
\centering
\includegraphics[width=\textwidth]{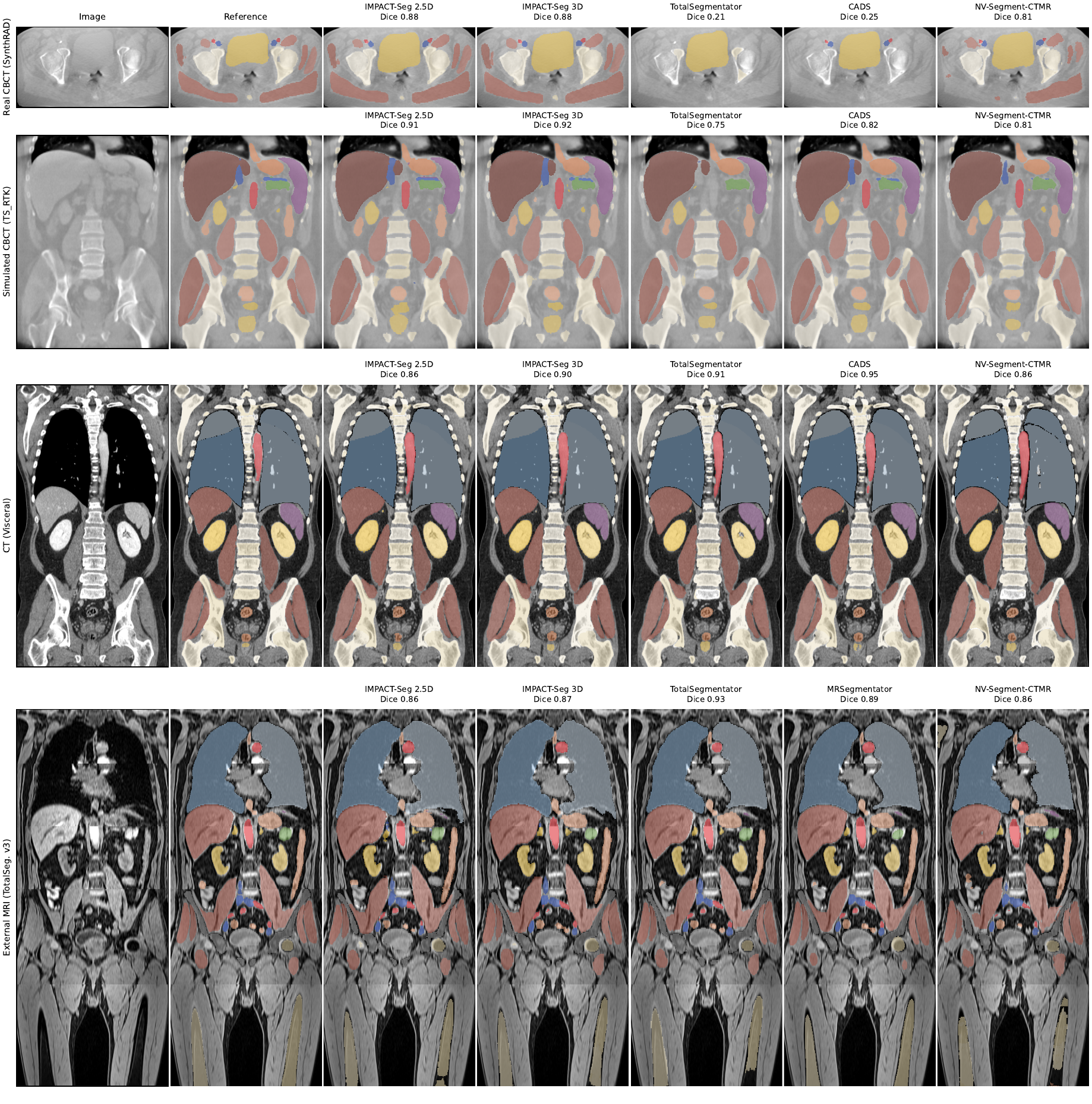}
\caption{Illustrative cases selected for reference agreement: for each benchmark, a case whose reference agrees most closely with the compared methods (competitors on the planning or source CT for the CBCT benchmarks), checked visually. Display as in Figure~\ref{fig:qualitative_median}. These cases are selected for readability and their Dice values are not representative of the benchmark means.}
\label{fig:qualitative_reference}
\end{figure}
\fi

\clearpage
\FloatBarrier
\Needspace{12\baselineskip}
\section{Patch Reconstruction} \label{sec:results_reconstruction}

Overlapping patches can be merged either by keeping, at each voxel, the label of the most confident patch (our default) or, as in nnU-Net, by accumulating the class probabilities of all patches with a Gaussian weighting window and taking the argmax at the end. We compared both strategies with the same 2.5D fold-1 model on the fold-1 cases of every benchmark and of the internal test set, using the same evaluated cases and labels as in the main text (Table~\ref{tab:reconstruction}). Gaussian-weighted averaging changed the mean Dice by at most 0.002 on every benchmark, with median paired differences of 0.000, and the mean HD95 by less than 0.2~mm except on the external TotalSegmentator CT cases (11.8 versus 10.6~mm). Overall Dice was insensitive to the fusion strategy in this 2.5D fold-1 comparison. The principal observed difference was memory use: on a whole-body CT, probability averaging raised the peak GPU memory of the 3D network from 4.3 to 12.1~GB, because the 132 class probabilities of the whole volume must be accumulated, whereas the 2.5D network (batch of 16 slices) was barely affected (3.1 versus 3.3~GB) and the inference time was unchanged. Given the comparable segmentation accuracy observed in 2.5D and the substantially lower memory footprint in 3D, we retained maximum-confidence fusion as the default reconstruction strategy.

% Source: Code/evaluation/paper_v2/results_experiments/reconstruction (summary logs from the archived analysis, internal_*, memory/memory.csv)
\begin{table}[H]
\centering
\caption{Patch reconstruction ablation (2.5D, fold~1). Per-case mean Dice / HD95 (mm) on the fold-1 cases of each benchmark, same evaluated cases and labels as the main comparison; internal: 126 structures, HD95 penalized for missed structures. Bottom: peak GPU memory (GB, $2^{30}$ bytes, as in Table~\ref{tab:runtime}) of one whole-body CT inference (VisceralGC case, RTX A6000; 2.5D with a batch of 16).}
\label{tab:reconstruction}
\small
\fittable{%
\begin{tabular}{lrcc}
\toprule
Benchmark & Cases & Confidence & Gaussian \\
\midrule
SynthRAD CBCT & 37 & 0.822 / 6.02 & 0.824 / 5.97 \\
Simulated CBCT (TS\_RTK) & 99 & 0.835 / 5.77 & 0.837 / 5.73 \\
VISCERAL CT & 14 & 0.822 / 7.10 & 0.823 / 7.09 \\
TotalSegmentator v3 CT & 54 & 0.804 / 11.75 & 0.806 / 10.60 \\
SynthRAD MRI & 11 & 0.798 / 7.79 & 0.799 / 7.95 \\
AMOS-MR + TotalSeg.-MR & 30 & 0.703 / 14.89 & 0.703 / 14.89 \\
TotalSegmentator v3 MRI & 58 & 0.732 / 15.15 & 0.731 / 15.32 \\
Internal test (126 structures) & 170 & 0.747 / 18.34 & 0.746 / 18.27 \\
\midrule
Peak GPU memory, 2.5D (GB) & 1 & 3.1 & 3.3 \\
Peak GPU memory, 3D (GB) & 1 & 4.3 & 12.1 \\
\bottomrule
\end{tabular}
}
\end{table}

\clearpage

\clearpage
\FloatBarrier
\Needspace{12\baselineskip}
\section{Label-wise Results} \label{app:labels}

\begin{landscape}
\setlength{\LTleft}{0pt}
\setlength{\LTright}{0pt}
\tiny
\begin{longtable}{lcccccccc}
\caption{Label-wise comparison on the held-out VISCERAL CT benchmark (unseen patients from the VisceralGC and VisceralGCsilver cohorts) (mean $\pm$ standard deviation over evaluated patient--label pairs; HD95 in mm). Best values for each label in bold.}\label{tab:ct_eval_framework_labels}\\
\toprule
 & \multicolumn{2}{c}{NV-Segment-CTMR} & \multicolumn{2}{c}{TotalSegmentator} & \multicolumn{2}{c}{IMPACT-Seg 2.5D} & \multicolumn{2}{c}{IMPACT-Seg 3D}\\
\cmidrule(lr){2-3}\cmidrule(lr){4-5}\cmidrule(lr){6-7}\cmidrule(lr){8-9}
Label & Dice & HD95 & Dice & HD95 & Dice & HD95 & Dice & HD95\\
\midrule
\endfirsthead
\caption[]{Label-wise comparison on the held-out VISCERAL CT benchmark (unseen patients from the VisceralGC and VisceralGCsilver cohorts) (mean $\pm$ standard deviation over evaluated patient--label pairs; HD95 in mm). Best values for each label in bold. (continued)}\\
\toprule
 & \multicolumn{2}{c}{NV-Segment-CTMR} & \multicolumn{2}{c}{TotalSegmentator} & \multicolumn{2}{c}{IMPACT-Seg 2.5D} & \multicolumn{2}{c}{IMPACT-Seg 3D}\\
\cmidrule(lr){2-3}\cmidrule(lr){4-5}\cmidrule(lr){6-7}\cmidrule(lr){8-9}
Label & Dice & HD95 & Dice & HD95 & Dice & HD95 & Dice & HD95\\
\midrule
\endhead
\bottomrule
\endfoot
Aorta & \textbf{0.857 ± 0.064} & \textbf{6.389 ± 8.085} & 0.851 ± 0.061 & 7.014 ± 8.175 & 0.856 ± 0.060 & 6.594 ± 8.114 & 0.851 ± 0.062 & 6.877 ± 8.317\\
Colon & 0.880 ± 0.038 & 7.376 ± 5.039 & \textbf{0.906 ± 0.031} & 4.930 ± 2.919 & 0.877 ± 0.040 & 7.794 ± 5.302 & 0.899 ± 0.028 & \textbf{4.695 ± 2.239}\\
Duodenum & 0.540 ± 0.084 & 6.567 ± 3.645 & \textbf{0.828 ± 0.081} & \textbf{4.884 ± 3.051} & 0.789 ± 0.110 & 6.980 ± 5.383 & 0.817 ± 0.097 & 5.203 ± 3.914\\
Esophagus & 0.812 ± 0.075 & 4.058 ± 4.327 & \textbf{0.834 ± 0.089} & 4.175 ± 5.321 & 0.827 ± 0.078 & \textbf{3.902 ± 4.396} & 0.822 ± 0.081 & 4.043 ± 4.684\\
Gallbladder & 0.700 ± 0.159 & 9.025 ± 8.500 & \textbf{0.727 ± 0.132} & \textbf{7.018 ± 4.499} & 0.701 ± 0.144 & 7.695 ± 6.328 & 0.706 ± 0.145 & 7.229 ± 4.304\\
Inferior vena cava & 0.764 ± 0.092 & 9.012 ± 8.597 & 0.753 ± 0.082 & 8.397 ± 2.967 & \textbf{0.786 ± 0.078} & 7.597 ± 4.444 & 0.782 ± 0.080 & \textbf{7.567 ± 3.802}\\
Left adrenal gland & 0.128 ± 0.081 & 8.117 ± 5.081 & \textbf{0.619 ± 0.148} & \textbf{7.321 ± 6.183} & 0.595 ± 0.129 & 9.362 ± 7.124 & 0.601 ± 0.138 & 7.655 ± 5.480\\
Left autochthon & 0.922 ± 0.048 & 3.386 ± 3.109 & \textbf{0.945 ± 0.053} & 2.520 ± 3.117 & 0.941 ± 0.038 & \textbf{2.503 ± 2.113} & 0.935 ± 0.053 & 2.783 ± 2.989\\
\end{longtable}
\end{landscape}

\begin{landscape}
\setlength{\LTleft}{0pt}
\setlength{\LTright}{0pt}
\tiny
\begin{longtable}{lcccccccccc}
\caption{Label-wise comparison on the external TotalSegmentator v3 CT cases (absent from v2) (mean $\pm$ standard deviation over evaluated patient--label pairs; HD95 in mm). Best values for each label in bold. IMPACT-Seg: five-fold ensembles, as these external cases were seen by no fold.}\label{tab:ts_v3_ct_framework_labels}\\
\toprule
 & \multicolumn{2}{c}{CADS} & \multicolumn{2}{c}{NV-Segment-CTMR} & \multicolumn{2}{c}{TotalSegmentator} & \multicolumn{2}{c}{IMPACT-Seg 2.5D (5-fold ens.)} & \multicolumn{2}{c}{IMPACT-Seg 3D (5-fold ens.)}\\
\cmidrule(lr){2-3}\cmidrule(lr){4-5}\cmidrule(lr){6-7}\cmidrule(lr){8-9}\cmidrule(lr){10-11}
Label & Dice & HD95 & Dice & HD95 & Dice & HD95 & Dice & HD95 & Dice & HD95\\
\midrule
\endfirsthead
\caption[]{Label-wise comparison on the external TotalSegmentator v3 CT cases (absent from v2) (mean $\pm$ standard deviation over evaluated patient--label pairs; HD95 in mm). Best values for each label in bold. IMPACT-Seg: five-fold ensembles, as these external cases were seen by no fold. (continued)}\\
\toprule
 & \multicolumn{2}{c}{CADS} & \multicolumn{2}{c}{NV-Segment-CTMR} & \multicolumn{2}{c}{TotalSegmentator} & \multicolumn{2}{c}{IMPACT-Seg 2.5D (5-fold ens.)} & \multicolumn{2}{c}{IMPACT-Seg 3D (5-fold ens.)}\\
\cmidrule(lr){2-3}\cmidrule(lr){4-5}\cmidrule(lr){6-7}\cmidrule(lr){8-9}\cmidrule(lr){10-11}
Label & Dice & HD95 & Dice & HD95 & Dice & HD95 & Dice & HD95 & Dice & HD95\\
\midrule
\endhead
\bottomrule
\endfoot
Aorta & 0.936 ± 0.084 & 3.831 ± 7.648 & 0.902 ± 0.107 & 18.837 ± 104.820 & \textbf{0.962 ± 0.084} & \textbf{2.436 ± 5.701} & 0.926 ± 0.124 & 3.499 ± 8.216 & 0.931 ± 0.105 & 3.880 ± 9.501\\
Colon & 0.883 ± 0.103 & 8.338 ± 9.845 & 0.827 ± 0.111 & 46.809 ± 110.132 & \textbf{0.936 ± 0.050} & \textbf{3.394 ± 3.444} & 0.836 ± 0.110 & 15.126 ± 14.943 & 0.874 ± 0.098 & 9.500 ± 10.774\\
Duodenum & 0.841 ± 0.121 & 6.870 ± 10.679 & 0.767 ± 0.146 & 14.693 ± 51.344 & \textbf{0.905 ± 0.074} & \textbf{3.228 ± 5.706} & 0.757 ± 0.153 & 10.522 ± 11.215 & 0.802 ± 0.150 & 8.027 ± 10.910\\
Esophagus & 0.873 ± 0.154 & 5.037 ± 17.976 & 0.822 ± 0.151 & 12.094 ± 68.747 & \textbf{0.934 ± 0.070} & \textbf{3.257 ± 15.519} & 0.838 ± 0.153 & 5.388 ± 16.428 & 0.843 ± 0.128 & 4.738 ± 15.711\\
Gallbladder & 0.803 ± 0.211 & 6.841 ± 8.706 & 0.785 ± 0.213 & 28.822 ± 94.853 & \textbf{0.917 ± 0.074} & \textbf{2.543 ± 1.891} & 0.702 ± 0.293 & 11.781 ± 15.698 & 0.740 ± 0.283 & 8.362 ± 11.584\\
Inferior vena cava & 0.899 ± 0.101 & 5.274 ± 11.616 & 0.830 ± 0.146 & 22.408 ± 82.952 & \textbf{0.926 ± 0.091} & \textbf{2.747 ± 4.402} & 0.821 ± 0.147 & 6.722 ± 9.481 & 0.850 ± 0.114 & 5.151 ± 5.151\\
Left adrenal gland & 0.883 ± 0.094 & 2.093 ± 1.645 & 0.816 ± 0.140 & 14.993 ± 62.687 & \textbf{0.932 ± 0.071} & \textbf{1.702 ± 1.185} & 0.793 ± 0.153 & 4.298 ± 5.023 & 0.817 ± 0.137 & 3.071 ± 3.554\\
Left autochthon & 0.966 ± 0.053 & 2.432 ± 6.772 & 0.882 ± 0.175 & 9.935 ± 45.462 & \textbf{0.979 ± 0.022} & \textbf{1.541 ± 0.262} & 0.941 ± 0.089 & 2.547 ± 2.916 & 0.946 ± 0.048 & 2.147 ± 1.790\\
\end{longtable}
\end{landscape}

\begin{landscape}
\setlength{\LTleft}{0pt}
\setlength{\LTright}{0pt}
\tiny
\begin{longtable}{lcccccccccc}
\caption{Label-wise comparison on the curated SynthRADSegCBCT subset (mean $\pm$ standard deviation over evaluated patient--label pairs; HD95 in mm). Best values for each label in bold.}\label{tab:synthrad_cbct_framework_organs}\\
\toprule
 & \multicolumn{2}{c}{CADS} & \multicolumn{2}{c}{NV-Segment-CTMR} & \multicolumn{2}{c}{TotalSegmentator} & \multicolumn{2}{c}{IMPACT-Seg 2.5D} & \multicolumn{2}{c}{IMPACT-Seg 3D}\\
\cmidrule(lr){2-3}\cmidrule(lr){4-5}\cmidrule(lr){6-7}\cmidrule(lr){8-9}\cmidrule(lr){10-11}
Label & Dice & HD95 & Dice & HD95 & Dice & HD95 & Dice & HD95 & Dice & HD95\\
\midrule
\endfirsthead
\caption[]{Label-wise comparison on the curated SynthRADSegCBCT subset (mean $\pm$ standard deviation over evaluated patient--label pairs; HD95 in mm). Best values for each label in bold. (continued)}\\
\toprule
 & \multicolumn{2}{c}{CADS} & \multicolumn{2}{c}{NV-Segment-CTMR} & \multicolumn{2}{c}{TotalSegmentator} & \multicolumn{2}{c}{IMPACT-Seg 2.5D} & \multicolumn{2}{c}{IMPACT-Seg 3D}\\
\cmidrule(lr){2-3}\cmidrule(lr){4-5}\cmidrule(lr){6-7}\cmidrule(lr){8-9}\cmidrule(lr){10-11}
Label & Dice & HD95 & Dice & HD95 & Dice & HD95 & Dice & HD95 & Dice & HD95\\
\midrule
\endhead
\bottomrule
\endfoot
Aorta & 0.764 ± 0.248 & 13.742 ± 19.299 & 0.781 ± 0.114 & 15.189 ± 19.277 & 0.457 ± 0.419 & 15.760 ± 22.593 & 0.851 ± 0.081 & 6.741 ± 9.256 & \textbf{0.861 ± 0.086} & \textbf{6.551 ± 10.990}\\
Colon & 0.533 ± 0.259 & 40.568 ± 35.919 & 0.629 ± 0.150 & 31.771 ± 23.608 & 0.504 ± 0.300 & 46.343 ± 45.732 & 0.662 ± 0.147 & 25.668 ± 19.851 & \textbf{0.704 ± 0.126} & \textbf{20.122 ± 14.567}\\
Duodenum & 0.590 ± 0.199 & 21.172 ± 22.667 & 0.440 ± 0.220 & 32.011 ± 23.631 & 0.549 ± 0.209 & 18.446 ± 20.342 & 0.606 ± 0.134 & 12.883 ± 9.014 & \textbf{0.648 ± 0.130} & \textbf{11.883 ± 11.738}\\
Esophagus & 0.646 ± 0.195 & 13.823 ± 23.911 & 0.648 ± 0.126 & 7.926 ± 7.393 & 0.670 ± 0.176 & 9.257 ± 12.337 & 0.724 ± 0.131 & 5.933 ± 5.953 & \textbf{0.770 ± 0.079} & \textbf{4.494 ± 2.431}\\
Gallbladder & 0.524 ± 0.263 & 10.432 ± 5.895 & 0.356 ± 0.401 & 17.480 ± 12.584 & 0.483 ± 0.373 & 12.954 ± 9.544 & \textbf{0.683 ± 0.157} & \textbf{6.468 ± 2.293} & 0.562 ± 0.326 & 8.191 ± 5.279\\
Inferior vena cava & 0.494 ± 0.275 & 27.037 ± 42.452 & 0.516 ± 0.262 & 22.347 ± 28.959 & 0.363 ± 0.327 & 13.779 ± 15.689 & 0.707 ± 0.108 & 7.744 ± 5.063 & \textbf{0.730 ± 0.106} & \textbf{7.256 ± 7.045}\\
Left adrenal gland & 0.580 ± 0.163 & \textbf{5.175 ± 2.222} & 0.515 ± 0.161 & 7.278 ± 4.549 & 0.551 ± 0.212 & 6.896 ± 5.986 & 0.600 ± 0.109 & 5.804 ± 3.884 & \textbf{0.603 ± 0.130} & 5.529 ± 3.077\\
Left autochthon & 0.447 ± 0.397 & 14.704 ± 20.957 & 0.785 ± 0.231 & 8.608 ± 13.951 & 0.490 ± 0.407 & 14.475 ± 32.062 & 0.896 ± 0.057 & 3.237 ± 1.291 & \textbf{0.901 ± 0.044} & \textbf{3.040 ± 1.075}\\
\end{longtable}
\end{landscape}

\begin{landscape}
\setlength{\LTleft}{0pt}
\setlength{\LTright}{0pt}
\tiny
\begin{longtable}{lcccccccccc}
\caption{Label-wise comparison on the TS\_RTK simulated CBCT benchmark (mean $\pm$ standard deviation over evaluated patient--label pairs; HD95 in mm). Best values for each label in bold.}\label{tab:cbct_sim_framework_labels}\\
\toprule
 & \multicolumn{2}{c}{CADS} & \multicolumn{2}{c}{NV-Segment-CTMR} & \multicolumn{2}{c}{TotalSegmentator} & \multicolumn{2}{c}{IMPACT-Seg 2.5D} & \multicolumn{2}{c}{IMPACT-Seg 3D}\\
\cmidrule(lr){2-3}\cmidrule(lr){4-5}\cmidrule(lr){6-7}\cmidrule(lr){8-9}\cmidrule(lr){10-11}
Label & Dice & HD95 & Dice & HD95 & Dice & HD95 & Dice & HD95 & Dice & HD95\\
\midrule
\endfirsthead
\caption[]{Label-wise comparison on the TS\_RTK simulated CBCT benchmark (mean $\pm$ standard deviation over evaluated patient--label pairs; HD95 in mm). Best values for each label in bold. (continued)}\\
\toprule
 & \multicolumn{2}{c}{CADS} & \multicolumn{2}{c}{NV-Segment-CTMR} & \multicolumn{2}{c}{TotalSegmentator} & \multicolumn{2}{c}{IMPACT-Seg 2.5D} & \multicolumn{2}{c}{IMPACT-Seg 3D}\\
\cmidrule(lr){2-3}\cmidrule(lr){4-5}\cmidrule(lr){6-7}\cmidrule(lr){8-9}\cmidrule(lr){10-11}
Label & Dice & HD95 & Dice & HD95 & Dice & HD95 & Dice & HD95 & Dice & HD95\\
\midrule
\endhead
\bottomrule
\endfoot
Aorta & 0.909 ± 0.137 & 3.415 ± 4.996 & 0.883 ± 0.104 & 5.502 ± 8.613 & 0.805 ± 0.228 & 15.576 ± 33.149 & \textbf{0.925 ± 0.079} & 3.211 ± 4.691 & 0.924 ± 0.075 & \textbf{2.890 ± 3.313}\\
Colon & 0.790 ± 0.189 & 15.339 ± 25.884 & 0.754 ± 0.184 & 15.788 ± 19.835 & 0.849 ± 0.150 & \textbf{9.251 ± 15.886} & 0.813 ± 0.122 & 14.703 ± 14.030 & \textbf{0.853 ± 0.099} & 10.655 ± 13.122\\
Duodenum & 0.780 ± 0.174 & 7.359 ± 9.866 & 0.723 ± 0.201 & 9.273 ± 9.516 & \textbf{0.847 ± 0.117} & \textbf{4.220 ± 5.201} & 0.750 ± 0.174 & 9.774 ± 10.939 & 0.777 ± 0.177 & 8.498 ± 10.963\\
Esophagus & 0.740 ± 0.218 & 6.957 ± 12.593 & 0.755 ± 0.133 & 5.074 ± 3.739 & \textbf{0.848 ± 0.083} & 3.203 ± 3.365 & 0.838 ± 0.111 & 3.231 ± 2.507 & 0.835 ± 0.111 & \textbf{2.909 ± 1.717}\\
Gallbladder & 0.696 ± 0.245 & 9.587 ± 10.033 & 0.663 ± 0.254 & 11.460 ± 11.368 & \textbf{0.769 ± 0.205} & 7.926 ± 9.135 & 0.751 ± 0.199 & 10.042 ± 12.082 & 0.756 ± 0.234 & \textbf{7.463 ± 10.857}\\
Inferior vena cava & 0.826 ± 0.153 & 6.795 ± 12.262 & 0.781 ± 0.171 & 9.866 ± 12.310 & 0.637 ± 0.293 & 26.219 ± 47.637 & 0.860 ± 0.089 & 4.611 ± 4.803 & \textbf{0.864 ± 0.100} & \textbf{4.514 ± 5.328}\\
Left adrenal gland & 0.772 ± 0.153 & 3.031 ± 2.121 & 0.704 ± 0.185 & 4.391 ± 3.835 & \textbf{0.826 ± 0.111} & \textbf{2.800 ± 9.094} & 0.747 ± 0.142 & 4.603 ± 4.405 & 0.767 ± 0.143 & 3.436 ± 3.793\\
Left autochthon & 0.796 ± 0.204 & 14.926 ± 23.482 & 0.849 ± 0.137 & 7.639 ± 8.079 & 0.848 ± 0.155 & 10.495 ± 18.184 & \textbf{0.940 ± 0.052} & 2.636 ± 2.636 & 0.938 ± 0.041 & \textbf{2.510 ± 1.374}\\
\end{longtable}
\end{landscape}

\begin{landscape}
\setlength{\LTleft}{0pt}
\setlength{\LTright}{0pt}
\tiny
\begin{longtable}{lcccccccccc}
\caption{Label-wise comparison on the curated SynthRAD MRI subset (mean $\pm$ standard deviation over evaluated patient--label pairs; HD95 in mm). Best values for each label in bold.}\label{tab:mri_framework_labels}\\
\toprule
 & \multicolumn{2}{c}{MRSegmentator} & \multicolumn{2}{c}{NV-Segment-CTMR} & \multicolumn{2}{c}{TotalSegmentator} & \multicolumn{2}{c}{IMPACT-Seg 2.5D} & \multicolumn{2}{c}{IMPACT-Seg 3D}\\
\cmidrule(lr){2-3}\cmidrule(lr){4-5}\cmidrule(lr){6-7}\cmidrule(lr){8-9}\cmidrule(lr){10-11}
Label & Dice & HD95 & Dice & HD95 & Dice & HD95 & Dice & HD95 & Dice & HD95\\
\midrule
\endfirsthead
\caption[]{Label-wise comparison on the curated SynthRAD MRI subset (mean $\pm$ standard deviation over evaluated patient--label pairs; HD95 in mm). Best values for each label in bold. (continued)}\\
\toprule
 & \multicolumn{2}{c}{MRSegmentator} & \multicolumn{2}{c}{NV-Segment-CTMR} & \multicolumn{2}{c}{TotalSegmentator} & \multicolumn{2}{c}{IMPACT-Seg 2.5D} & \multicolumn{2}{c}{IMPACT-Seg 3D}\\
\cmidrule(lr){2-3}\cmidrule(lr){4-5}\cmidrule(lr){6-7}\cmidrule(lr){8-9}\cmidrule(lr){10-11}
Label & Dice & HD95 & Dice & HD95 & Dice & HD95 & Dice & HD95 & Dice & HD95\\
\midrule
\endhead
\bottomrule
\endfoot
Aorta & \textbf{0.846 ± 0.120} & \textbf{3.957 ± 3.767} & 0.751 ± 0.174 & 12.489 ± 19.131 & 0.828 ± 0.126 & 11.945 ± 21.050 & 0.828 ± 0.124 & 4.404 ± 2.911 & 0.834 ± 0.147 & 6.748 ± 14.006\\
Bladder & 0.901 ± 0.087 & 6.462 ± 9.216 & 0.873 ± 0.091 & 6.627 ± 8.117 & \textbf{0.960 ± 0.067} & \textbf{2.608 ± 4.100} & 0.882 ± 0.075 & 13.397 ± 32.674 & 0.881 ± 0.102 & 5.634 ± 4.859\\
Colon & 0.802 ± 0.111 & 12.701 ± 9.247 & 0.746 ± 0.097 & 17.391 ± 12.880 & \textbf{0.854 ± 0.138} & \textbf{10.748 ± 18.025} & 0.740 ± 0.096 & 17.602 ± 9.982 & 0.784 ± 0.088 & 15.098 ± 10.310\\
Duodenum & 0.681 ± 0.200 & 9.681 ± 12.567 & 0.477 ± 0.257 & 22.286 ± 20.051 & \textbf{0.737 ± 0.154} & \textbf{9.258 ± 11.439} & 0.559 ± 0.253 & 20.729 ± 43.774 & 0.584 ± 0.290 & 9.602 ± 7.008\\
Esophagus & 0.756 ± 0.090 & 6.141 ± 5.971 & 0.646 ± 0.131 & 7.360 ± 4.998 & 0.741 ± 0.138 & 7.722 ± 9.023 & \textbf{0.767 ± 0.063} & 5.464 ± 7.165 & 0.766 ± 0.062 & \textbf{4.839 ± 4.875}\\
Gallbladder & \textbf{0.782 ± 0.108} & \textbf{5.421 ± 3.230} & 0.714 ± 0.191 & 9.248 ± 11.420 & 0.719 ± 0.239 & 8.199 ± 8.583 & 0.650 ± 0.181 & 13.363 ± 9.708 & 0.621 ± 0.208 & 8.053 ± 3.304\\
Heart & 0.869 ± 0.032 & 16.401 ± 9.649 & 0.837 ± 0.062 & 16.675 ± 25.222 & 0.869 ± 0.050 & 11.333 ± 5.472 & \textbf{0.904 ± 0.049} & \textbf{5.228 ± 1.739} & 0.903 ± 0.041 & 5.795 ± 2.544\\
Inferior vena cava & \textbf{0.708 ± 0.178} & \textbf{6.021 ± 2.663} & 0.608 ± 0.216 & 15.331 ± 24.213 & 0.688 ± 0.287 & 10.745 ± 11.876 & 0.697 ± 0.175 & 8.388 ± 8.458 & 0.692 ± 0.159 & 7.163 ± 4.734\\
\end{longtable}
\end{landscape}

\begin{landscape}
\setlength{\LTleft}{0pt}
\setlength{\LTright}{0pt}
\tiny
\begin{longtable}{lcccccccccc}
\caption{Label-wise comparison on the AMOS22 MRI test cases (mean $\pm$ standard deviation over evaluated patient--label pairs; HD95 in mm). Best values for each label in bold.}\label{tab:mri_amos_framework_labels}\\
\toprule
 & \multicolumn{2}{c}{MRSegmentator} & \multicolumn{2}{c}{NV-Segment-CTMR} & \multicolumn{2}{c}{TotalSegmentator} & \multicolumn{2}{c}{IMPACT-Seg 2.5D} & \multicolumn{2}{c}{IMPACT-Seg 3D}\\
\cmidrule(lr){2-3}\cmidrule(lr){4-5}\cmidrule(lr){6-7}\cmidrule(lr){8-9}\cmidrule(lr){10-11}
Label & Dice & HD95 & Dice & HD95 & Dice & HD95 & Dice & HD95 & Dice & HD95\\
\midrule
\endfirsthead
\caption[]{Label-wise comparison on the AMOS22 MRI test cases (mean $\pm$ standard deviation over evaluated patient--label pairs; HD95 in mm). Best values for each label in bold. (continued)}\\
\toprule
 & \multicolumn{2}{c}{MRSegmentator} & \multicolumn{2}{c}{NV-Segment-CTMR} & \multicolumn{2}{c}{TotalSegmentator} & \multicolumn{2}{c}{IMPACT-Seg 2.5D} & \multicolumn{2}{c}{IMPACT-Seg 3D}\\
\cmidrule(lr){2-3}\cmidrule(lr){4-5}\cmidrule(lr){6-7}\cmidrule(lr){8-9}\cmidrule(lr){10-11}
Label & Dice & HD95 & Dice & HD95 & Dice & HD95 & Dice & HD95 & Dice & HD95\\
\midrule
\endhead
\bottomrule
\endfoot
Aorta & 0.892 ± 0.065 & 12.610 ± 15.955 & \textbf{0.904 ± 0.044} & \textbf{2.490 ± 1.429} & 0.834 ± 0.055 & 6.000 ± 7.257 & 0.895 ± 0.045 & 3.615 ± 3.556 & 0.891 ± 0.050 & 4.296 ± 5.658\\
Colon & 0.840 ± 0.079 & 7.097 ± 5.508 & 0.796 ± 0.090 & 8.726 ± 7.486 & \textbf{0.848 ± 0.071} & \textbf{4.997 ± 4.088} & 0.821 ± 0.064 & 9.087 ± 7.527 & 0.827 ± 0.075 & 7.412 ± 5.677\\
Duodenum & 0.596 ± 0.158 & 14.624 ± 8.721 & \textbf{0.752 ± 0.119} & \textbf{8.392 ± 7.210} & 0.551 ± 0.185 & 17.415 ± 11.536 & 0.629 ± 0.139 & 13.589 ± 9.853 & 0.648 ± 0.118 & 10.928 ± 6.128\\
Esophagus & 0.723 ± 0.095 & 6.368 ± 5.309 & \textbf{0.759 ± 0.118} & 5.241 ± 7.040 & 0.720 ± 0.095 & 6.772 ± 4.963 & 0.691 ± 0.100 & 5.410 ± 2.508 & 0.700 ± 0.079 & \textbf{4.572 ± 1.856}\\
Gallbladder & 0.678 ± 0.299 & 8.834 ± 9.079 & \textbf{0.871 ± 0.078} & \textbf{3.740 ± 3.714} & 0.667 ± 0.279 & 8.659 ± 6.710 & 0.696 ± 0.260 & 8.999 ± 8.150 & 0.771 ± 0.183 & 6.904 ± 6.770\\
Heart & 0.858 ± 0.037 & 9.099 ± 5.573 & 0.869 ± 0.039 & 7.603 ± 3.488 & 0.859 ± 0.040 & 8.656 ± 4.871 & \textbf{0.940 ± 0.016} & \textbf{3.342 ± 0.646} & 0.923 ± 0.033 & 4.177 ± 1.923\\
Inferior vena cava & 0.785 ± 0.057 & 6.192 ± 6.072 & 0.832 ± 0.029 & 3.740 ± 0.915 & 0.694 ± 0.101 & 11.934 ± 10.653 & 0.852 ± 0.047 & 3.546 ± 2.304 & \textbf{0.868 ± 0.032} & \textbf{2.861 ± 0.624}\\
Left adrenal gland & 0.545 ± 0.168 & 8.365 ± 5.130 & \textbf{0.732 ± 0.067} & \textbf{3.357 ± 2.392} & 0.609 ± 0.118 & 6.238 ± 2.746 & 0.533 ± 0.174 & 9.044 ± 5.724 & 0.581 ± 0.121 & 5.642 ± 2.481\\
\end{longtable}
\end{landscape}

\begin{landscape}
\setlength{\LTleft}{0pt}
\setlength{\LTright}{0pt}
\tiny
\begin{longtable}{lcccccccc}
\caption{Label-wise comparison on the curated TotalSegmentator MRI test cases (mean $\pm$ standard deviation over evaluated patient--label pairs; HD95 in mm). Best values for each label in bold.}\label{tab:mri_tsmr_framework_labels}\\
\toprule
 & \multicolumn{2}{c}{MRSegmentator} & \multicolumn{2}{c}{NV-Segment-CTMR} & \multicolumn{2}{c}{IMPACT-Seg 2.5D} & \multicolumn{2}{c}{IMPACT-Seg 3D}\\
\cmidrule(lr){2-3}\cmidrule(lr){4-5}\cmidrule(lr){6-7}\cmidrule(lr){8-9}
Label & Dice & HD95 & Dice & HD95 & Dice & HD95 & Dice & HD95\\
\midrule
\endfirsthead
\caption[]{Label-wise comparison on the curated TotalSegmentator MRI test cases (mean $\pm$ standard deviation over evaluated patient--label pairs; HD95 in mm). Best values for each label in bold. (continued)}\\
\toprule
 & \multicolumn{2}{c}{MRSegmentator} & \multicolumn{2}{c}{NV-Segment-CTMR} & \multicolumn{2}{c}{IMPACT-Seg 2.5D} & \multicolumn{2}{c}{IMPACT-Seg 3D}\\
\cmidrule(lr){2-3}\cmidrule(lr){4-5}\cmidrule(lr){6-7}\cmidrule(lr){8-9}
Label & Dice & HD95 & Dice & HD95 & Dice & HD95 & Dice & HD95\\
\midrule
\endhead
\bottomrule
\endfoot
Aorta & 0.712 ± 0.189 & 21.422 ± 33.709 & \textbf{0.771 ± 0.203} & \textbf{13.694 ± 33.312} & 0.769 ± 0.171 & 17.701 ± 38.085 & 0.757 ± 0.184 & 14.031 ± 30.019\\
Bladder & 0.845 ± 0.166 & 7.773 ± 9.523 & \textbf{0.865 ± 0.147} & \textbf{5.540 ± 5.236} & 0.761 ± 0.210 & 13.999 ± 16.215 & 0.784 ± 0.208 & 13.533 ± 18.002\\
Colon & \textbf{0.841 ± 0.107} & 13.777 ± 25.535 & 0.803 ± 0.139 & \textbf{12.179 ± 21.002} & 0.735 ± 0.132 & 28.000 ± 35.725 & 0.805 ± 0.098 & 15.075 ± 24.424\\
Duodenum & 0.765 ± 0.189 & 11.398 ± 16.636 & \textbf{0.785 ± 0.153} & \textbf{7.000 ± 9.944} & 0.665 ± 0.195 & 16.123 ± 17.820 & 0.675 ± 0.240 & 15.303 ± 19.775\\
Esophagus & 0.711 ± 0.183 & 11.056 ± 22.149 & \textbf{0.713 ± 0.162} & \textbf{8.287 ± 11.991} & 0.616 ± 0.159 & 14.659 ± 22.322 & 0.647 ± 0.159 & 10.778 ± 12.217\\
Gallbladder & 0.780 ± 0.197 & 6.881 ± 7.791 & \textbf{0.791 ± 0.179} & \textbf{5.647 ± 5.819} & 0.638 ± 0.270 & 11.676 ± 13.561 & 0.739 ± 0.205 & 8.762 ± 12.028\\
Heart & 0.755 ± 0.178 & 14.148 ± 10.003 & 0.733 ± 0.218 & 15.137 ± 16.069 & \textbf{0.856 ± 0.125} & \textbf{8.091 ± 8.388} & 0.827 ± 0.159 & 8.911 ± 6.502\\
Inferior vena cava & 0.601 ± 0.214 & 17.668 ± 22.174 & 0.543 ± 0.238 & 24.602 ± 28.227 & 0.790 ± 0.125 & 5.750 ± 5.214 & \textbf{0.799 ± 0.118} & \textbf{5.676 ± 5.762}\\
\end{longtable}
\end{landscape}

\begin{landscape}
\setlength{\LTleft}{0pt}
\setlength{\LTright}{0pt}
\tiny
\begin{longtable}{lcccccccccc}
\caption{Label-wise comparison on the external TotalSegmentator v3 axial MRI cases (absent from v2) (mean $\pm$ standard deviation over evaluated patient--label pairs; HD95 in mm). Best values for each label in bold. IMPACT-Seg: five-fold ensembles, as these external cases were seen by no fold.}\label{tab:ts_v3_mri_framework_labels}\\
\toprule
 & \multicolumn{2}{c}{MRSegmentator} & \multicolumn{2}{c}{NV-Segment-CTMR} & \multicolumn{2}{c}{TotalSegmentator} & \multicolumn{2}{c}{IMPACT-Seg 2.5D (5-fold ens.)} & \multicolumn{2}{c}{IMPACT-Seg 3D (5-fold ens.)}\\
\cmidrule(lr){2-3}\cmidrule(lr){4-5}\cmidrule(lr){6-7}\cmidrule(lr){8-9}\cmidrule(lr){10-11}
Label & Dice & HD95 & Dice & HD95 & Dice & HD95 & Dice & HD95 & Dice & HD95\\
\midrule
\endfirsthead
\caption[]{Label-wise comparison on the external TotalSegmentator v3 axial MRI cases (absent from v2) (mean $\pm$ standard deviation over evaluated patient--label pairs; HD95 in mm). Best values for each label in bold. IMPACT-Seg: five-fold ensembles, as these external cases were seen by no fold. (continued)}\\
\toprule
 & \multicolumn{2}{c}{MRSegmentator} & \multicolumn{2}{c}{NV-Segment-CTMR} & \multicolumn{2}{c}{TotalSegmentator} & \multicolumn{2}{c}{IMPACT-Seg 2.5D (5-fold ens.)} & \multicolumn{2}{c}{IMPACT-Seg 3D (5-fold ens.)}\\
\cmidrule(lr){2-3}\cmidrule(lr){4-5}\cmidrule(lr){6-7}\cmidrule(lr){8-9}\cmidrule(lr){10-11}
Label & Dice & HD95 & Dice & HD95 & Dice & HD95 & Dice & HD95 & Dice & HD95\\
\midrule
\endhead
\bottomrule
\endfoot
Aorta & 0.824 ± 0.103 & 11.887 ± 15.796 & 0.809 ± 0.170 & 10.216 ± 15.833 & \textbf{0.836 ± 0.152} & 7.758 ± 13.414 & 0.825 ± 0.135 & 8.038 ± 12.661 & 0.825 ± 0.159 & \textbf{6.975 ± 13.227}\\
Bladder & 0.878 ± 0.159 & 5.877 ± 8.446 & 0.831 ± 0.195 & 15.600 ± 70.189 & \textbf{0.892 ± 0.131} & \textbf{5.242 ± 6.366} & 0.846 ± 0.150 & 9.612 ± 23.478 & 0.868 ± 0.162 & 6.254 ± 9.743\\
Colon & 0.820 ± 0.117 & 12.027 ± 16.650 & 0.762 ± 0.131 & 24.938 ± 38.168 & \textbf{0.844 ± 0.126} & \textbf{10.721 ± 17.065} & 0.767 ± 0.152 & 18.866 ± 23.528 & 0.811 ± 0.133 & 12.575 ± 18.947\\
Duodenum & 0.738 ± 0.177 & 11.282 ± 13.755 & 0.673 ± 0.202 & 19.305 ± 53.588 & \textbf{0.768 ± 0.176} & \textbf{9.302 ± 11.685} & 0.684 ± 0.206 & 12.155 ± 11.221 & 0.712 ± 0.210 & 11.995 ± 14.388\\
Esophagus & 0.791 ± 0.108 & 4.811 ± 5.798 & 0.723 ± 0.164 & 9.898 ± 44.316 & \textbf{0.812 ± 0.107} & \textbf{4.586 ± 11.143} & 0.713 ± 0.153 & 8.568 ± 14.179 & 0.736 ± 0.137 & 7.246 ± 11.305\\
Gallbladder & \textbf{0.796 ± 0.194} & \textbf{7.493 ± 25.692} & 0.746 ± 0.209 & 12.967 ± 45.792 & 0.781 ± 0.214 & 10.243 ± 29.346 & 0.712 ± 0.245 & 9.113 ± 12.368 & 0.762 ± 0.234 & 8.405 ± 21.347\\
Heart & 0.914 ± 0.046 & 8.890 ± 6.856 & 0.884 ± 0.120 & 33.535 ± 134.742 & \textbf{0.924 ± 0.048} & \textbf{7.684 ± 6.361} & 0.805 ± 0.149 & 12.569 ± 5.719 & 0.832 ± 0.103 & 11.433 ± 5.729\\
Inferior vena cava & 0.746 ± 0.140 & 15.136 ± 17.571 & 0.760 ± 0.145 & 12.521 ± 16.721 & \textbf{0.808 ± 0.134} & \textbf{8.000 ± 11.809} & 0.594 ± 0.161 & 24.327 ± 28.331 & 0.616 ± 0.146 & 21.013 ± 21.362\\
\end{longtable}
\end{landscape}

\clearpage
\FloatBarrier
\Needspace{12\baselineskip}
\section{Confidence Intervals}
\begin{table}[H]
\centering
\small
\caption{Per-case mean Dice and 95\% patient-bootstrap confidence intervals for the two IMPACT-Seg networks. Values correspond to the main benchmarks; the bootstrap uses $10\,000$ resamples.}
\label{tab:dice_confidence_intervals}
\begin{tabular}{lcc}
\toprule
Benchmark & IMPACT-Seg 2.5D & IMPACT-Seg 3D\\
\midrule
SynthRAD CBCT & \res{synthradcbct.ims25.dice} \res{synthradcbct.ims25.ci} & \res{synthradcbct.ims3d.dice} \res{synthradcbct.ims3d.ci}\\
Simulated CBCT & \res{tsrtk.ims25.dice} \res{tsrtk.ims25.ci} & \res{tsrtk.ims3d.dice} \res{tsrtk.ims3d.ci}\\
VISCERAL CT & \res{visceral.ims25.dice} \res{visceral.ims25.ci} & \res{visceral.ims3d.dice} \res{visceral.ims3d.ci}\\
TotalSegmentator v3 CT & \res{tsvthreect.ims25.dice} \res{tsvthreect.ims25.ci} & \res{tsvthreect.ims3d.dice} \res{tsvthreect.ims3d.ci}\\
SynthRAD MRI & \res{synthradmri.ims25.dice} \res{synthradmri.ims25.ci} & \res{synthradmri.ims3d.dice} \res{synthradmri.ims3d.ci}\\
AMOS-MR & \res{amos.ims25.dice} \res{amos.ims25.ci} & \res{amos.ims3d.dice} \res{amos.ims3d.ci}\\
TotalSegmentator-MR & \res{tsmr.ims25.dice} \res{tsmr.ims25.ci} & \res{tsmr.ims3d.dice} \res{tsmr.ims3d.ci}\\
TotalSegmentator v3 MRI & \res{tsvthreemri.ims25.dice} \res{tsvthreemri.ims25.ci} & \res{tsvthreemri.ims3d.dice} \res{tsvthreemri.ims3d.ci}\\
\bottomrule
\end{tabular}
\end{table}

\end{document}